\documentclass[journal]{IEEEtran}
\ifCLASSINFOpdf
\else
\fi

\usepackage{cite}
\usepackage{amsmath,mathtools,amsfonts,amsthm} 
\usepackage{amssymb}
\usepackage{bbm}
\usepackage{verbatim}
\usepackage{graphicx}  
\usepackage{enumerate}
\usepackage{caption}
\usepackage{color}
\usepackage{subcaption}

\newcommand{\scr}{\scriptscriptstyle}
\renewcommand{\baselinestretch}{0.9965}  

\newtheorem{ass}{Assumption}
\newtheorem{problem}{Problem}

\newtheorem{lemma}{Lemma}
\newtheorem{proposition}{Proposition}
\newtheorem{remark}{Remark}
\newtheorem{definition}{Definition}
\newtheorem{theorem}{Theorem}

\begin{document}
%
\title{Robust Cooperative Manipulation without Force/Torque Measurements: Control Design and Experiments}
%
%
%

\author{Christos~K. Verginis,~\IEEEmembership{Member,~IEEE,}
        Matteo~Mastellaro,
        and~Dimos~V. Dimarogonas,~\IEEEmembership{Member,~IEEE}
\thanks{The authors are with the Centre for Autonomous Systems and ACCESS Linnaeus Centre, KTH Royal Institute of Technology, Stockholm 10044, Sweden. Emails: \{cverginis, matteoma, dimos\}@kth.se.}
\thanks{This work was supported by the H2020 ERC Starting Grant BUCOPHSYS, the Swedish Research Council (VR), the Knut och Alice Wallenberg Foundation (KAW), the European Union's Horizon 2020 Research and Innovation Programme under the Grant Agreement No. 644128 (AEROWORKS), the Swedish Foundation for Strategic Research (SSF), and the EU H2020 Research and Innovation Programme under GA No. 731869 (Co4Robots).}}
\maketitle

\begin{abstract}
This paper presents two novel control methodologies for the cooperative manipulation of an object by $N$ robotic agents. Firstly, we design an adaptive control protocol which employs quaternion feedback for the object orientation to avoid potential representation singularities. Secondly, we propose a control protocol that guarantees predefined transient and steady-state performance for the object trajectory. Both methodologies are decentralized, since the agents calculate their own signals without communicating with each other, as well as robust to external disturbances and model uncertainties. Moreover, we consider that the grasping points are rigid, and avoid the need for force/torque measurements. Load distribution is also included via a grasp matrix pseudo-inverse to account for potential differences in the agents' power capabilities. Finally, simulation and experimental results with two robotic arms verify the theoretical findings. 
\end{abstract}


%
\IEEEpeerreviewmaketitle

\section{Introduction}
%
%
%
%
\IEEEPARstart{M}{ulti}-agent systems have gained significant attention the last years due to the numerous advantages they yield with respect to single-agent setups. In the case of robotic manipulation, heavy payloads and challenging maneuvers necessitate the employment of multiple robotic agents. Although collaborative manipulation of a single object, both in terms of transportation (regulation) and trajectory tracking, has been considered in the research community the last decades, there still exist several challenges that need to be {taken into} account by on-going research, both in control design as well as experimental evaluation. 

Early works develop control architectures where the robotic agents communicate and share information with each other, and completely decentralized schemes, where each agent uses only local information or observers, avoiding potential communication delays (see, indicatively, \cite{schneider1992object,sugar2002control,khatib1996decentralized,liu1996decentralized,liu1998decentralized,zribi1992adaptive,gudino2004control,wen1992motion,yoshikawa1993coordinated,kopf1989dynamic}). Impedance and hybrid force/position control is the most common methodology used in the related literature \cite{caccavale2000task,caccavale2008six,heck2013internal,erhart2013adaptive,erhart2013impedance,kume2007coordinated,szewczyk2002planning,tsiamis2015cooperative,ficuciello2014cartesian,ponce2016cooperative,gueaieb2007robust,Li_fuzzy2015,wen1992motion,yoshikawa1993coordinated,tzierakis2003independent,kopf1989dynamic,marino2017distributed}, where a desired impedance behavior is imposed potentially with force regulation. Most of the aforementioned works employ force/torque sensors to acquire feedback of the object-robots contact forces/torques, which however {may result in performance decline} due to sensor noise or mounting difficulties. When the grasping object-agents contacts are rigid, the need for such sensors is redundant, since the overall system can be seen as a closed-chain robot. Regarding grasp rigidity, recent technological advances allow  end-effectors to  grasp \textit{rigidly} certain objects, motivating the specific analysis.  



In addition, many works in the related literature consider known dynamic parameters regarding the object and the robotic agents. However, the accurate knowledge of such parameters, such as masses or moments of inertia, can be a challenging issue, especially for complex robotic manipulators. 

Force/torque sensor-free methodologies have been developed in \cite{kume2007coordinated,wen1992motion,zribi1992adaptive,liu1996decentralized,aghili2013adaptive,Li_fuzzy2015,jean1993adaptive,ficuciello2014cartesian,gueaieb2007robust}; \cite{kume2007coordinated} develops a leader-follower communication-based scheme by partly accounting for dynamic parametric uncertainty, whereas \cite{wen1992motion} and \cite{liu1996decentralized} employ partial and full model information, respectively; \cite{zribi1992adaptive} develops an adaptive control scheme that achieves boundedness of the errors based on known disturbance bounds, and \cite{aghili2013adaptive} proposes an adaptive estimator for kinematic uncertainties, whose convergence affects the asymptotic stability of the overall scheme. In \cite{gueaieb2007robust} and \cite{Li_fuzzy2015}  adaptive fuzzy estimators for structural and parametric uncertainty are introduced, with the latter not taking into account the object dynamics; \cite{jean1993adaptive} develops an adaptive protocol that guarantees boundedness of the internal forces, and \cite{ficuciello2014cartesian} employs an approximate force estimator for a human-robot cooperative task. 

{Another important feature is the representation of the agent and object orientation. The most commonly used tools for orientation representation are rotation matrices, Euler angles, unit quaternions, and the angle/axis convention. In this work,  we employ unit quaternions, which do not suffer from representation singularities and can be tuned to avoid undesired local equilibria, issues that characterize the other methods.}

Unit quaternions in the control design of cooperative manipulation tasks have been employed in \cite{caccavale2000task}, where the authors address the gravity-compensated pose regulation of the grasped object, as well as in \cite{caccavale2008six}, where a model-based force-feedback scheme is developed.  

Full model information is employed in the works \cite{heck2013internal,kopf1989dynamic,gudino2004control,schneider1992object,tzierakis2003independent,szewczyk2002planning,yoshikawa1993coordinated,erhart2013impedance}; \cite{gudino2004control} employs a velocity estimator, \cite{tzierakis2003independent} uses a linearized model, and \cite{erhart2013impedance,erhart2013adaptive} considers kinematic and grasping uncertainties.  Adaptive control schemes are developed in \cite{ponce2016cooperative}, where redundancy is used for obstacle avoidance and \cite{sun2002adaptive}, where the {object dynamics are not taken} into account; \cite{petitti2016decentralized} and \cite{wang2015multi} propose protocols based on graph-based communication by neglecting parts of the overall system dynamics, and \cite{tsiamis2015cooperative}, \cite{wang2015multi} consider leader-follower approaches. An observer-based (for state and task estimation) adaptive control scheme is proposed in \cite{marino2017distributed}. Model-based force-control control protocols with unilateral constraints are developed in \cite{yun1993object,alonso2017multi}. Formation control approaches are considered in \cite{bai2010cooperative,alonso2017multi} and a navigation-function scheme is used in \cite{tanner2003nonholonomic}; \cite{murphey2008adaptive} includes hybrid control with intermittent contacts and in our previous works \cite{MED17_mpc,ECC18_mpc} we considered MPC approaches for cooperative object transportation. Finally, internal force and load distribution analysis in cooperative manipulation tasks is performed in a variety of works (e.g., \cite{walker1991analysis,williams1993virtual,chung2005analysis,erhart2015internal,erhart2016model}).

Note that most of the aforementioned adaptive control schemes (except e.g., \cite{gueaieb2007robust}) employ the usual regressor matrix technique to compensate for unknown dynamic parameters \cite{Slotine_adaptive87,siciliano2010robotics}, which assumes a known structure of the dynamic terms. {Such structures can still be difficult to obtain accurately, especially} when complex manipulators are considered. Moreover, in terms of load distribution, many of the related works use load sharing coefficients (e.g., \cite{liu1996decentralized,liu1998decentralized,gueaieb2007robust}), without proving that undesired internal forces do not arise, or the standard Moore-Penrose inverse of the grasp matrix (e.g. \cite{zribi1992adaptive,szewczyk2002planning}), which has been questioned in \cite{walker1991analysis}.


\subsection{Contribution and Outline}

{In this paper we propose two novel nonlinear control protocols for the trajectory tracking of an object that is rigidly grasped by $N$ robotic agents, without using force/torque measurements at the grasping points. More specifically, our contribution lies in the following attributes:}

{
	\begin{enumerate}
\item Firstly, we develop a decentralized control scheme that combines 
	\begin{itemize}
		\item adaptive control ideas to compensate for external disturbances and uncertainties of the agents' and the object's dynamic parameters,
		\item quaternion modeling of the object's orientation that avoids undesired representation singularities.		 
	\end{itemize}
\item Secondly, we propose a decentralized control scheme that does not depend on the dynamic structure or parameters of the overall system and guarantees \textit{predefined} transient and steady-state performance for the object's center of mass, using the Prescribed Performance Control (PPC) scheme \cite{bechlioulis2008robust}.
\item We carry out extensive simulation studies and experimental results that verify the 
theoretical findings.
\end{enumerate}
Moreover, both control schemes employ the load distribution proposed in \cite{erhart2015internal} that provably avoids undesired internal forces.}

{The first control scheme is an extension of our preliminary work \cite{Verginis_ifac}, where  
we designed a similar adaptive quaternion-based controller, guaranteeing, however, only \textit{local} stability, and no experimental validation was provided. Furthermore, we have employed the PPC scheme in our
previous work \cite{verginis2017timed} to design timed transition systems for a cooperatively manipulated object. In this work, however, we perform a more extended and detailed analysis by deriving specific bounds for the inputs of the robotic arms (i.e., joint velocities and torques), as well as real-time experiments. It is worth noting that PPC has been also used for single manipulation tasks in \cite{bechlioulis2014robust,karayiannidis2012model,doulgeri2009prescribed}. }

The rest of the paper is organized as follows. Section \ref{sec:Notation and Preliminaries} provides the notation used throughout the paper and necessary background. The modeling of the system as well as the problem formulation are given in Section \ref{sec: Problem Formulation}. Section \ref{sec:Main Results} presents the details of the two proposed control schemes with the corresponding stability analysis, and Section \ref{sec:sim and exps} illustrates the simulation and experimental results. Finally, Section \ref{sec:Conclusion and FW} concludes the paper.

\section{Notation and Preliminaries} \label{sec:Notation and Preliminaries}

\subsection{Notation} \label{subsec:Notation}
The set of positive integers is denoted by $\mathbb{N}$ and the real $n$-coordinate space, with $n\in\mathbb{N}$, by $\mathbb{R}^n$;
$\mathbb{R}^n_{\geq 0}$ and $\mathbb{R}^n_{> 0}$ are the sets of real $n$-vectors with all elements nonnegative and positive, respectively. The $n\times n$ identity matrix is denoted by $I_n$, the $n$-dimensional zero vector by $0_n$ and the $n\times m$ matrix with zero entries by $0_{n\times m}$. 
Given a matrix $A\in\mathbb{R}^{n\times m}$, we use $\|A\| \coloneqq \sqrt{\lambda_{\max}(A^\top A)}$, where $\lambda_{\max}(\cdot)$ is the maximum eigenvalue of a matrix.  The vector connecting the origins of coordinate frames $\{A\}$ and $\{B$\} expressed in frame $\{C\}$ coordinates in $3$-D space is denoted as $p^{\scriptscriptstyle C}_{{\scriptscriptstyle B/A}}\in{\mathbb{R}}^{3}$. Given $a\in\mathbb{R}^3$, $S(a)$ is the skew-symmetric matrix
defined according to $S(a)b = a\times b$. The rotation matrix from $\{A\}$ to $\{B\}$ is denoted as $R_{{\scriptscriptstyle B/A}}\in SO(3)$, where $SO(3)$ is the $3$-D rotation group.
The angular velocity of frame $\{B\}$ with respect to $\{A\}$ is
denoted as $\omega_{{\scriptscriptstyle B/A}}\in \mathbb{R}^{3}$ and it holds that \cite{siciliano2010robotics} $\dot{R}_{{\scriptscriptstyle B/A}}=S(\omega_{{\scriptscriptstyle B/A}})R_{{\scriptscriptstyle B/A}}$. We further denote as $\eta_{\scriptscriptstyle A/B}\in\mathbb{T}$ the Euler angles representing the orientation of $\left\{B\right\}$ with respect to $\left\{A \right\}$, with $\mathbb{T} \coloneqq (-\pi,\pi)\times(-\frac{\pi}{2},\frac{\pi}{2})\times(-\pi,\pi)$. We also define the set $\mathbb{M} \coloneqq \mathbb{R}^3\times\mathbb{T}$. In addition, $ S^n$ denotes the $(n+1)$-dimensional sphere. 
For notational brevity, when a coordinate frame corresponds to an inertial frame of reference $\left\{I\right\}$, we will omit its explicit notation (e.g., $p_{\scriptscriptstyle B} = p^{\scriptscriptstyle I}_{\scriptscriptstyle B/I}, \omega_{\scriptscriptstyle B} = \omega^{\scriptscriptstyle I}_{\scriptscriptstyle B/I}, R_{\scriptscriptstyle B} = R_{\scriptscriptstyle B/I}$ etc.). Finally, all vector and matrix differentiations are expressed with respect to an inertial frame $\{I\}$, unless otherwise stated.

\subsection{Unit Quaternions} \label{subsec:Quaternions}
Given two frames $\{A\}$ and $\{B\}$, we define a unit quaternion $\zeta_{\scriptscriptstyle B/A}\coloneqq[\varphi_{\scriptscriptstyle B/A}, \epsilon^\top _{\scriptscriptstyle B/A}]^\top \in S^{3}$ describing the orientation of $\{B\}$ with respect to $\{A\}$, with $\varphi_{\scriptscriptstyle B/A}\in\mathbb{R}, \epsilon_{\scriptscriptstyle B/A}\in \mathbb{R}^3$, subject to the constraint $\varphi^2_{\scriptscriptstyle B/A} + \epsilon^\top _{\scriptscriptstyle B/A}\epsilon_{\scriptscriptstyle B/A} = 1$. The relation between $\zeta_{\scriptscriptstyle B/A}$ and the corresponding rotation matrix $R_{\scriptscriptstyle B/A}$ as well as the axis/angle representation can be found in \cite{siciliano2010robotics}.
For a given quaternion $\zeta_{\scriptscriptstyle B/A}=[
\varphi_{\scriptscriptstyle B/A}, \epsilon^\top _{\scriptscriptstyle B/A}]^\top \in S^3$, its conjugate, that corresponds to the orientation of $ \{A\}$ with respect to $\{B\}$, is \cite{siciliano2010robotics} $\zeta^{+}_{\scriptscriptstyle B/A} \coloneqq [\varphi_{\scriptscriptstyle B/A}, -\epsilon^\top _{\scriptscriptstyle B/A}]^\top  \in S^3$.
Moreover, given two quaternions $\zeta_i \coloneqq \zeta_{\scriptscriptstyle B_i/A_i}= [
\varphi_{\scriptscriptstyle B_i/A_i}, \epsilon_{\scriptscriptstyle B_i/A_i}^\top ]^\top , \forall i\in\{1,2\}$, the quaternion product is defined as \cite{siciliano2010robotics}
\begin{equation}
\zeta_1\otimes\zeta_2 \coloneqq \left[\begin{array}{c}
\varphi_1\varphi_2-\epsilon_1^\top \epsilon_2 \\
\varphi_1\epsilon_2+\varphi_2\epsilon_1+S(\epsilon_1)\epsilon_2
\end{array}\right]\in S^{3},  \label{eq:quat_prod}
\end{equation}
where $\varphi_i \coloneqq \varphi_{\scriptscriptstyle B_i/A_i}, \epsilon_i \coloneqq \epsilon_{\scriptscriptstyle B_i/A_i}, \forall i\in\{1,2\}$.

For a moving frame $\{B\}$ (with respect to $\{A\}$), the time derivative of the quaternion $\zeta_{\scriptscriptstyle B/A}=[\varphi_{\scriptscriptstyle B/A}, \epsilon^\top _{\scriptscriptstyle B/A}]^\top \in S^3$ is given by \cite{siciliano2010robotics}:
\begin{subequations} \label{eq:propagation rule}
\begin{equation}
\dot{\zeta}_{\scriptscriptstyle B/A}=\frac{1}{2}{E(\zeta_{\scriptscriptstyle B/A})\omega^{\scriptscriptstyle A}_{\scriptscriptstyle B/A}},\label{eq:propagation rule_1}
\end{equation}
where $E:S^3\to\mathbb{R}^{4\times3}$ is defined as:
\begin{equation}
E(\zeta) \coloneqq \left[\begin{array}{c}
-\epsilon^\top \\
\varphi I_3-S(\epsilon)
\end{array}\right], \forall \zeta=[\varphi,\epsilon^\top]^\top\in S^3. \notag
\end{equation}
Finally, it can be shown that $E(\zeta)^\top E(\zeta) = I_3, \forall \zeta\in S^3 $ and hence \eqref{eq:propagation rule_1} implies
\begin{equation}
\omega^{\scriptscriptstyle A}_{\scriptscriptstyle B/A} = 2E(\zeta_{\scriptscriptstyle B/A})^\top \dot{\zeta}_{\scriptscriptstyle B/A}. \label{eq:propagation rule_2}
\end{equation}
\end{subequations}

\subsection{Prescribed Performance} \label{subsec:PPC}
\label{subsec:ppc}
Prescribed performance control, recently proposed in \cite{bechlioulis2008robust}, describes the behavior where a tracking error $e:\mathbb{R}_{\geq 0}\rightarrow\mathbb{R}$ evolves strictly within a predefined region that is bounded by certain functions of time, achieving prescribed transient and steady state performance.
The mathematical expression of prescribed performance is given by the inequalities
$-\rho_L(t) < e(t) < \rho_U(t),\ \ \forall t\in\mathbb{R}_{\geq 0}$, 
where $\rho_L(t),\rho_U(t)$ are smooth and bounded decaying functions of time satisfying $\lim\limits_{t\rightarrow\infty}\rho_L(t) > 0$ and $\lim\limits_{t\rightarrow\infty}\rho_U(t) > 0$, called performance functions. Specifically, for the exponential performance functions $\rho_i(t) \coloneqq (\rho_{i, \scriptscriptstyle 0}-\rho_{i,\scriptscriptstyle \infty})\exp(-l_it)+\rho_{i,\scriptscriptstyle \infty}$, with $\rho_{i,\scriptscriptstyle 0},\rho_{i,\scriptscriptstyle \infty}, l_i\in\mathbb{R}_{>0}, i\in\{U,L\}$, appropriately chosen constants, the terms $\rho_{L,\scriptscriptstyle 0} \coloneqq\rho_L(0),\rho_{U,\scriptscriptstyle 0}\coloneqq\rho_U(0)$ are selected such that $\rho_{U,\scriptscriptstyle 0} > e(0) > \rho_{L,\scriptscriptstyle 0}$ and the terms $\rho_{L,\scriptscriptstyle \infty} \coloneqq\lim\limits_{t\rightarrow\infty}\rho_L(t),\rho_{U,\scriptscriptstyle \infty}\coloneqq\lim\limits_{t\rightarrow\infty}\rho_U(t)$ represent the maximum allowable size of the tracking error $e(t)$ at steady state, which may be set arbitrarily small to a value reflecting the resolution of the measurement device, thus achieving practical convergence of $e(t)$ to zero. Moreover, the decreasing rate of $\rho_L(t),\rho_U(t)$, which is affected by the constants $l_L, l_U$ in this case, introduces a lower bound on the required speed of convergence of $e(t)$. Therefore, the appropriate selection of the performance functions $\rho_L(t),\rho_U(t)$ imposes performance characteristics on the tracking error $e(t)$.

\subsection{Dynamical Systems} \label{subsec:Dynamical Systems}
Consider the initial value problem: 
\begin{equation}
\dot{\sigma} = H(\sigma,t), \sigma(0)\in\Omega, \label{eq:initial value pr}
\end{equation}
with $H:\Omega\times\mathbb{R}_{\geq 0}\rightarrow\mathbb{R}^n$ where $\Omega\subset\mathbb{R}^n$ is a non-empty open set.
\begin{definition}
\cite{sontag2013mathematical} A solution $\sigma(t)$ of the initial value problem \eqref{eq:initial value pr} is maximal if it has no proper right extension that is also a solution of \eqref{eq:initial value pr}.
\end{definition}
\begin{theorem} \label{Th:dynamical systems}
\cite{sontag2013mathematical} Consider problem \eqref{eq:initial value pr}. Assume that $H(\sigma,t)$ is: a) locally Lipschitz on $\sigma$ for almost all $t\in\mathbb{R}_{\geq 0}$, b) piecewise continuous on $t$ for each fixed $\sigma\in\Omega$ and c) locally integrable on $t$ for each fixed $\sigma\in\Omega$. Then, there exists a maximal solution $\sigma(t)$ of \eqref{eq:initial value pr} on $[0,t_{\max})$ with $t_{\max} > 0$ such that $\sigma(t)\in\Omega,\forall t\in[0,t_{\max})$.
\end{theorem}
\begin{proposition} \label{Prop:dynamical systems}
\cite{sontag2013mathematical} Assume that the hypotheses of Theorem \ref{Th:dynamical systems} hold. For a maximal solution $\sigma(t)$ on the time interval $[0,t_{\max})$ with $t_{\max} < \infty$ and for any compact set $\Omega'\subset \Omega$ there exists a time instant $t'\in[0,t_{\max})$ such that $\sigma(t')\notin\Omega'$.
\end{proposition}

\section{Problem Formulation} \label{sec: Problem Formulation}

\begin{figure}
\centering
\includegraphics[width = 0.3\textwidth]{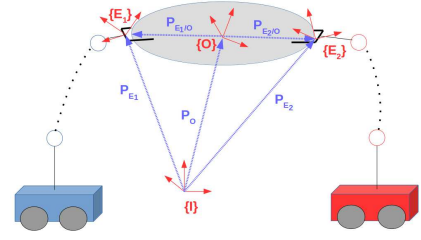}
\caption{Two robotic agents rigidly grasping an object.\label{fig:Two-robotic-arms}}
\end{figure}

Consider $N$ fully actuated robotic agents (e.g. robotic arms mounted on omnidirectional mobile bases) rigidly grasping an object  
(see Fig. \ref{fig:Two-robotic-arms}). We denote by $\left\{ E_{i}\right\}$, $\left\{ O\right\}$
the end-effector and
object's center of mass frames, respectively; $\left\{ I\right\} $
corresponds to an inertial frame of reference, as mentioned in Section \ref{subsec:Notation}. The rigidity assumption implies
that the agents can exert both forces and torques along all
directions to the object. In the following, we present the modeling of the coupled kinematics and dynamics of the object and the agents, which follows closely the one in \cite{zribi1992adaptive, liu1996decentralized}.

\subsection{Robotic Agents} \label{subsec:Robotics Agents}

We denote by $q_i,\dot{q}_i \in\mathbb{R}^{n_i}$, with $n_i\in\mathbb{N}, \forall i\in\mathcal{N}$, the generalized joint-space variables and their time derivatives of agent $i$, with $q_i \coloneqq [q_{i_1},\dots, q_{i_{n_i}}]$. Here, $q_i$ consists of the degrees of freedom of the robotic arm as well as the moving base. The overall joint configuration is then $q \coloneqq [q^\top _1,\dots,q^\top _N]^\top , \dot{q} \coloneqq [\dot{q}^\top _1,\dots,\dot{q}^\top _N]^\top \in\mathbb{R}^{n}$, with $n \coloneqq \sum_{i\in\mathcal{N}}n_i$. In addition, the inertial position and Euler-angle orientation of the $i$th end-effector, denoted by $p_{\scriptscriptstyle E_i}$ and $\eta_{\scriptscriptstyle E_i}$, respectively, can be derived by the forward kinematics and are smooth functions of $q_i$, i.e. $p_{E_i}:\mathbb{R}^{n_i}\to\mathbb{R}^3$, $\eta_{E_i}:\mathbb{R}^{n_i}\to\mathbb{T}$.  The generalized velocity of each agent's end-effector $v_i \coloneqq [\dot{p}^\top_{E_i},\omega^\top_{\scriptscriptstyle E_i}]^\top \in\mathbb{R}^6$, can be computed through the differential kinematics $v_i = J_i(q_i)\dot{q}_i$ \cite{siciliano2010robotics}, where $J_i:\mathbb{R}^{n_i}\to\mathbb{R}^{6\times n_i}$ is a smooth function representing the geometric Jacobian matrix, $\forall i\in\mathcal{N}$ \cite{siciliano2010robotics}. 
We define also the set $\mathbb{S}_i \coloneqq \{q_i\in\mathbb{R}^{n_i} : \det( J_i(q_i)J_i(q_i)^\top) > 0 \}$ which contains all the singularity-free configurations.
The differential equation describing the dynamics of each agent in task-space coordinates is \cite{siciliano2010robotics}:
\begin{equation}
M_i(q_i)\dot{v}_i+C_i(q_i,\dot{q}_i)v_i+ g_i(q_i) + d_i(q_i,\dot{q}_i,t) =u_i-f_i,\label{eq:manipulator dynamics}
\end{equation}
where $M_i:\mathbb{S}_i\to\mathbb{R}^{n_i\times n_i}$ is the positive definite inertia matrix, $C_i:\mathbb{S}_i\times\mathbb{R}^{n_i}\to\mathbb{R}^{n_i\times n_i}$ is the Coriolis matrix, $g_i:\mathbb{S}_i\to\mathbb{R}^{n_i}$ is the gravity term, $d_i:\mathbb{S}_i\times\mathbb{R}^{n_i}\times\mathbb{R}_{\geq 0}\to\mathbb{R}^{n_i}$ is a vector representing unmodeled friction, uncertainties and external disturbances, $f_i\in\mathbb{R}^{6}$ is the vector of generalized forces that agent $i$ exerts on the grasping point with the object and $u_i\in\mathbb{R}^6$ is the task space wrench, that acts as the control input; $u_i$ is related to the input torques, denoted by $\tau_i$, via $\tau_i = J^\top _i(q_i)u_i + (I_{n_i} - J^\top _i(q_i)[J^{+}_i(q_i)]^\top)\tau_{i0}$, where $J^{+}_i$ is a generalized inverse of $J_i$ \cite{siciliano2010robotics}. Moreover, $\tau_{i0}$ concerns redundant agents ($n_i > 6$) and does not contribute to end-effector forces. 
The agent task-space dynamics \eqref{eq:manipulator dynamics} can be written in vector form as:
\begin{equation}
{M}({q})\dot{v}+{C}({q},\dot{{q}})v + {g}({q}) + d(q,\dot{q},t) = {u}-{f},\label{eq:manipulator dynamics_vector_form}
\end{equation}
where $v \coloneqq [v^\top _1,\dots,v^\top _N]\in\mathbb{R}^{6N}$, $M \coloneqq \text{diag}\{[M_i]_{i\in\mathcal{N}}\}\in\mathbb{R}^{6N\times6N}$, ${C} \coloneqq \text{diag}\{[C_i]_{i\in\mathcal{N}}\}\in\mathbb{R}^{6N\times6N}$, $f \coloneqq [f^\top _1,\dots,f^\top _N]^\top $, $u \coloneqq [u^\top _1,\dots,u^\top _N]^\top $, $g \coloneqq [g^\top _1,\dots,g^\top _N]^\top $, $d \coloneqq [d^\top _1,\dots,d^\top _N]^\top \in\mathbb{R}^{6N}$.



\subsection{Object} \label{subsec:Object}

Regarding the object, we denote by $x_{\scriptscriptstyle O}\coloneqq [p^\top _{\scriptscriptstyle O},\eta^\top _{\scriptscriptstyle O}]^\top \in\mathbb{M}$, $v_{\scriptscriptstyle O} \coloneqq [\dot{p}^\top_{\scriptscriptstyle O}, \omega^\top _{\scriptscriptstyle O}]^\top \in\mathbb{R}^{12}$ the pose and generalized velocity of its center of mass, with $\eta_{\scr O}\coloneqq [\phi_{\scr O}, \theta_{\scr O}, \psi_{\scr O}]^\top$. We consider the following second order dynamics, which can be derived based on the Newton-Euler formulation: 
\begin{subequations} \label{eq:object dynamics}
\begin{align}
& \dot{x}_{\scriptscriptstyle O} = J_{\scriptscriptstyle O}(\eta_{\scriptscriptstyle O})v_{\scriptscriptstyle O}, \label{eq:object dynamics 1}\\
& M_{\scriptscriptstyle O}(x_{\scriptscriptstyle O})\dot{v}_{\scriptscriptstyle O}+C_{\scriptscriptstyle O}(x_{\scriptscriptstyle O},\dot{x}_{\scriptscriptstyle O})v_{\scriptscriptstyle O}+g_{\scriptscriptstyle O}(x_{\scriptscriptstyle O}) + d_{\scriptscriptstyle O}(x_{\scriptscriptstyle O},\dot{x}_{\scriptscriptstyle O},t) = f_{\scriptscriptstyle O}, \label{eq:object dynamics 2}
\end{align}
\end{subequations}
where $M_{\scriptscriptstyle O}:\mathbb{M}\to\mathbb{R}^{6\times6}$ is the positive definite inertia matrix, $C_{\scriptscriptstyle O}:\mathbb{M}\times\mathbb{R}^{6}\to\mathbb{R}^{6\times6}$ is the Coriolis matrix, $g_{\scriptscriptstyle O}:\mathbb{M}\to\mathbb{R}^6$ is the gravity vector, $d_{\scriptscriptstyle O}:\mathbb{M}\times\mathbb{R}^6\times\mathbb{R}_{\geq 0}\to\mathbb{R}^6$ a vector representing modeling uncertainties and external disturbances, and $f_{\scriptscriptstyle O}\in\mathbb{R}^6$ is the vector of generalized forces acting on the object's center of mass. Moreover, $J_{\scriptscriptstyle O}:\mathbb{T}\to\mathbb{R}^{6\times6}$ is the well-known object representation Jacobian 
and is not well-defined when $\theta_{\scriptscriptstyle O}= \pm \tfrac{\pi}{2}$, which is referred to as \textit{representation singularity}. 
A way to avoid the aforementioned singularity is to transform the Euler angles to a unit quaternion for the orientation. Hence,  $\eta_{\scriptscriptstyle O}$ can be transformed to the unit quaternion $\zeta_{\scriptscriptstyle O}=[\varphi_{\scriptscriptstyle O}, \epsilon^\top _{\scriptscriptstyle O}]^\top \in S^3$ \cite{siciliano2010robotics}, for which, following Section \ref{subsec:Quaternions} and \eqref{eq:propagation rule}, one obtains ${\dot{\zeta}_{\scriptscriptstyle O}} = \frac{1}{2}E(\zeta_{\scriptscriptstyle O})\omega_{\scriptscriptstyle O}$ and $\omega_{\scriptscriptstyle O} = 2[E(\zeta_{\scriptscriptstyle O})]^\top\dot{\zeta}_{\scriptscriptstyle O}$,
Moreover, it can be proved that 
\begin{subequations} \label{eq:J_norm}
	\begin{align}
	&\lVert J_{\scriptscriptstyle O}(\eta_{\scriptscriptstyle O})\rVert  = \sqrt{\tfrac{\lvert \sin(\theta_{\scriptscriptstyle O}) \rvert + 1}{1 - \sin^2(\theta_{\scriptscriptstyle O})}}, \\ 
	&\| J_{\scr O}(\eta_{\scr O})^{-1} \| = \sqrt{1 + \sin(\theta_{\scr O})} \leq \sqrt{2},
	\end{align}
\end{subequations}
where $J_{\scr O}(\cdot)^{-1}$ is the matrix inverse.
which constitutes a singularity-free representation.  

\subsection{Coupled Dynamics}

In view of Fig. \ref{fig:Two-robotic-arms}, one concludes that the pose of the agents and the object's center of mass are related as  
\begin{subequations} \label{eq:coupled_kinematics}
\begin{align}
p_{\scriptscriptstyle E_i}(q_i) &= p_{\scriptscriptstyle O}+p_{\scriptscriptstyle E_i/O}(q_i) =p_{\scriptscriptstyle O} +  R_{\scriptscriptstyle E_{i}}(q_i) p^{\scriptscriptstyle E_i}_{\scriptscriptstyle E_{i}/O},\label{eq:coupled_kinematics_1} \\
\eta_{\scriptscriptstyle E_i}(q_i) &= \eta_{\scriptscriptstyle O} + \eta_{\scriptscriptstyle E_i/O}, \label{eq:coupled_kinematics_2}
\end{align}
\end{subequations}
$\forall i\in\mathcal{N}$, where $p^{\scriptscriptstyle E_i}_{\scriptscriptstyle E_i/O}$ and $\eta_{\scriptscriptstyle E_i/O}$ are the \textit{constant} distance and orientation offset vectors between $\{O\}$ and $\{E_i\}$. 
Following \eqref{eq:coupled_kinematics}, along with the
fact that, due to the grasping rigidity, it holds that $\omega_{\scriptscriptstyle E_i}=\omega_{\scriptscriptstyle O}, \forall i\in\mathcal{N}$,
one obtains
\begin{equation}
v_i=J_{\scriptscriptstyle O_i}(q_i) v_{\scriptscriptstyle O}, \label{eq:J_o_i}
\end{equation}
where $J_{\scriptscriptstyle O_i}:\mathbb{R}^{n_i}\to\mathbb{R}^{6\times6}$ is the object-to-agent Jacobian matrix \cite{Verginis_ifac} 
for which it can be further proved that
\begin{equation}
\| J_{\scriptscriptstyle O_i}(x) \| \leq \Big\| p^{\scriptscriptstyle E_i}_{\scriptscriptstyle O/E_i} \Big\| + 1, \forall  x\in\mathbb{R}^{n_i}, i\in\mathcal{N}. \label{eq:J_O_i bound}
\end{equation}


The kineto-statics duality along with the grasp rigidity suggest that the force $f_{\scriptscriptstyle O}$ acting on the object's center of mass and the generalized forces $f_i,i\in\mathcal{N}$, exerted by the agents at the grasping points, are related through $f_{\scriptscriptstyle O}=[G(q)]^\top{f}$,
where $G:\mathbb{R}^{n}\to\mathbb{R}^{6N\times 6}$, with $G({q}) \coloneqq [[J_{\scriptscriptstyle O_1}(q_1)]^\top,\dots,[J_{\scriptscriptstyle O_N}(q_N)]^\top]^\top$, 
is the full column-rank grasp matrix. 
By using the latter along with \eqref{eq:manipulator dynamics_vector_form}, \eqref{eq:object dynamics}, \eqref{eq:J_o_i} and its derivative,
  we obtain the overall system coupled dynamics: 
\begin{equation}
\widetilde{M}(x)\dot{v}_{\scriptscriptstyle O}+\widetilde{C}(x)v_{\scriptscriptstyle O}+\widetilde{g}(x)+\widetilde{d}(x,t)  = [G(q)]^\top{u},\label{eq:coupled dynamics}
\end{equation}
where $\widetilde{M}  \coloneqq  M_{\scriptscriptstyle O} + G^\top M G$, $\widetilde{C}  \coloneqq C_{\scriptscriptstyle O}+ G^\top C G \notag + G^\top M\dot{G}$, $\widetilde{g}  \coloneqq  g_{\scriptscriptstyle O}+[G(q)]^\top g(q)$,  $\widetilde{d}  \coloneqq d_{\scriptscriptstyle O}+ G^\top d$, 
$x$ is the overall state $x\coloneqq [q^\top,\dot{q}^\top,x^\top_{\scriptscriptstyle O},\dot{x}^\top_{\scriptscriptstyle O}]^\top\in\mathbb{S}\times\mathbb{R}^{n+6}\times\mathbb{M}$, $\mathbb{S}\coloneqq \mathbb{S}_1\times\dots\times\mathbb{S}_N$, and we have omitted the arguments for notational brevity.
Moreover, the following Lemma, whose proof can be found in \cite{Verginis_ifac}, is necessary for the following analysis. 
\begin{lemma} \label{lem:coupled dynamics skew symmetry}
The matrix $\widetilde{M}(x)$ is symmetric and positive definite and the matrix $\dot{\widetilde{M}}(x) - 2\widetilde{C}(x)$ is skew symmetric. 
\end{lemma}

\noindent The positive definiteness of $\widetilde{M}(x)$ implies $\underline{m}I_6 \leq \widetilde{M}(x) \leq \bar{m} I_6$,
$\forall x\in \mathbb{S}\times\mathbb{R}^{n+6}\times\mathbb{M}$, where $\underline{m}$ and $\bar{m}$ are positive unknown constants. 

We are now ready to state the problem treated in this paper:

\begin{problem}
Given a desired bounded object smooth pose trajectory specified by $x_{\text{d}}(t) \coloneqq [p_{\text{d}}(t)^\top, \eta_{\text{d}}(t)^\top]^\top$, $p_\text{d}(t)\in\mathbb{R}^3,
\eta_\text{d}(t) \coloneqq [\varphi_\text{d}(t),\theta_\text{d}(t), \psi_\text{d}(t)]\in\mathbb{T}$, with bounded first
and second derivatives, and $\theta_\text{d}(t)\in[-\bar{\theta},\bar{\theta}]\subset(-\tfrac{\pi}{2},\tfrac{\pi}{2}), \forall t\in\mathbb{R}_{\geq 0}$, as well as $v_{\scriptscriptstyle O}(0)=0_{6}$, determine a decentralized control law $u$ in \eqref{eq:coupled dynamics} such that one of the following holds:\\
1) ${\lim\limits_{t\rightarrow\infty}\left[
[p_{\scriptscriptstyle O}(t) - p_\text{d}(t)]^\top, 
[\eta_{\scriptscriptstyle O}(t) - \eta_\text{d}(t)]^\top
\right]^\top= 0_{3}}, \notag$ \\
2) {$\|\left[
	[p_{\scriptscriptstyle O}(t) - p_\text{d}(t)]^\top, 
	[\eta_{\scriptscriptstyle O}(t) - \eta_\text{d}(t)]^\top
	\right] \| \leq \lambda \exp(-lt) + \rho$, $\forall t\in\mathbb{R}_{\geq 0}$, for positive $\lambda, l, \rho$}.
\label{prob:problem1} 
\end{problem}
{Part $1$ in the aforementioned problem statement corresponds to the asymptotic stability that will be guaranteed by the control scheme of Section \ref{subsec:Quaternion Controller}, and part $2$ is associated with the predefined transient and steady state performance that will be guaranteed in Section \ref{subsec:PPC Controller}.}
 The requirement $\theta_\text{d}(t)\in[-\bar{\theta},\bar{\theta}]\subset(-\tfrac{\pi}{2},\tfrac{\pi}{2}), \forall t\in\mathbb{R}_{\geq 0}$ is a necessary condition needed to ensure that tracking of $\theta_\text{d}$ will not result in singular configurations of $J_{\scriptscriptstyle O}(\eta_{\scriptscriptstyle O})$, which is needed for the control protocol of Section \ref{subsec:PPC Controller}. The constant $\bar{\theta}\in[0,\tfrac{\pi}{2})$ can be taken arbitrarily close to $\tfrac{\pi}{2}$.

To solve the aforementioned problem, we need the following assumptions regarding the agent feedback, the bounds of the uncertainties/disturbances, and the kinematic singularities.

\begin{ass} [Feedback] \label{ass:feedback}
Each agent $i\in\mathcal{N}$ has continuous feedback of its own state $q_i, \dot{q}_i$.
\end{ass}
\begin{ass} [Object Geometry] \label{ass:object geometry}
Each agent $i\in\mathcal{N}$ knows the constant offsets $p^{\scriptscriptstyle E_i}_{\scriptscriptstyle E_i/O}$ and $\eta_{\scriptscriptstyle E_i/O}, \forall i\in\mathcal{N}$.
\end{ass}
\begin{ass} [Kinematic Singularities] \label{ass:kinematic singularities}
The agents operate away from kinematic singularities, i.e., $q_i(t)$ evolves in a closed subset of $\mathbb{S}_i$, $\forall i\in\mathcal{N}$.
\end{ass}

Assumption \ref{ass:feedback} is realistic for real manipulation systems, since on-board sensor can provide accurately the measurements $q_i,\dot{q}_i$. The object geometric characteristics in Assumption \ref{ass:object geometry} can be obtained by on-board sensors, whose inaccuracies are not modeled in this work. Finally, Assumption \ref{ass:kinematic singularities} states that the $q_i$ that achieve $x_{\scriptscriptstyle O}(t) = x_{\text{d}}(t), \forall t\in\mathbb{R}_{\geq 0}$ are sufficiently far from singular configurations.
Since each agent has feedback from its state $q_i, \dot{q}_i$, it can compute through the forward and differential kinematics the end-effector pose $p_{\scriptscriptstyle E_i}(q_i), \eta_{\scriptscriptstyle E_i}(q_i)$ and the velocity $v_i$, $\forall i\in\mathcal{N}$. Moreover, since it knows $p^{\scriptscriptstyle E_i}_{\scriptscriptstyle E_i/O}$ and $\eta_{\scriptscriptstyle E_i/O}$, it can compute $J_{\scriptscriptstyle O_i}(q_i)$ and $x_{\scriptscriptstyle O}$, $v_{\scriptscriptstyle O}$ by inverting \eqref{eq:coupled_kinematics} and \eqref{eq:J_o_i}, respectively. Consequently, each agent can then compute the quaternion signals $\zeta_{\scriptscriptstyle O}$ and $\dot{\zeta}_{\scriptscriptstyle O}$. 

{Note that, due to Assumption \ref{ass:object geometry} and the grasp rigidity, the object-agents configuration is similar to a single closed-chain robot. The considered multi-agent setup, however, renders the problem more challenging, since the agents must calculate their own control signal in a decentralized manner, without communicating with each other. Moreover, each agent needs to compensate its own part of the (possibly uncertain/unknown) dynamics of the coupled dynamic equation \eqref{eq:coupled dynamics}, while respecting the rigidity kinematic constraints. 
Regarding Assumption \ref{ass:object geometry}, our future directions include its relaxation to uncertain/unknown object offsets for some agents, which would then not have exact feedback of the object's pose. In that case, the team would need to cooperate in a leader-follower fashion for the compensation/estimation of the state by these agents.}

\section{Main Results} \label{sec:Main Results}
In this section we present two control schemes for the solution of Problem \ref{prob:problem1}. The proposed controllers are decentralized, in the sense that the agents calculate their control signal on their own, without communicating with each other, as well as robust, since they do not take into account the dynamic properties of the agents or the object (mass/inertia moments) or the uncertainties/external disturbances modeled by the function $\widetilde{d}(x,t)$ in \eqref{eq:coupled dynamics}. The first control scheme is presented in Section \ref{subsec:Quaternion Controller}, and is based on quaternion feedback and adaptation laws, while the second control scheme is given in Section \ref{subsec:PPC Controller} and is inspired by the Prescribed Performance Control (PPC) Methodology introduced in \cite{bechlioulis2008robust}. 

\subsection{Adaptive Control with Quaternion Feedback} \label{subsec:Quaternion Controller}
{The proposed controller of this section is based on the techniques of adaptive control, whose aim is the design of control systems that are robust to constant or slowly varying  unknown parameters. For more details, we refer the reviewer to the related literature (e.g., \cite{lavretsky13adaptive} and the references therein).}

Firstly, we need the following assumption regarding the model uncertainties/external disturbances.
\begin{ass} [Uncertainties/Disturbance parameterization] \label{ass:disturbance bound}
There exist constant \textit{unknown} vectors $\bar{d}_{\scriptscriptstyle O}\in\mathbb{R}^{\mu_{\scr O}}, \bar{d}_i\in\mathbb{R}^{\mu}$ and known functions $\delta_{\scriptscriptstyle O}:\mathbb{M}\times\mathbb{R}^6\times\mathbb{R}_{\geq 0}\to\mathbb{R}^{6\times\mu_{\scr O}}, \delta_i:\mathbb{R}^{2n_i}\times\mathbb{R}_{\geq 0}\to\mathbb{R}^{6\times\mu}$, such that $d_{\scriptscriptstyle O}(x_{\scriptscriptstyle O},\dot{x}_{\scriptscriptstyle O},t) = \delta_{\scriptscriptstyle O}(x_{\scriptscriptstyle O},\dot{x}_{\scriptscriptstyle O},t)\bar{d}_{\scriptscriptstyle O}$, 
$d_i(q_i,\dot{q}_i,t) = \delta_i(q_i,\dot{q}_i,t)\bar{d}_i$,
$\forall q_i,\dot{q}_i\in\mathbb{R}^{n_i}, x_{\scriptscriptstyle O}\in\mathbb{M}, \dot{x}_{\scriptscriptstyle O}\in\mathbb{R}^6, t\in\mathbb{R}_{\geq 0}, i\in\mathcal{N}$,
{where $\delta_{\scriptscriptstyle  O}(x_{\scriptscriptstyle  O},\dot{x}_{\scriptscriptstyle  O},t)$ and  $\delta_i(q_i,\dot{q}_i,t)$ are continuous in $(x_{\scr O},\dot{x}_{\scr O})$ and $(q_i,\dot{q}_i)$, respectively, and bounded in $t$.}
\end{ass} 
\noindent {The aforementioned assumption is motivated by the use of Neural Networks for approximating unknown functions in compact sets \cite{lavretsky13adaptive}. More specifically, any continuous function $f(x):\mathbb{R}^n \to \mathbb{R}^m$ can be approximated on a known compact set $X\subset \mathbb{R}^n$ by a Neural Network equipped with $N$ Radial Basis Functions (RBFs) $\Phi(x)$ and using unknown ideal constant connection weights that are stored in a matrix $\Theta \in\mathbb{R}^{N\times m}$ as $f(x) = \Theta ^\top \Phi(x) + \varepsilon(x)$;
$\Theta ^\top \Phi(x)$ represents the parametric uncertainty and $\varepsilon(x)$ represents the unknown nonparametric uncertainty, which is bounded as $\|\varepsilon(x)\| \leq  \bar{\varepsilon}$ in $X$. In our case, the functions $\delta_{\scriptscriptstyle O}$, $\delta_i$ play the role of the known function $\Phi(x)$ and $\bar{d}_{\scriptscriptstyle O}$, $\bar{d}_i$ and $\mu$, $\mu_{\scriptscriptstyle O}$ represent the unknown constants $\Theta$ and the number of layers of the Neural Network, respectively. 
Nevertheless, in view of Neural Network approximation, Assumption $4$ implies that the nonparametric uncertainty is zero and that $d_{\scr O}$ and $d_i$ are \textit{known} functions of \textit{time}. These properties can be relaxed with non-zero bounded nonparametric uncertainties and \textit{unknown} but bounded time-dependent disturbances, i.e. $d_i(q_i,\dot{q}_i,t) = \delta_{i,q}(q_i,\dot{q}_i)\bar{d}_i + d_{i,t}(t) + \varepsilon_{i,q}(q_i,\dot{q}_i)$ and  
$d_{\scriptscriptstyle O}(x_{\scriptscriptstyle O},\dot{x}_{\scriptscriptstyle O},t) = \delta_{\scriptscriptstyle O, x}(x_{\scriptscriptstyle O},\dot{x}_{\scriptscriptstyle O})\bar{d}_{\scr O} + d_{\scriptscriptstyle O, t}(t) + \varepsilon_{\scriptscriptstyle O, x}(x_{\scriptscriptstyle O},\dot{x}_{\scriptscriptstyle O})$, where $d_{i,t}, d_{\scr O,t}$, $\varepsilon_{i,q}, \varepsilon_{\scr O,x}$ are bounded. In that case, instead of asymptotic convergence of the pose to the desired one, we can show convergence of the respective errors to a compact set around the origin. For more details on Neural Network approximation and adaptive control with illustrative examples, we refer the reader to \cite[Ch. 12]{lavretsky13adaptive}.}


%

The desired Euler angle orientation vector $\eta_\text{d}:\mathbb{R}_{\geq 0}\to\mathbb{T}$ is transformed first to the unit quaternion $\zeta_\text{d}:\mathbb{R}_{\geq 0}\to S^3$ \cite{siciliano2010robotics} 
and we define the position error $e_p  \coloneqq p_{\scriptscriptstyle O}-p_\text{d}$.
Since unit quaternions do not form a vector space, they cannot be subtracted to form an orientation error; instead we should use the properties of the quaternion group algebra. Let $e_\zeta = [e_{\varphi}, e^\top _{\epsilon}]^\top \in S^3$ be the unit quaternion describing the orientation error. Then, it holds that \cite{siciliano2010robotics}
\begin{equation}
e_{\zeta}=\zeta_\text{d}\otimes\zeta_{\scriptscriptstyle O}^{+} = 
\begin{bmatrix}
\varphi_\text{d} \\ \epsilon_\text{d}
\end{bmatrix} \otimes
\begin{bmatrix}
\varphi_{\scriptscriptstyle O} \\ -\epsilon_{\scriptscriptstyle O}
\end{bmatrix}, \notag
\end{equation}
which, by using \eqref{eq:quat_prod}, becomes:
\begin{align}
e_\zeta = \begin{bmatrix}
e_\varphi\\ e_\epsilon 
\end{bmatrix} \coloneqq
\begin{bmatrix}
\varphi_{\scriptscriptstyle O}\varphi_\text{d}+\epsilon^\top _{\scriptscriptstyle O}\epsilon_\text{d} \\
\varphi_{\scriptscriptstyle O}\epsilon_\text{d}-\varphi_\text{d}\epsilon_{\scriptscriptstyle O}+S(\epsilon_{\scriptscriptstyle O})\epsilon_\text{d} \label{eq:quat_error}
\end{bmatrix}.
\end{align}
By employing \eqref{eq:propagation rule} and certain properties of skew-symmetric matrices \cite{campa2006kinematic}, the dynamics of $e_p, e_\varphi$ can {be} shown to be:
\begin{subequations} \label{eq:error_dynamics}
\begin{align}
\dot{e}_{p}  = & \dot{p}_{\scriptscriptstyle O}-\dot{p}_\text{d} \label{eq:position error dynamics}\\
\dot{e}_{\varphi}  = & \tfrac{1}{2}e^\top _{\epsilon}e_\omega \label{eq:eta_error dynamics}  \\
\dot{e}_{\epsilon} = & -\tfrac{1}{2}\left[e_\varphi I_3 + S(e_{\epsilon}) \right]e_\omega - S(e_{\epsilon})\omega_\text{d},\label{eq:epsilon error dynamics} 
\end{align}
\end{subequations}
where $e_\omega \coloneqq \omega_{\scriptscriptstyle O} - \omega_\text{d}$ is the angular velocity error, with $\omega_\text{d} = 2E (\zeta_\text{d})^\top \dot{\zeta}_\text{d}$, as indicated by \eqref{eq:propagation rule_2}. 
Due to the ambiguity of unit quaternions, when $\zeta_{\scriptscriptstyle O} = \zeta_\text{d}$, then $e_{\zeta} = [1, 0^\top_3]^\top \in S^3$. If $\zeta_{\scriptscriptstyle O} = -\zeta_\text{d}$, then $e_{\zeta} = [-1, 0^\top_3]^\top \in S^3$, which, however, represents the same orientation. Therefore,  the control objective is equivalent to 
$\lim\limits_{t\to\infty} 
\begin{bmatrix}
e_p(t)^\top,  \lvert e_\varphi(t) \rvert, e_\epsilon(t)^\top
\end{bmatrix}^\top = 
\begin{bmatrix}
0_3^\top, 1, 0_3^\top
\end{bmatrix}^\top$. 

The left hand side of \eqref{eq:manipulator dynamics}, after employing \eqref{eq:J_o_i} and its derivative, becomes
\begin{align}
& M_i(q_i)\dot{v}_i + C_i(q_i,\dot{q}_i)v_i + g_i(q_i) + d_i(q_i,\dot{q}_i,t) = \notag \\
& M_i(q_i)\Big(J_{\scriptscriptstyle O_i}(q_i)\dot{v}_{\scriptscriptstyle O} + \dot{J}_{\scriptscriptstyle O_i}(q_i)v_{\scriptscriptstyle O}\Big)  + C_i(q_i,\dot{q}_i)J_{\scriptscriptstyle O_i}(q_i)v_{\scriptscriptstyle O} + \notag \\
&\hspace{10mm} g_i(q_i) + d_i(q_i,\dot{q}_i,t). \notag
\end{align}
which, according to Assumption \ref{ass:disturbance bound} and the fact that the manipulator dynamics can be linearly parameterized with respect to dynamic parameters \cite{Slotine_adaptive87}, becomes
\begin{align*}
& M_i(q_i)J_{\scriptscriptstyle O_i}(q_i)\dot{v}_{\scriptscriptstyle O} + \Big(M_i(q_i)\dot{J}_{\scriptscriptstyle O_i}(q_i)+C_i(q_i,\dot{q}_i)J_{\scriptscriptstyle O_i}(q_i)\Big)v_{\scriptscriptstyle O} \notag \\
&+ g_i(q_i) + d_i(q_i,\dot{q}_i,t) =  Y_i(q_i,\dot{q}_i,v_{\scriptscriptstyle O}, \dot{v}_{\scriptscriptstyle O})\vartheta_i + \delta_i(q_i,\dot{q}_i,t)\bar{d}_i, 
\end{align*}
$\forall i\in\mathcal{N}$, where $\vartheta_i \in\mathbb{R}^\ell,\ell\in\mathbb{N}$, are vectors of unknown but constant dynamic parameters of the agents, appearing in the terms $M_i,C_i,g_i$, and $Y_i:\mathbb{S}\times \mathbb{R}^{n_i+12}\to\mathbb{R}^{6\times\ell}$ are known regressor matrices, independent of $\vartheta_i,i\in\mathcal{N}$. Without loss of generality, we assume here {that $\ell$ is the same} for all agents. Similarly, the dynamical terms of the left hand side of \eqref{eq:object dynamics 2} can be written as 
\begin{align*}
& M_{\scriptscriptstyle O}(x_{\scriptscriptstyle O})\dot{v}_{\scriptscriptstyle O} + C_{\scriptscriptstyle O}(x_{\scriptscriptstyle O}, \dot{x}_{\scriptscriptstyle O})v_{\scriptscriptstyle O} + g_{\scriptscriptstyle O}(x_{\scriptscriptstyle O})  + 
 d_{\scriptscriptstyle O}(x_{\scriptscriptstyle O},\dot{x}_{\scriptscriptstyle O}, t) \notag \\ 
&\hspace{5mm} =  Y_{\scriptscriptstyle O}(x_{\scriptscriptstyle O},\dot{x}_{\scriptscriptstyle O},v_{\scriptscriptstyle O}, \dot{v}_{\scriptscriptstyle O})\vartheta_{\scriptscriptstyle O} + \delta_{\scriptscriptstyle O}(x_{\scriptscriptstyle O},\dot{x}_{\scriptscriptstyle O}, t)\bar{d}_{\scriptscriptstyle O},
\end{align*}
where $\vartheta_{\scriptscriptstyle O}\in\mathbb{R}^{\ell_{\scriptscriptstyle O}}, \ell_{\scriptscriptstyle O}\in\mathbb{N}$ is a vector of unknown but constant dynamic parameters of the object, appearing in the terms $M_{\scriptscriptstyle O},C_{\scriptscriptstyle O},g_{\scriptscriptstyle O}$, and 
$Y_{\scriptscriptstyle O}:\mathbb{M}\times\mathbb{R}^{18}\to\mathbb{R}^{6\times\ell_{\scriptscriptstyle O}}$ is a known regressor matrix, independent of $\vartheta_{\scriptscriptstyle O}$. It is worth noting that
the choice for $\ell$ and $\ell_{\scriptscriptstyle O}$ is not unique. 
In view of the aforementioned expressions, the left-hand side of \eqref{eq:coupled dynamics} can be written as: 	

\begin{align}
& \widetilde{M}(x)\dot{v}_{\scriptscriptstyle O}+ \widetilde{C}(x)v_{\scriptscriptstyle O}+\widetilde{g}(x) + \widetilde{d}(x,t) = Y_{\scriptscriptstyle O}(x_{\scriptscriptstyle O},\dot{x}_{\scriptscriptstyle O},v_{\scriptscriptstyle O},\dot{v}_{\scriptscriptstyle O})\vartheta_{\scriptscriptstyle O} \notag \\ 
&+ \delta_{\scriptscriptstyle O}(x_{\scriptscriptstyle O},\dot{x}_{\scriptscriptstyle O},t)\bar{d}_{\scriptscriptstyle O} +  [G(q)]^\top\Big(\widetilde{Y}(q,\dot{q},v_{\scriptscriptstyle O},\dot{v}_{\scriptscriptstyle O})\vartheta  + \widetilde{\delta}(q,\dot{q},t)\bar{d}\Big), \label{eq:parameter linearity 2}
\end{align}
where $\widetilde{Y}(q,\dot{q},v_{\scriptscriptstyle O}, \dot{v}_{\scriptscriptstyle O}) \coloneqq \text{diag}\{[Y_i(q_i,\dot{q}_i,v_{\scriptscriptstyle O},\dot{v}_{\scriptscriptstyle O})]_{i\in\mathcal{N}} \} \in\mathbb{R}^{6N \times Nl}$, $\vartheta \coloneqq [\vartheta^\top_1,\dots,\vartheta^\top_N]^\top \in\mathbb{R}^{N\ell}$, $\bar{d}$ $\coloneqq$ $[\bar{d}^\top_1,\dots,\bar{d}^\top_N]^\top\in\mathbb{R}^{N\mu}$, and $\widetilde{\delta}(q,\dot{q},t) \coloneqq \text{diag}\{[\delta_i(q_i,\dot{q}_i,t)]_{i\in\mathcal{N}}\}$ $\in$ $\mathbb{R}^{6N\times N\mu}$.

Let us now introduce the states $\hat{\vartheta}_{\scriptscriptstyle O}\in\mathbb{R}^{\ell_{\scriptscriptstyle O}}$ and $\hat{\vartheta}_i\in\mathbb{R}^{\ell}$ which represent the estimates of $\vartheta_{\scriptscriptstyle O}$ and $\vartheta_i$, respectively, by agent $i\in\mathcal{N}$, and the corresponding stack vector
$\hat{\vartheta} \coloneqq [\hat{\vartheta}_1^\top,\dots,\hat{\vartheta}_N^\top]^\top \in\mathbb{R}^{N\ell}$,
for which the associated errors are 
\begin{subequations} \label{eq:adaptation errors}
\begin{align}
e_{\vartheta_{\scriptscriptstyle O}} \coloneqq &
 \vartheta_{\scriptscriptstyle O} - \hat{\vartheta}_{\scriptscriptstyle O}  \in\mathbb{R}^{\ell_{\scr O}} \label{eq:adaptation errors_1}\\
e_{\vartheta} \coloneqq & \begin{bmatrix}
e_{\vartheta_1}^\top, \dots, e_{\vartheta_N}^\top
\end{bmatrix}^\top  \coloneqq 
\vartheta- \hat{\vartheta}\in\mathbb{R}^{N\ell}.  \label{eq:adaptation errors_2}
\end{align}
\end{subequations}
In the same vein, we introduce the states $\hat{d}_{\scriptscriptstyle O}\in\mathbb{R}^{\mu_{\scr O}}$ and $\hat{d}_i\in\mathbb{R}^{\mu}$ that correspond to the estimates of $\bar{d}_{\scriptscriptstyle O}$ and $\bar{d}_i$, respectively, by agent $i\in\mathcal{N}$, and the corresponding stack vector $\hat{d} \coloneqq [\hat{d}_1,\dots,\hat{d}_N]^\top\in\mathbb{R}^{N\mu}$, for which we also formulate the associated errors as
\begin{subequations} \label{eq:d bounde errors}
\begin{align}
e_{d_{\scriptscriptstyle O}} \coloneqq & 
 \bar{d}_{\scriptscriptstyle O} - \hat{d}_{\scriptscriptstyle O} \in\mathbb{R}^{\mu_{\scr O}}   \label{eq:d bounde errors_1}\\
e_d \coloneqq & \begin{bmatrix}
e_{d_1}^\top, \dots, e_{d_N}^\top
\end{bmatrix}^\top  \coloneqq
\bar{d} - \hat{d} \in\mathbb{R}^{N\mu}. \label{eq:d bounde errors_2}
\end{align}
\end{subequations}

\noindent Next, we design the reference velocity 
\begin{equation}
v_f \coloneqq v_\text{d} - K_f e =
\begin{bmatrix}
\dot{p}_\text{d} - k_p e_p \\ \omega_{\text{d}} + k_\zeta e_\epsilon
\end{bmatrix} \label{eq:v_f}
\end{equation} 
where $v_\text{d} \coloneqq [\dot{p}_\text{d}^\top, \omega_\text{d}^\top]^\top$, $e \coloneqq [e^\top_p, -e^\top_\epsilon]^\top\in\mathbb{R}^6$, and $K_f \coloneqq \text{diag}\{k_pI_3,k_\zeta I_3\}$, with $k_p, k_\zeta$ positive control gains.
We also introduce the respective velocity error $e_v$ as
\begin{equation}
e_{v_f} \coloneqq v_{\scriptscriptstyle O} - v_f, \label{eq:e_v_f}
\end{equation} 
and design the adaptive control law $u_i$ in \eqref{eq:coupled dynamics}, for each agent $i\in\mathcal{N}$, as:
\begin{align}
& u_i = Y_i\Big(q_i,\dot{q}_i,v_f,\dot{v}_f \Big)\hat{\vartheta}_i + \delta_i(q_i,\dot{q}_i,t)\hat{d}_i + J_{M_i}(q) \Big[ - e \notag \\
& Y_{\scriptscriptstyle O}\Big(x_{\scriptscriptstyle O},\dot{x}_{\scriptscriptstyle O},v_f,\dot{v}_f)\hat{\vartheta}_{\scriptscriptstyle O}  + \delta_{\scriptscriptstyle O}(x_{\scriptscriptstyle O},\dot{x}_{\scriptscriptstyle O},t)\hat{d}_{\scriptscriptstyle O} - K_ve_{v_f} \Big],  \label{eq:control laws adaptive quat}
\end{align}
where $K_{v}$ is a diagonal positive definite gain matrix,
and $J_{M_i}:\mathbb{R}^{n}\to\mathbb{R}^{6\times 6} $ is the matrix \cite{erhart2015internal}
\begin{equation}\label{eq:J_Hirche}
	J_{M_i}(q) \coloneqq \begin{bmatrix}
	m^\star_i[m^\star_{\scr O}]^{-1} I_3 & m^\star_i [J^\star_{\scr O}(q)]^{-1} S(p_{\scr O/E_i}(q_i)) \\
	0_{3\times 3} & J^\star_i [J^\star_{\scr O}(q)]^{-1}
	\end{bmatrix}
\end{equation}
for some positive coefficients $m^\star_i\in\mathbb{R}_{> 0}$ and positive definite matrices $J^\star_i\in\mathbb{R}^{3\times3}$, $\forall i\in\mathcal{N}$, satisfying
\begin{align*}
	&m^\star_{\scr O} = \sum_{i\in\mathcal{N}} m^\star_i, \ \ \sum_{i\in\mathcal{N}} p_{\scr O/E_i}(q_i) m^\star_i = 0_{3} \notag \\
	&J^\star_{\scr O}(q) = \sum_{i\in\mathcal{N}} J^\star_i - \sum_{i\in\mathcal{N}} m^\star_i[S(p_{\scr O/E_i}(q_i))]^2.
\end{align*}
In addition, we design the following adaptation laws:
\begin{subequations} \label{eq:adaptation laws}
\begin{align}
& \dot{\hat{\theta}}_i = -\gamma_i \Big[Y_i\Big(q_i,\dot{q}_i,v_f,\dot{v}_f \Big)\Big]^\top J_{\scr O_i}(q_i) e_{v_f} \label{eq:adaptation laws_theta_i}\\
& \dot{\hat{\theta}}_{\scriptscriptstyle O} = -\gamma_{\scriptscriptstyle O}\Big[Y_{\scriptscriptstyle O}\Big(x_{\scriptscriptstyle O},\dot{x}_{\scriptscriptstyle O},v_f,\dot{v}_f \Big)\Big]^\top e_{v_f} \label{eq:adaptation laws_theta_o}\\
& \dot{\hat{d}}_i = -\beta_i[\delta_i(q_i,\dot{q}_i,t)]^\top J_{\scriptscriptstyle O_i}(q_i)e_{v_f} \label{eq:adaptation laws_d_i} \\
& \dot{\hat{d}}_{\scriptscriptstyle O} =  -\beta_{\scriptscriptstyle O}[\delta_{\scriptscriptstyle O}(x_{\scriptscriptstyle O},\dot{x}_{\scriptscriptstyle O},t)]^\top e_{v_f}, \label{eq:adaptation laws_d_o}
\end{align}
\end{subequations}
with arbitrary bounded initial conditions, where $\beta_i, \beta_{\scriptscriptstyle O}, \gamma_i, \gamma_{\scriptscriptstyle O} \in\mathbb{R}_{> 0}$ are positive gains, $\forall i\in\mathcal{N}$.  The control and adaptation laws can be written in vector form 
\begin{subequations} \label{eq:laws vector forms}
\begin{align}
	&u = \widetilde{Y}(q,\dot{q},v_{\scr O},\dot{v}_{\scr O})\hat{\vartheta} + \widetilde{\delta}(q,\dot{q},t)\hat{d} + G^{+}_M(q)\Big[ - e \notag\\ &  Y_{\scriptscriptstyle O}\Big(x_{\scriptscriptstyle O},\dot{x}_{\scriptscriptstyle O},v_f,\dot{v}_f)\hat{\vartheta}_{\scriptscriptstyle O}  + \delta_{\scriptscriptstyle O}(x_{\scriptscriptstyle O},\dot{x}_{\scriptscriptstyle O},t)\hat{d}_{\scriptscriptstyle O}  - K_ve_{v_f} \Big] \label{eq:control laws adaptive quat vector form} \\ 
	& \dot{\hat{\vartheta}} = -\Gamma [\widetilde{Y}(q,\dot{q},v_f,\dot{v}_f)]^\top G(q)e_{v_f}  \label{eq:adaptation laws_theta_i vector form} \\
	& \dot{\hat{d}} = -B[\widetilde{\delta}(q,\dot{q},t)]^\top G(q)e_{v_f} \label{eq:adaptation laws_d_i vector form} \\
	& \dot{\hat{\theta}}_{\scriptscriptstyle O} = -\gamma_{\scriptscriptstyle O}\Big[Y_{\scriptscriptstyle O}\Big(x_{\scriptscriptstyle O},\dot{x}_{\scriptscriptstyle O},v_f,\dot{v}_f \Big)\Big]^\top e_{v_f} \label{eq:adaptation laws_theta_o vector form}\\
	& \dot{\hat{d}}_{\scriptscriptstyle O} =  -\beta_{\scriptscriptstyle O}[\delta_{\scriptscriptstyle O}(x_{\scriptscriptstyle O},\dot{x}_{\scriptscriptstyle O},t)]^\top e_{v_f}, \label{eq:adaptation laws_d_o vector form} 
\end{align}
\end{subequations}
where $G^{+}_M(q) \coloneqq [J^\top_{M_1}(q),\dots,J^\top_{M_N}(q)]^\top \in\mathbb{R}^{6N\times6}$, $B \coloneqq \text{diag}\{[\beta_i]_{i\in\mathcal{N}}\}$, and $\Gamma \coloneqq \text{diag}\{[\gamma_i]_{i\in\mathcal{N}}\}$. The matrix $G^{+}_M(q)$ was introduced in \cite{erhart2015internal}, where it was proved that it yields a load distribution that is free of internal forces. The parameters $m^\star_{\scr O}, m^\star_i$ are used to distribute the object's needed effort (the term that right multiplies $G^{+}_M(q)$ in \eqref{eq:control laws adaptive quat vector form}) to the agents.

\begin{remark}[\textbf{Decentralized manner (adaptive controller)}]
	Notice from \eqref{eq:control laws adaptive quat} and \eqref{eq:adaptation laws} that the overall control protocol is decentralized in the sense that the agents calculate their own control signals without communicating with each other. In particular, the control gains and the desired trajectory can be transmitted off-line to the agents, which
	can compute the object's pose and velocity, and hence the signals $e$, $v_f$, $e_{v_f}$ from the inverse kinematics. For the computation of $J_{M_i}(q)$, each agent needs feedback from all $q_i$ to compute $S(p_{\scr O/E_i}(q_i), \forall i\in\mathcal{N}$. However, by exploiting the rigidity of the grasps, it holds that $p_{\scr O/E_i}(q_i) = R_{\scr O}(q_i) p^{\scr O}_{\scr O/E_i}$. Therefore, since all agents can compute $R_{\scr O}$, the computation of $J_{M_i}(q)$ reduces to knowledge of the offsets $p^{\scr O}_{\scr E_i/O}$, which can also be transmitted off-line to the agents. Moreover, by also transmitting off-line to the agents the initial conditions $\hat{\theta}_{\scr O}$, $\hat{d}_{\scr O}$, and via the adaptation laws \eqref{eq:adaptation laws_theta_o vector form}, \eqref{eq:adaptation laws_d_o vector form}, each agent has access to the adaptation signals $\hat{\theta}_{\scr O}(t), \hat{d}_{\scr O}(t)$, $\forall t\in\mathbb{R}_{\geq 0}$.
	Finally, the structure of the functions $\delta_i$ ,$\delta_{\scr O}$, $Y_i$, $Y_{\scr O}$, as well as the constants $m^\star_i$, $J^\star_i$ can be also known by the agents a priori.
\end{remark}

The following theorem summarizes the main results of this subsection.

\begin{theorem}
Consider $N$ robotic agents rigidly grasping an object with coupled
dynamics described by (\ref{eq:coupled dynamics}) and unknown dynamic
parameters. Then, under Assumptions \ref{ass:feedback}-\ref{ass:disturbance bound}, by applying the control protocol \eqref{eq:control laws adaptive quat}
with the adaptation laws \eqref{eq:adaptation laws},  
the object pose converges asymptotically to the desired pose trajectory. 
Moreover, all closed loop signals are bounded. 
\end{theorem}

\begin{IEEEproof}
Consider the nonnegative function 
\begin{align}
& V \coloneqq \tfrac{1}{2}e_p^\top e_p+2(1 - e_\varphi) + \tfrac{1}{2}e^\top_{v_f}\widetilde{M}(x)e_{v_f} +\tfrac{1}{2}e^\top _\vartheta \Gamma^{-1} e_\vartheta + \notag\\  & + \tfrac{1}{2\gamma_{\scr O}}e^\top _{\vartheta_{\scriptscriptstyle O}}e_{\vartheta_{\scriptscriptstyle O}} + \tfrac{1}{2}e^\top _d B^{-1} e_d + \tfrac{1}{2\beta_{\scr O}}e^\top _{d_{\scriptscriptstyle O}}e_{d_{\scriptscriptstyle O}}, \label{eq:Lyap_f}
\end{align}
By taking the derivative of $V$ and using \eqref{eq:e_v_f}, \eqref{eq:v_f}, \eqref{eq:parameter linearity 2}, and Lemma \ref{lem:coupled dynamics skew symmetry}, we obtain
\begin{align}
& \dot{V} = -e^\top K_f e + e^\top_{v_f} \Big[ [G(q)]^\top\Big(u - \widetilde{Y}(q,\dot{q},v_f,\dot{v}_f)\vartheta - \notag \\ & \widetilde{\delta}(q,\dot{q},t)\bar{d}\Big) + e   
  - Y_{\scriptscriptstyle O}\Big(x_{\scriptscriptstyle O},\dot{x}_{\scriptscriptstyle O},\dot{v}_f,\dot{v}_f \Big) \vartheta_{\scriptscriptstyle O} - \delta_{\scriptscriptstyle O}(x_{\scriptscriptstyle O}, \dot{x}_{\scriptscriptstyle O},t)\bar{d}_{\scriptscriptstyle O} \notag\\ 
&-e^\top_{\vartheta}\Gamma^{-1}\dot{\hat{\vartheta}} - \tfrac{1}{\gamma_{\scr O}}e^\top_{\vartheta_{\scriptscriptstyle O}}\dot{\hat{\vartheta}}_{\scriptscriptstyle O} - e^\top_d B^{-1}\dot{\hat{d}} - \tfrac{1}{\beta_{\scr O}}e^\top_{d_{\scriptscriptstyle O}} \dot{\hat{d}}_{\scriptscriptstyle O}, \notag
\end{align}
and after substituting the adaptive control and adaptation laws \eqref{eq:laws vector forms} and using the fact that $[G(q)]^\top G^{+}_M = I_6$, 
\begin{align}
& \dot{V} = -e^\top K_f e - e^\top_{v_f}K_{v}e_{v_f} - \notag \\
& e^\top_{v_f} \Big[ [G(q)]^\top \Big( 
  \widetilde{Y}(q,\dot{q},v_f,\dot{v}_f) e_{\vartheta} + \widetilde{\delta}(q,\dot{q},t) e_d\Big)  +  \notag\\ 
&  Y_{\scriptscriptstyle O}(x_{\scriptscriptstyle O},\dot{x}_{\scriptscriptstyle O},v_f,\dot{v}_f) e_{\vartheta_{\scriptscriptstyle O}} + \delta_{\scriptscriptstyle O}(x_{\scriptscriptstyle O}, \dot{x}_{\scriptscriptstyle O},t) e_{d_{\scriptscriptstyle O}}\Big] + \notag \\
& e^\top_{\vartheta}[\widetilde{Y}(q,\dot{q},v_f,\dot{v}_f)]^\top G(q)e_{v_f} +  e^\top_d[\widetilde{\delta}(q,\dot{q},t)]^\top e_{v_f} + \notag \\ &e^\top_{\vartheta_{\scriptscriptstyle O}}[Y_{\scr O}(x_{\scr O},\dot{x}_{\scr O},v_f,\dot{v}_f)]^\top e_{v_f} + e^\top_{d_{\scriptscriptstyle O}} [\delta_{\scr O}(x_{\scr O},\dot{x}_{\scr O},t)]^\top e_{v_f},  \notag \\
&= -k_p\|e_p \|^2 - k_\zeta\| e_\epsilon \|^2 - e^\top_{v_f}K_{v}e_{v_f},   \label{eq:V_dot adaptive final}
\end{align}
which is non-positive. {Note, however, that $\dot{V}$ is not negative definite, and we need to invoke invariance-like properties to conclude the asymptotic stability of $e_p, e_\epsilon, e_{v_f}$. Since the closed-loop system is non-autonomous (this can be verified by inspecting \eqref{eq:error_dynamics}, the derivative of \eqref{eq:e_v_f} and \eqref{eq:laws vector forms}), LaSalle's invariance principle is not applicable, and we thus employ Barbalat's lemma \cite[Lemma 8.1]{lavretsky13adaptive}}. From \eqref{eq:V_dot adaptive final} we conclude 
the boundedness of $V$ and of $\chi$, which implies  the boundedness of the dynamic terms $\widetilde{M}(x), \widetilde{C}(x), \widetilde{g}(x)$. 
Moreover, by invoking the boundedness of $p_\text{d}(t), v_\text{d}(t),\omega_\text{d}(t), \dot{v}_\text{d}(t), \dot{\omega}_\text{d}(t)$, we conclude the boundedness of $v_f,v_{\scriptscriptstyle O},v_i,\hat{\vartheta}_{\scriptscriptstyle O}$, $\hat{\vartheta}$, $\hat{d}$, $\hat{d}_{\scriptscriptstyle O}$. By differentiating \eqref{eq:error_dynamics}, we also conclude the boundedness of $\dot{v}_f$ and therefore, the boundedness of the control and adaptation laws \eqref{eq:control laws adaptive quat} and \eqref{eq:adaptation laws}. Thus, we can conclude the boundedness of the second derivative $\ddot{V}$ and {by invoking Corollary 8.1 of \cite{lavretsky13adaptive}}, the uniform continuity of $\dot{V}$. Therefore, according to Barbalat's lemma, we deduce that $\lim_{t\to\infty}\dot{V}(t) = 0$ and, consequently, that $\lim_{t\to\infty}e_p(t) = 0_3$, $\lim_{t\to\infty}e_{v_f}(t) = 0_6$, and $\lim_{t\to\infty} \| e_\epsilon(t) \|^2 = 0$, which, given that $e_\zeta$ is a unit quaternion, leads to the configuration $(e_p, e_{v_f}, e_\varphi, e_\epsilon) = (0_3,0_6,\pm 1,0_3)$. 

\end{IEEEproof}

\begin{remark}[\textbf{Unwinding}]
Note that the two configurations where $e_\varphi = 1$ and $e_\varphi = -1$  represent the same orientation. The closed loop dynamics of $e_\varphi$, as given in \eqref{eq:eta_error dynamics}, can be written, in view of \eqref{eq:v_f}, as $\dot{e}_\varphi = k_\zeta\tfrac{1}{2}\|e_\epsilon\|^2 +  \tfrac{1}{2}[0^\top_3, e^\top_\epsilon]e_{v_f}$. Since the first term is always positive, we conclude that the equilibrium point $(e_p, e_{v_f}, e_\varphi, e_\epsilon) = (0_3,0_6,-1,0_3)$ is unstable. Therefore, there might be trajectories close to the configuration $e_\varphi = -1$ that will move away and approach $e_\varphi = 1$, i.e., a full rotation will be performed to reach the desired orientation (of course, if the system starts at the equilibrium $(e_p, e_{v_f}, e_\varphi,e_\epsilon) = (0_3,0_6,-1,0_3)$, it will stay there, which also corresponds to the desired orientation behavior). This is the so-called \textit{unwinding phenomenon} \cite{bhat2000topological}.
Note, however, that the desired equilibrium point $(e_p, e_{v_f}, e_\varphi,e_\epsilon) = (0_3,0_6,1,0_3)$ is \textit{\textbf{eventually attractive}}, meaning that for each $\delta_\varepsilon >0$, there exist finite a time instant $T\geq 0$ such that $1 - e_\varphi(t) < \delta_\varepsilon, \forall t > T \geq 0$. A similar behavior is observed if we stabilize the point $e_\varphi = -1$ instead of $e_\varphi = 1$, by setting $e \coloneqq [e^\top_p, e^\top_\epsilon]^\top$ in \eqref{eq:v_f} and considering the term $2(1 + e_\varphi)$ instead of $2(1 - e_\varphi)$ in the function \eqref{eq:Lyap_f}. 

In order to avoid the unwinding phenomenon, instead of the error $e = [e^\top_p, -e^\top _\epsilon]^\top$, we {can choose} $e = [e^\top_p, -e_\varphi e^\top _\epsilon]^\top$ (see our preliminary result \cite{Verginis_ifac}). Then by replacing the term $1 - e_\varphi$ with $1-e^2_\varphi$ in \eqref{eq:Lyap_f} and using 
\eqref{eq:laws vector forms}, we conclude by proceeding with a similar analysis that 
$(e_p, \|e_\epsilon\|e_\varphi, e_{v_f}) \rightarrow (0_3,0,0_6)$, which implies that the system is asymptotically driven to either the configuration $(e_p, e_{v_f}, e_\varphi,e_\epsilon) = (0_3,0_6,\pm 1,0_3)$, which is the desired one, or a configuration  $(e_p, e_{v_f}, e_\varphi,e_\epsilon) = (0_3,0_6,0,\widetilde{e}_\epsilon)$, where $\widetilde{e}_\epsilon\in S^2$ is a unit vector. The latter represents a set of invariant undesired equilibrium points. The closed loop dynamics are $\dot{e}_\varphi = \frac{1}{2}e_\varphi\|e_\epsilon\|^2 + \frac{1}{2}[0^\top_3, e^\top_\epsilon]e_v$, and 
$\dot{\|e_\epsilon\|^2} = -e_\varphi^2\|e_\epsilon\|^2  - e_\varphi[0^\top_3, e^\top_\epsilon]e_v$. 
We can conclude from the term $[0^\top_3, e^\top_\epsilon]e_v$ that there exist trajectories that can bring the system close to the undesired equilibrium, rendering thus the point  $(e_p, e_{v_f}, e_\varphi,e_\epsilon) = (0_3,0_6,\pm 1,0_3)$ only locally asymptotically stable. 
It has been proved that $e_\varphi = \pm 1$ cannot be globally stabilized with a purely continuous controller \cite{bhat2000topological}. Discontinuous control laws have also been proposed (e.g., \cite{mayhew2011quaternion}), whose  combination with adaptation techniques constitutes part of our future research directions. Another possible direction is tracking on $SO(3)$ (see e.g., \cite{lee10, Automatica_formation_18}).
\end{remark}

\begin{remark} [\textbf{Robustness (adaptive controller)}]
Notice also that the control protocol compensates the uncertain dynamic parameters and external disturbances through the adaptation laws \eqref{eq:adaptation laws}, although the errors \eqref{eq:adaptation errors}, \eqref{eq:d bounde errors} do not converge to zero, but remain bounded. Finally, the control gains $k_p, k_\zeta, K_{v}$ can be tuned appropriately so that the proposed control inputs do not reach motor saturations in real scenarios.
\end{remark}

	

\subsection{Prescribed Performance Control} \label{subsec:PPC Controller}

In this section, we adopt the concepts and techniques of prescribed performance control, recently proposed in \cite{bechlioulis2008robust}, in order to achieve predefined  transient and steady state response for the derived error, as well as ensure that $\theta_{\scriptscriptstyle O}(t)\in(-\tfrac{\pi}{2},\tfrac{\pi}{2}), \forall t\in\mathbb{R}_{\geq 0}$. As stated in Section \ref{subsec:PPC}, prescribed performance characterizes the behavior where a signal evolves strictly within a predefined region that is bounded by absolutely decaying functions of time, called performance functions. This signal is represented by the object's pose error 
\begin{equation}\label{eq:ppc errors}
e_s \coloneqq \begin{bmatrix}
e_{s_x}, e_{s_y}, e_{s_z}, e_{s_\phi}, e_{s_\theta}, e_{s_\psi}
\end{bmatrix}^\top
\coloneqq 
x_{\scriptscriptstyle O} - x_\text{d}
\end{equation}
Firstly, we relax Assumption \ref{ass:disturbance bound}:
\begin{ass} [Uncertainties/Disturbance bound]  \label{ass:disturbance bound_ppc}$\ $
	{The functions $d_{\scr O}(x_{\scr O},\dot{x}_{\scr O},t)$ and $d_{\scr O}(q_i,\dot{q}_i,t)$ are continuous in $(x_{\scr O},\dot{x}_{\scr O})$ and $(q_i,\dot{q}_i)$, respectively, and bounded in $t$ by unknown positive constants $\bar{d}_{\scr O}$ and $\bar{d}_i$, respectively, $\forall i\in\mathcal{N}$.}
\end{ass} 
The mathematical expressions of prescribed performance are given by the following inequalities:
\begin{equation}
-\rho_{s_k}(t)< e_{s_k}(t) <\rho_{s_k}(t), \forall k\in\mathcal{K},  \label{eq:ppc}
\end{equation}
where $\mathcal{K} \coloneqq \{x,y,z,\phi,\theta,\psi\}$ and $\rho_k:\mathbb{R}_{\geq 0}\rightarrow\mathbb{R}_{> 0}$, with 
\begin{equation}
\rho_{s_k}(t) \coloneqq (\rho_{s_k,\scriptscriptstyle 0}-\rho_{s_k,\scriptscriptstyle\infty})\exp(-l_{s_k}t)+\rho_{s_k,\scriptscriptstyle \infty},  \ \forall k\in\mathcal{K},  \label{eq:rho}
\end{equation}
are designer-specified, smooth, bounded and decreasing positive functions of time with $l_{s_k}, \rho_{s_k,\scriptscriptstyle \infty}, k\in\mathcal{K}$, positive parameters incorporating the desired transient and steady state performance respectively. The terms $\rho_{s_k,\scriptscriptstyle \infty}$ can be set arbitrarily small, achieving thus practical convergence of the errors to zero. Next, we propose a state feedback control protocol that does not incorporate any information on the agents' or the object's dynamics or the external disturbances and guarantees \eqref{eq:ppc} for all $t\in\mathbb{R}_{\geq 0}$. 
Given the errors \eqref{eq:ppc errors}:\\
\textbf{Step I-a}. Select the functions $\rho_{s_k}$ as in \eqref{eq:rho} with 
\begin{enumerate}[(i)]
\item 	$\rho_{s_\theta,\scriptscriptstyle 0}  = \rho_{s_\theta}(0)= \theta^*, \rho_{s_k,\scriptscriptstyle 0} = \rho_{s_k}(0) > \lvert e_{s_k}(0) \rvert, \forall k\in\mathcal{K}\backslash\{\theta\}$, 
\item $l_{s_k} \in\mathbb{R}_{>0}, \forall k\in\mathcal{K}$,
\item $\rho_{s_k,\scriptscriptstyle \infty}\in(0,\rho_{s_k,0}), \forall k\in\mathcal{K}$,
\end{enumerate}
where $\theta^*$ is a positive constant satisfying $\theta^* + \bar{\theta} < \frac{\pi}{2}$ and $\bar{\theta}$ is the desired trajectory bound (see statement of Problem \ref{prob:problem1}).  
\textbf{Step I-b}. Introduce the normalized errors 
\begin{equation}
\xi_{s} \coloneqq \begin{bmatrix}
\xi_{s_x}, \dots, \xi_{s_\psi} \end{bmatrix}^\top
\coloneqq \rho_s^{-1}e_s,	\label{eq:ksi_s}
\end{equation}
where $\rho_s \coloneqq \text{diag}\{\left[\rho_{s_k}\right]_{k\in\mathcal{K}}\}\in\mathbb{R}^{6\times6}$, as well as the transformed state functions $\varepsilon_s$, and signals $r_s:(-1,1)^6\to\mathbb{R}^{6\times 6}$, with  
\begin{align}
& \varepsilon_s \coloneqq  \begin{bmatrix}
\varepsilon_{s_x}, \dots, \varepsilon_{s_\psi} \end{bmatrix}^\top 
\coloneqq 
\begin{bmatrix}
\ln\Big(\frac{1 + \xi_{s_x}}{1 - \xi_{s_x}} \Big), \dots, \ln\Big(\frac{1 + \xi_{s_\psi}}{1 - \xi_{s_\psi}}\Big) \end{bmatrix}^\top \label{eq:epsilon_s}\\
& r_s(\xi_s) \coloneqq 
\text{diag}\{[r_{s_k}(\xi_{s_k})]_{k\in\mathcal{K}}\}
 \coloneqq 
\text{diag}\Big\{ \Big [\frac{\partial \varepsilon_{v_k}}{\partial \xi_{s_k}} \Big]_{k\in\mathcal{K}} \Big \} \notag \\
&\hspace{10mm} = \text{diag}\Big\{ \Big [\frac{2}{1-\xi^2_{s_k} } \Big]_{k\in\mathcal{K}} \Big \} \label{eq:r_s},
\end{align} 
and design the reference velocity vector: 
\begin{align}
& v_{r} \coloneqq 
 -g_s J_{\scriptscriptstyle O}\Big( \eta_\text{d} + \rho_{s_\eta}\xi_{s_\eta} \Big)^{-1}\rho_s^{-1}r_s(\xi_s)\varepsilon_{s}, \label{eq:v_r}
\end{align}
where $\rho_{s_\eta} \coloneqq \text{diag}\{\rho_{s_\phi},\rho_{s_\theta},\rho_{s_\psi}\}$, $\xi_{s_\eta}\coloneqq [\xi_{s_\phi},\xi_{s_\eta},\xi_{s_\phi}]^\top$, and we have further used the relation $\xi_s = \rho_s^{-1}(x_{\scriptscriptstyle O}- x_\text{d})$ from \eqref{eq:ppc errors} and \eqref{eq:ksi_s}.\\
\textbf{Step II-a}. Define the velocity error vector 
\begin{equation}
e_v \coloneqq \begin{bmatrix}
e_{v_x}, \dots,  e_{v_\psi}  	
\end{bmatrix}^\top
\coloneqq v_{\scriptscriptstyle O} - v_r,  \label{eq:e_v_r}
\end{equation} 
and select the corresponding positive performance functions $\rho_{v_k}:\mathbb{R}_{\geq 0}\rightarrow\mathbb{R}_{>0}$ with $\rho_{v_k}(t) \coloneqq (\rho_{v_k, \scriptscriptstyle 0} - \rho_{v_k,\scriptscriptstyle \infty})\exp(-l_{v_k}t) + \rho_{v_k,\scriptscriptstyle \infty}$, such that $\rho_{v_k,\scriptscriptstyle 0}  = \lVert e_{v}(0) \rVert + \alpha, l_{v_k}>0$ and $\rho_{v_k,\scriptscriptstyle \infty}\in(0,\rho_{v_k,0}), \forall k\in\mathcal{K}$, where $\alpha$ is an arbitrary positive constant.\\
\textbf{Step II-b}. Define the normalized velocity error   
\begin{equation}
\xi_v \coloneqq \begin{bmatrix}
\xi_{v_x}, \dots, \xi_{v_\psi} \end{bmatrix}^\top
\coloneqq \rho_v^{-1}e_v,	\label{eq:ksi_v}
\end{equation}	
where $\rho_v\coloneqq\text{diag}\{\left[\rho_{v_k}\right]_{k\in\mathcal{K}}\}$, as well as the transformed states 
$\varepsilon_v$ and signals $r_v:(-1,1)^6\to\mathbb{R}^{6\times6}$, with 
\begin{align}
&\varepsilon_v \coloneqq  \begin{bmatrix}
\varepsilon_{v_x}, \dots, \varepsilon_{v_\psi} \end{bmatrix}^\top
 \coloneqq 
\begin{bmatrix}
\ln\Big(\frac{1 + \xi_{v_x}}{1 - \xi_{v_x}} \Big), \dots, \ln\Big(\frac{1 + \xi_{v_\psi}}{1 - \xi_{v_\psi}}\Big) \end{bmatrix}^\top \notag \\  
&r_v(\xi_v) \coloneqq 
\text{diag}\{[r_{v_k}(\xi_{v_k})]_{k\in\mathcal{K}}\}
 \coloneqq 
\text{diag}\Big\{ \Big [\frac{\partial \varepsilon_{v_k}}{\partial \xi_{v_k}} \Big]_{k\in\mathcal{K}} \Big \} \notag \\
&\hspace{10mm} = \text{diag}\Big\{ \Big [\frac{2}{1-\xi^2_{v_k} } \Big]_{k\in\mathcal{K}} \Big \} \label{eq:r_v},
\end{align} 
and design the decentralized feedback control protocol for each agent $i\in\mathcal{N}$ as 
\begin{equation}
u_i \coloneqq -g_v J_{M_i}(q) \rho_v^{-1}r_v(\xi_v)\varepsilon_v, \label{eq:control_law_ppc}
\end{equation}
where $g_v$ is a positive constant gain and $J_{M_i}$ as defined in \eqref{eq:J_Hirche}.
The control laws \eqref{eq:control_law_ppc} can be written in vector form $u \coloneqq [u^\top_1,\dots,u^\top_N]^\top$, with:
\begin{align}
\hspace{-2mm}u = -g_v G^{+}_M(q)\rho_v^{-1}r_v(\xi_v)\varepsilon_v. \label{eq:control_law_ppc_vector_form}
\end{align}

\begin{remark} [\textbf{Decentralized manner and robustness (PPC)}]
Similarly to \eqref{eq:laws vector forms},  notice from \eqref{eq:control_law_ppc} that each agent $i\in\mathcal{N}$ can calculate its own control signal, without communicating with the rest of the team, rendering thus the overall control scheme decentralized. The  terms $l_k$, $\rho_{k,0}$, $\rho_{k,\scriptscriptstyle \infty}$, $\alpha$, $l_{v_k}$, and $\rho_{v_k,\infty}$, $k\in\mathcal{K}$ needed for the calculation of the performance functions can be transmitted off-line to the agents. Moreover, the Prescribed Performance Control protocol is also robust to uncertainties of model uncertainties and external disturbances. In particular, note that the control laws do not even require the structure of the terms $\widetilde{M}, \widetilde{C}, \widetilde{g}, \widetilde{d}$, but only the positive definiteness of $\widetilde{M}$, as will be observed in the subsequent proof of Theorem \ref{th:thorem_ppc}. It is worth noting that, in the case that one or more agent failed to participate in the task, then the remaining agents would need to appropriately update their control protocols (e.g., update $J_{M_i}$) to compensate for the failure.   
\end{remark}

\begin{remark} [\textbf{Internal forces}]
Internal force regulation can be also guaranteed by including in the control laws \eqref{eq:control laws adaptive quat vector form} and \eqref{eq:control_law_ppc_vector_form} a term of the form $(I_{6N} - G^{+}_M(q)G(q)]^\top)\hat{f}_{\text{int,d}}$, where $\hat{f}_{\text{int,d}}\in\mathbb{R}^{6N}$ represents desired internal forces (e.g. to avoid grasp sliding) that can be transmitted off-line to the agents. 
\end{remark}

The main results of this subsection are summarized in the following theorem. 
\begin{theorem}\label{th:thorem_ppc}
Consider $N$ agents rigidly grasping an object with unknown coupled dynamics \eqref{eq:coupled dynamics}. Then, under Assumptions \ref{ass:feedback}-\ref{ass:kinematic singularities}, \ref{ass:disturbance bound_ppc},   
the decentralized control protocol \eqref{eq:ksi_s}-\eqref{eq:control_law_ppc} guarantees that $-\rho_{s_k}(t) < e_{s_k}(t) < \rho_{s_k}(t), \forall k\in\mathcal{K},t\in\mathbb{R}_{\geq 0}$ from all initial conditions satisfying $\lvert \theta(0)-\theta_\text{d}(0) \rvert < \theta^*$ (from \textbf{Step I-a} (i)), with
all closed loop signals being bounded.
\end{theorem}
\begin{IEEEproof}
{The proof consists of two main parts. Firstly, we prove that there exists a maximal solution $(\xi_s(t),\xi_v(t))\in(-1,1)^{12}$ for $t\in[0,\tau_{\max})$, where $\tau_{\max}>0$. Secondly, we prove that $(\xi_s(t), \xi_v(t))$ is contained in a compact subset of $(-1,1)^{12}$ and consequently, that $\tau_{\max} = \infty$.}

\noindent {\underline{Part A}: Consider the combined state $\sigma \coloneqq [q,\xi_s,\xi_v]\in \mathbb{S}\times\mathbb{R}^{12}$. Differentiation of $\sigma$ yields, in view of \eqref{eq:J_o_i}, \eqref{eq:ksi_s} and \eqref{eq:ksi_v}
\begin{align} \label{eq:sigma_dot_1}
	\dot{\sigma} = \begin{bmatrix}
	\widetilde{J}(q) G(q) v_{\scr O} \\
	\rho_s^{-1}(\dot{x}_{\scriptscriptstyle O} - \dot{x}_\text{d} - \dot{\rho}_s \xi_s) \\
	\rho_v^{-1}(\dot{v}_{\scr O} - \dot{v}_r - \dot{\rho}_v \xi_v),
	\end{bmatrix},
\end{align}
where $\widetilde{J}(q)\coloneqq \text{diag}\{[J_i(q_i)^\top(J_i(q_i)J_i(q_i)^\top)^{-1}]_{i\in\mathcal{N}}\}\in\mathbb{R}^{6N\times n}$ is well defined due to Assumption \ref{ass:kinematic singularities}. Then, by employing \eqref{eq:object dynamics}, \eqref{eq:ppc errors}, \eqref{eq:ksi_s}, and \eqref{eq:v_r}-\eqref{eq:control_law_ppc_vector_form} as well as $[G(q)]^\top G^{+}_M = I_6$, we can express the right-hand side of \eqref{eq:sigma_dot_1} as a function of $\sigma$ and $t$, i.e.,
$\dot{\sigma} = f_{\text{cl}}(\sigma,t) \coloneqq 
[
f_{\text{cl},q}(\sigma,t)^\top,
f_{\text{cl},s}(\sigma,t)^\top, 
f_{\text{cl},v}(\sigma,t)^\top 
]^\top$.
The analytic expressions for  $f_{\text{cl},q}(\sigma,t), f_{\text{cl},s}(\sigma,t),
f_{\text{cl},v}(\sigma,t)$ can be found in Appendix \ref{app: A}.}
{Consider now the open and nonempty set $\Omega \coloneqq \mathbb{S}\times(-1,1)^{12}$. The choice of the parameters $\rho_{s_k,0}$ and $\rho_{v_k,0}, k\in\mathcal{K}$ in \textbf{Step I-a} and \textbf{Step II-a}, respectively, along with the fact that the initial conditions satisfy $|\theta_{\scriptscriptstyle O}(0) - \theta_\text{d}(0) | < \theta^*$ imply that $| e_{s_k}(0) | < \rho_{s_k}(0), | e_{v_k}(0) | < \rho_{v_k}(0), \forall k\in\mathcal{K}$ and hence $[ \xi_s(0)^\top, \xi_v(0)^\top ]^\top\in(-1,1)^{12}$. 
Moreover, it can be verified that $f_{\text{cl}}: \Omega\times\mathbb{R}_{\geq 0} \to \mathbb{R}^{n+12}$ is locally Lipschitz in $\sigma$ over the set $\Omega$ and continuous and locally integrable in $t$ for each fixed $\sigma\in\Omega$. Therefore, the hypotheses of Theorem \ref{Th:dynamical systems} stated in Subsection \ref{subsec:Dynamical Systems} hold and the existence of a maximal solution $\sigma:[0,\tau_{\max})\to\Omega$, for $\tau_{\max} > 0$, is ensured. We thus conclude  
\begin{align} \label{eq:ksi_bounded_open}
\xi_{s_k}(t), \  
\xi_{v_k}(t)\in (-1,1)
\end{align}
$\forall k\in\mathcal{K}, t\in[0,\tau_{\max})$, which also implies that $\| \xi_s(t) \| \leq \sqrt{6}$, and $\|\xi_v(t) \| \leq \sqrt{6}, \forall t\in[0,\tau_{\max})$. In the following, we show the boundedness of all closed loop signals and $\tau_{\max} = \infty$. }

{\noindent \underline{Part B}:}
{Note first from \eqref{eq:ksi_bounded_open}, that  $\lvert \theta_{\scriptscriptstyle O}(t) - \theta_\text{d}(t) \rvert < \rho_\theta(t) \leq \rho_{\theta}(0) = \theta^*$, which, since $\theta_\text{d}(t)\in[-\bar{\theta}, \bar{\theta}],\forall t\in\mathbb{R}_{\geq 0}$, implies that $ | \theta_{\scriptscriptstyle O}(t) | \leq \widetilde{\theta}\coloneqq \bar{\theta} + \theta^* < \frac{\pi}{2}, \forall t\in[0,\tau_{\max})$. Therefore, by employing \eqref{eq:J_norm}, one obtains that, $\forall t\in[0,\tau_{\max})$,
\begin{equation}
\| J_{\scriptscriptstyle O}(\eta_{\scriptscriptstyle O}(t)) \| \leq \bar{J}_{\scriptscriptstyle O} \coloneqq \sqrt{\frac{ |\sin(\widetilde{\theta}) |+1}{1-\sin^2(\widetilde{\theta})}} < \infty. \label{eq:J_O_bar}
\end{equation}
Consider now the positive definite function $V_s = \tfrac{1}{2}\| \varepsilon_s\|^2$. Differentiating $V_s$ along the solutions of the closed loop system yields $\dot{V}_s = \varepsilon_s^\top r_s(\xi_s)\rho_s^{-1}\dot{\xi}_s$, which, in view of \eqref{eq:sigma_dot_1}, \eqref{eq:ksi_v}, \eqref{eq:v_r} and the fact that $\dot{x}_{\scr O} = J_{\scr O}(\eta_{\scr O})v_{\scr O} = J_{\scr O}(\eta_{\scr O})(v_r + e_v)$, becomes
\small
\begin{align} 
&\dot{V}_s = - g_s \|\rho_s^{-1}r_s(\xi_s)\varepsilon_s \|^2 - \varepsilon_s^\top r_s(\xi_s)\rho_s^{-1}\Big(\dot{x}_\text{d} +\dot{\rho}_s\xi_s   - J_{\scriptscriptstyle O}(\eta_{\scr O})e_v \Big) \notag \\
& \leq  g_s \|\rho_s^{-1}r_s(\xi_s)\varepsilon_s \|^2 + \|\rho_s^{-1}r_s(\xi_s)\varepsilon_s \| \Big(\|\dot{x}_\text{d}\|    + \| J_{\scriptscriptstyle O}(\eta_{\scr O})\rho_v\xi_v \|  + \notag\\ 
& \| \dot{\rho}_s\xi_s \| \Big).  \label{eq:V_s_dot for Bs}
\end{align}
\normalsize
In view of \eqref{eq:J_O_bar}, \eqref{eq:ksi_bounded_open}, and the structure of $\rho_{s_k}, \rho_{v_k}, k\in\mathcal{K}$, as well as the fact that $v_{\scriptscriptstyle O}(0) = 0$ and the boundedness of $\dot{x}_\text{d} $, the last inequality becomes  
\small
\begin{align}
\dot{V}_s \leq & -g_s \|\rho_s^{-1}r_s(\xi_s)\varepsilon_s \|^2 +   \|\rho_s^{-1}r_s(\xi_s)\varepsilon_s \| \bar{B}_s,   \label{eq:V_s_dot for Bs_2} 
\end{align}
\normalsize
$\forall t\in[0,\tau_{\max})$, where $\bar{B}_s$  
is a positive constant independent of $\tau_{\max}$.
Therefore, $\dot{V}_s$ is negative when $\|\rho_s^{-1}r_s(\xi_s)\varepsilon_s\| > \frac{\bar{B}_s}{g_s}$, which, by employing \eqref{eq:r_s}, the decreasing property of $\rho_{s_k}, k\in\mathcal{K}$ as well as \eqref{eq:ksi_bounded_open}, is satisfied when $\| \varepsilon_s \| > \frac{\max_{k\in\mathcal{K}}\{\rho_{s_k,0}\} \bar{B_s}}{2 g_s}$. Hence, we conclude that
\begin{align}
\| \varepsilon_s(t) \| \leq \bar{\varepsilon}_s \coloneqq \max\Bigg\{ \| \varepsilon_s(0) \|, \frac{\max\limits_{k\in\mathcal{K}}\{\rho_{s_k,0}\} \bar{B_s}}{2 g_s}   \Bigg\}, \label{eq:epsilon_s_bar}
\end{align}   
$\forall t\in[0,\tau_{\max})$. Furthermore, since $| \varepsilon_{s_k}| \leq \|\varepsilon_s\|, \forall k\in\mathcal{K}$, taking the inverse logarithm function from \eqref{eq:epsilon_s}, we obtain 
\small
\begin{align}
-1 < \frac{\exp(-\bar{\varepsilon}_s)-1}{\exp(-\bar{\varepsilon}_s)+1} =: -\bar{\xi}_{s} \leq  \xi_{s_k}(t) \leq \bar{\xi}_{s} \coloneqq \frac{\exp(\bar{\varepsilon}_s)-1}{\exp(\bar{\varepsilon}_s)+1} < 1, \label{eq:ksi_s_bar}
\end{align}
\normalsize
$\forall t\in[0,\tau_{\max})$. Hence, recalling \eqref{eq:r_s} and \eqref{eq:v_r},
we obtain the boundedness of $v_r(t)$, $\forall t\in [0,\tau_{\max})$, and in view of $v_o = v_r + e_v$, \eqref{eq:e_v_r}, \eqref{eq:ksi_bounded_open},  \eqref{eq:J_o_i} and \eqref{eq:J_O_i bound}, the boundedness of $v_o(t)$ and $v_i(t)$, $\forall t\in [0,\tau_{\max})$.
From \eqref{eq:ksi_s_bar}, \eqref{eq:object dynamics 1}, and \eqref{eq:ppc errors} we also conclude the boundedness of $x_{\scr O}(t)$, $\dot{x}_{\scr O}(t)$, $\forall t\in [0,\tau_{\max})$. The coupled kinematics \eqref{eq:coupled_kinematics} and Assumption \ref{ass:kinematic singularities} imply also the boundedness of $p_{\scr E_i}(t)$, $q_i(t)$, and $\dot{q}_i(t)$, $\forall i\in\mathcal{N}$, $[0,\tau_{\max})$.
In a similar vein, by differentiating the reference velocity \eqref{eq:v_r} and using \eqref{eq:epsilon_s}, \eqref{eq:r_s}, and \eqref{eq:epsilon_s_bar}, we also conclude the boundedness of $\dot{v}_r(t)$, $\forall t\in [0,\tau_{\max})$. }

{Applying the aforementioned line of proof, we consider the positive definite function $V_v = \tfrac{1}{2}\| \varepsilon_v\|^2$. By differentiating $V_v$ we obtain $\dot{V}_v = \varepsilon_v^\top r_v(\xi_v)\rho_v^{-1}\dot{\xi}_v$, which, in view of \eqref{eq:sigma_dot_1}, \eqref{eq:e_v_r}, \eqref{eq:coupled dynamics}, becomes 
\begin{align}
& \dot{V}_v = - g_v \varepsilon_v^\top r_v(\xi_v)\rho_v^{-1}\widetilde{M}(x) \rho_v^{-1}r_v(\xi_v)\varepsilon_v \notag\\
&+ \varepsilon_v^\top r_v(\xi_v)\rho_v^{-1}\Big( 
 -  \dot{\rho}_v \xi_v  - \widetilde{M}(x)\Big[\widetilde{C}(x)[\rho_v\xi_v + \notag \\
 & v_r] + \widetilde{g}(x)  + \widetilde{d}(x,t)\Big] -  \dot{v}_r \Big). \label{eq:V_v dot 1} 
\end{align}
Invoking Assumption \ref{ass:disturbance bound_ppc} and the boundedness of $q_i(t)$, $\dot{q}_i(t)$, $x_{\scr O}(t)$, $\dot{x}_{\scr O}(t)$, $\forall t\in[0,\tau_{\max})$, we conclude the boundedness of  $d_{\scr O}(x_{\scr O}(t),\dot{x}_{\scr O}(t),t)$ and $d_i(q_i(t),\dot{q}_i(t),t)$, $\forall t\in[0,\tau_{\max})$. Hence, from  \eqref{eq:J_O_i bound} and \eqref{eq:coupled dynamics}, we also obtain the boundedness of $\widetilde{d}(x(t))$. 
In addition, the continuity of $\widetilde{M}(x), \widetilde{C}(x), \widetilde{g}(x)$ implies their boundedness $\forall t\in[0,\tau_{\max})$.}

{Thus, by combining the aforementioned discussion with the boundedness of $\dot{v}_r$, the positive definitiveness of $\widetilde{M}(x)$, and \eqref{eq:ksi_bounded_open}, we obtain from  \eqref{eq:V_v dot 1} 
\small
\begin{align}
& \dot{V}_v \leq -g_v \underline{m} \| \rho_v^{-1}r_v(\xi_v)\varepsilon_v \|^2  +   \| \rho_v^{-1}r_v(\xi_v)\varepsilon_v \| \bar{B}_v,  \label{eq:V_v dot 2}
\end{align}
\normalsize
$\forall t\in[0,\tau_{\max})$, where $\bar{B}_v$ is a positive and finite constant, independent of $\tau_{\max}$.}

\begin{figure}[t]
	\centering
	\begin{subfigure}[b]{\columnwidth}
		\centering
		\includegraphics[width=.95\columnwidth]{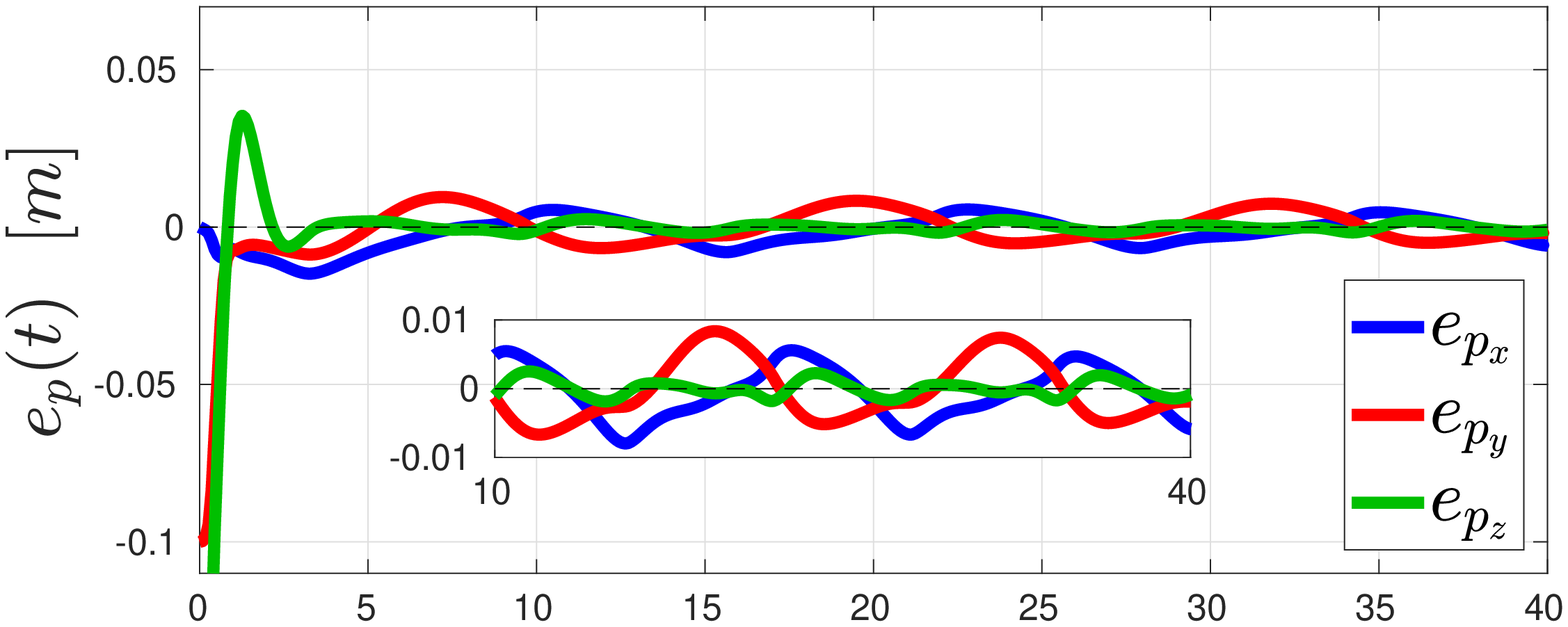}
		\caption{}
		\label{fig:Ng1} 
	\end{subfigure}
	
	\begin{subfigure}[b]{\columnwidth}
		\centering
		\includegraphics[width=.95\columnwidth]{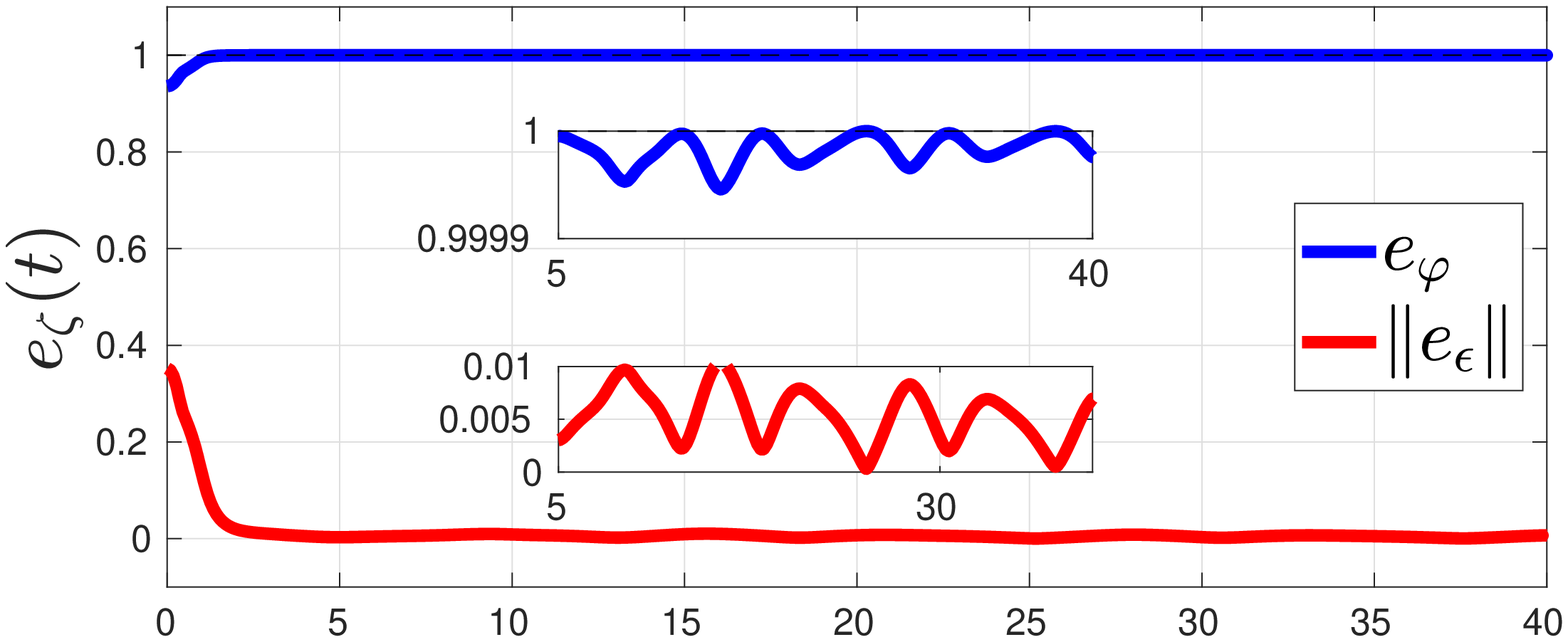}
		\caption{}
		\label{fig:Ng2}
	\end{subfigure}
	
	\begin{subfigure}[b]{\columnwidth}
		\centering
		\includegraphics[width=.95\columnwidth]{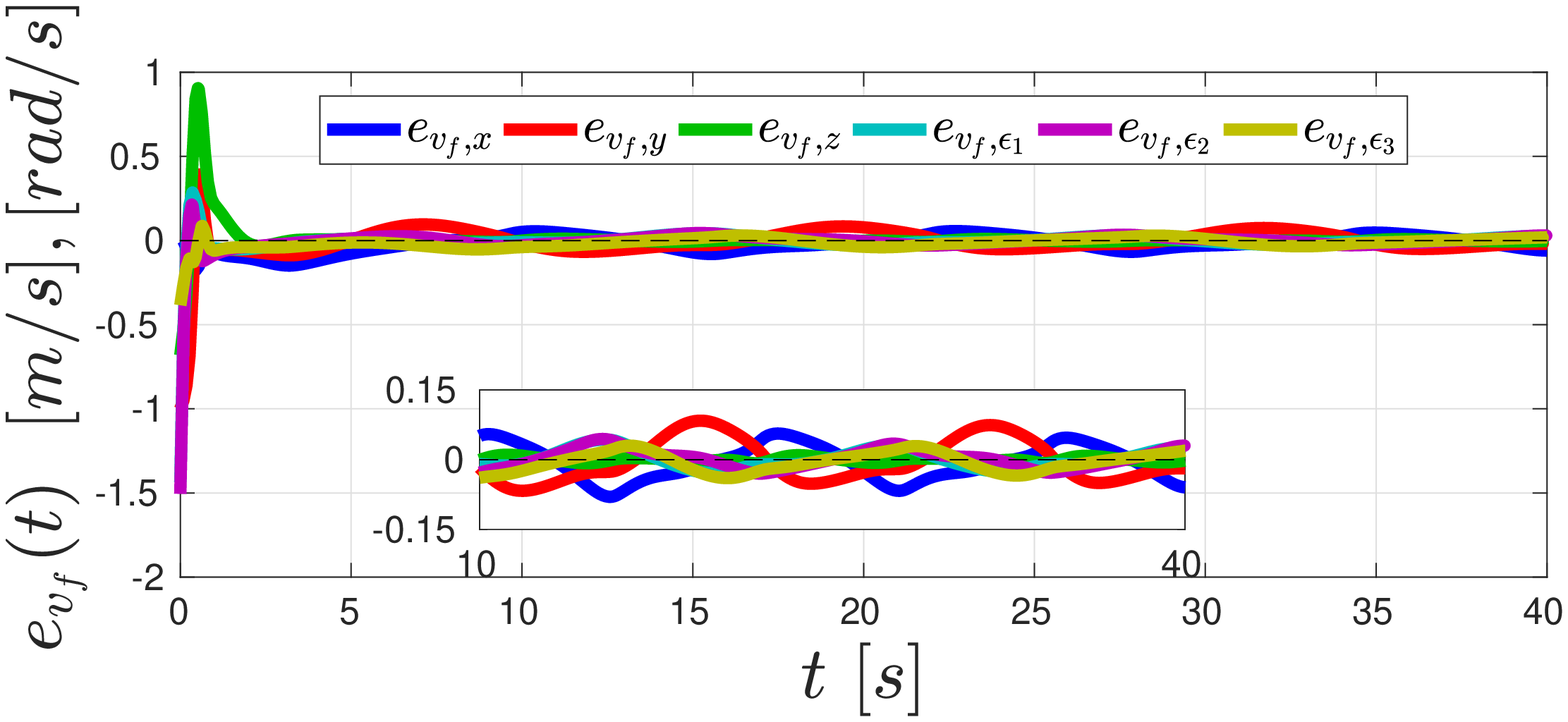}
		\caption{}
		\label{fig:Ng2}
	\end{subfigure}
	
	\caption[]{Simulation results for the control scheme of Section \ref{subsec:Quaternion Controller}; (a): The position errors $e_p(t)$; (b): The quaternion errors $e_\varphi(t)$, $\|e_\varepsilon(t)\|$; (c) The velocity errors $e_{v_f}(t)$, $\forall t\in[0,40]$. A zoomed version of the steady state response has been included in all plots.} \label{fig:adapt_sim_errors}
\end{figure}

\begin{figure}[t]
	\centering
	\begin{subfigure}[b]{\columnwidth}
		\centering
		\includegraphics[width=.95\columnwidth]{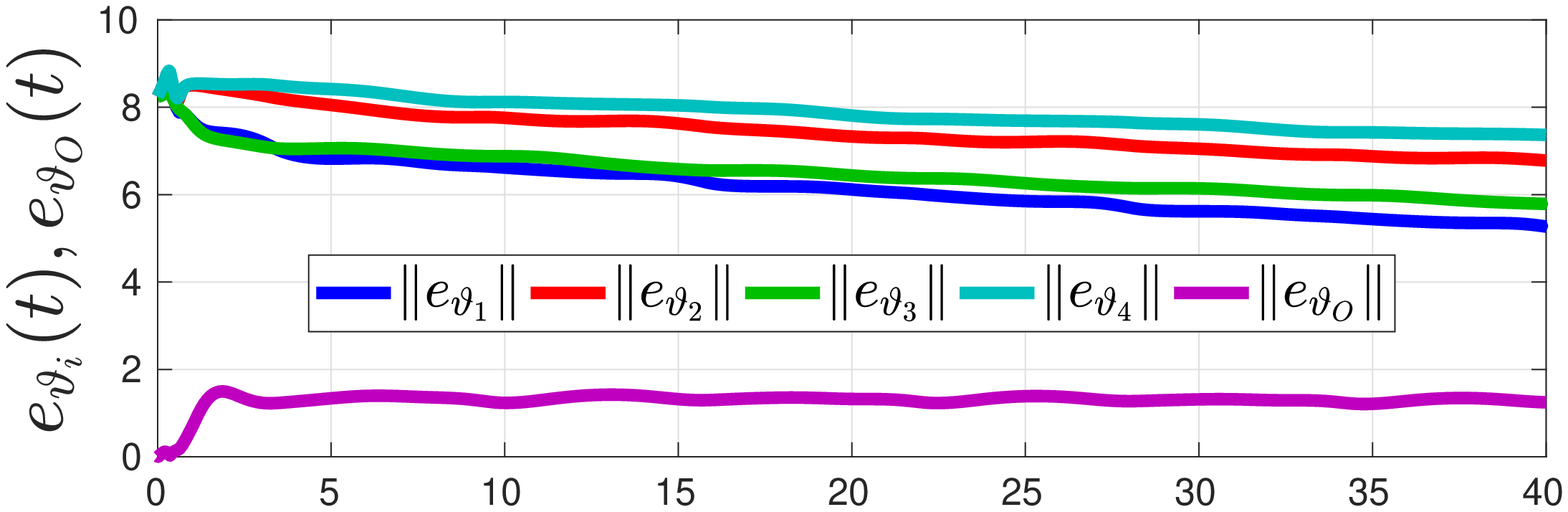}
		\caption{}
		\label{fig:Ng1} 
	\end{subfigure}
	
	\begin{subfigure}[b]{\columnwidth}
		\centering
		\includegraphics[width=.95\columnwidth]{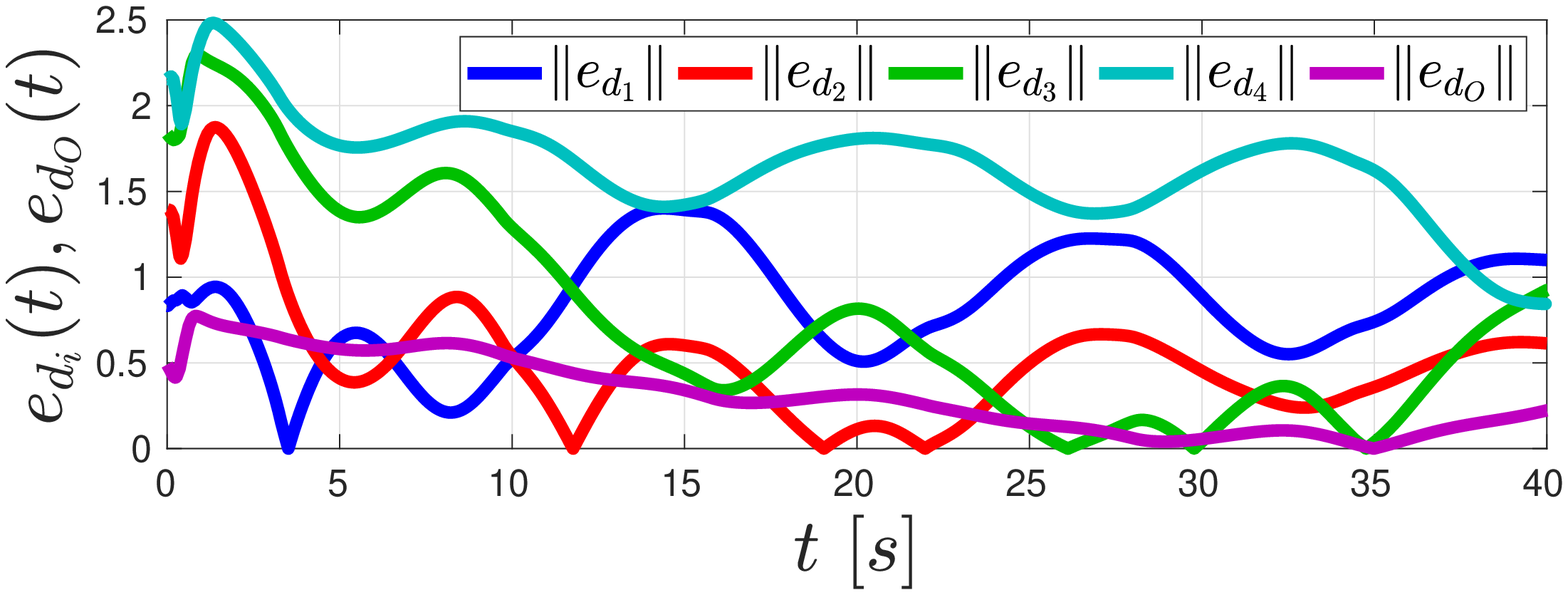}
		\caption{}
		\label{fig:Ng2}
	\end{subfigure}
	
	\caption[]{The adaptation error norms $\|e_{\vartheta_i}(t)\|$, $i\in\mathcal{N}$, $\|e_{\vartheta_{\scr O}}(t)\|$ (a), $\|e_{d_i}(t)\|$, $i\in\mathcal{N}$, $\|e_{d_{\scr O}}(t)\|$ (b), of the control scheme of Section \ref{subsec:Quaternion Controller} $\forall t\in[0,40]$.} \label{fig:adapt_sim_theta_dis_errors}
\end{figure}

{By proceeding similarly as with $\dot{V}_s$, we conclude that 
\begin{align}
\|\varepsilon_v(t) \| \leq \bar{\varepsilon}_v \coloneqq \max\Bigg\{\| \varepsilon_v(0)\|, \frac{\max\limits_{k\in\mathcal{K}}\{\rho_{v_k,0}\}\bar{B}_v}{2g_v\underline{m}} \Bigg\}, \label{eq:epsilon_v_bar}
\end{align}
$\forall t\in[0,\tau_{\max})$, from which we obtain
\small
\begin{align}
-1 < \frac{\exp(-\bar{\varepsilon}_v)-1}{\exp(-\bar{\varepsilon}_v)+1} =: -\bar{\xi}_{v} \leq  \xi_{v_k}(t) \leq \bar{\xi}_{v} \coloneqq \frac{\exp(\bar{\varepsilon}_v)-1}{\exp(\bar{\varepsilon}_v)+1} < 1, \label{eq:ksi_v_bar}
\end{align}
\normalsize
$\forall t\in[0,\tau_{\max})$. 
What remains to be shown is that $\tau_{\max} = \infty$. We can conclude from the aforementioned analysis, Assumption \ref{ass:kinematic singularities}, and \eqref{eq:ksi_s_bar}, \eqref{eq:ksi_v_bar} that the solution $\sigma(t)$ remains in a compact subset $\Omega'$ of $\Omega$, $\forall t\in[0,\tau_{\max})$, namely $\sigma(t) \in \Omega'$,
$\forall t\in[0,\tau_{\max})$. Hence, assuming $\tau_{\max} < \infty$ and since $\Omega'\subset \Omega$, Proposition \ref{Prop:dynamical systems} in Subsection \ref{subsec:Dynamical Systems} dictates the existence of a time instant $t'\in[0,\tau_{\max})$ such that $\sigma(t')\notin\Omega'$, which is a contradiction. Therefore, $\tau_{\max} = \infty$. Thus, all closed loop signals remain bounded and moreover $\sigma(t)\in\Omega' \subset \Omega, \forall t\in\mathbb{R}_{\geq 0}$. Finally, by multiplying \eqref{eq:ksi_s_bar} by $\rho_k(t), k\in\mathcal{K}$, we obtain
\begin{equation}
-\rho_{s_k}(t) < -\bar{\xi}_s\rho_{s_k}(t) \leq e_{s_k}(t) \leq \bar{\xi}_s\rho_{s_k}(t) < \rho_{s_k}(t), \label{eq:ppc_final}
\end{equation}
$\forall t\in\mathbb{R}_{\geq 0}$, which leads to the conclusion of the proof.}
\end{IEEEproof}

\begin{remark} [\textbf{Prescribed Performance}]
From the aforementioned proof it can be deduced that the Prescribed Performance Control scheme achieves its goal without resorting to the need of rendering the ultimate bounds $\bar{\varepsilon}_s,\bar{\varepsilon}_v$ of the modulated pose and velocity errors $\varepsilon_s(t), \varepsilon_v(t)$ arbitrarily small by adopting extreme values of the control gains $g_s$ and $g_v$ (see \eqref{eq:epsilon_s_bar} and \eqref{eq:epsilon_v_bar}). More specifically, notice that \eqref{eq:ksi_s_bar} and \eqref{eq:ksi_v_bar} hold no matter how large the finite bounds $\bar{\varepsilon}_s, \bar{\varepsilon}_v$ are. In the same spirit, large uncertainties involved in the coupled model \eqref{eq:coupled dynamics} can be compensated, as they affect only the size of $\varepsilon_v$ through $\bar{B}_v$, but leave unaltered the achieved stability properties. Hence, the actual performance given in \eqref{eq:ppc_final}, which is solely determined by the designed-specified performance functions $\rho_{s_k}(t), \rho_{v_k}(t), k\in\mathcal{K}$, becomes isolated against model uncertainties, thus extending greatly the robustness of the proposed control scheme.
\end{remark}

\begin{remark}[\textbf{Control Input Bounds}]
The aforementioned analysis of the Prescribed Performance Control methodology reveals the derivation of bounds for the velocity $v_i$ and control input $u_i$ of each agent. {In contrast to our previous work \cite{verginis2017timed}, we derive in Appendix \ref{app: A} explicit bounds $\bar{v}_i$  and $\bar{u}_i$ for $v_i$ and $u_i$ (see \eqref{eq:v_i_bar}, \eqref{eq:u_i_bar}), respectively, which depend on the control gains, the bounds of the dynamic terms, the desired trajectory, and the performance functions.
Therefore, given desired bounds for the agents' velocity $\bar{v}_{i,b}$ and input $\bar{u}_{i,b}$ (derived from bounds on the joint velocities and torques $\dot{q}_i$, $\tau_i$, respectively) and that the upper bounds of the dynamic terms are known, we can tune appropriately the control gain $g_s$, $g_v$ as well as the parameters $\rho_{s_k,0}, \rho_{v_k,0}, \rho_{s_k,\infty}, \rho_{v_k,\infty}, l_{s_k}, l_{v_k}$ in order to achieve $\bar{v}_i \leq \bar{v}_{i,b}, \bar{u}_i \leq \bar{u}_{i,b}, \forall i\in\mathcal{N}$. }
It is also worth noting that the selection of the control gains $g_s, g_v$ affects the evolution of the errors $e_s, e_v$ inside the corresponding performance envelopes.   
\end{remark}

\begin{figure}
	\centering
	\begin{subfigure}[b]{\columnwidth}
		\centering
		\includegraphics[width=.95\columnwidth]{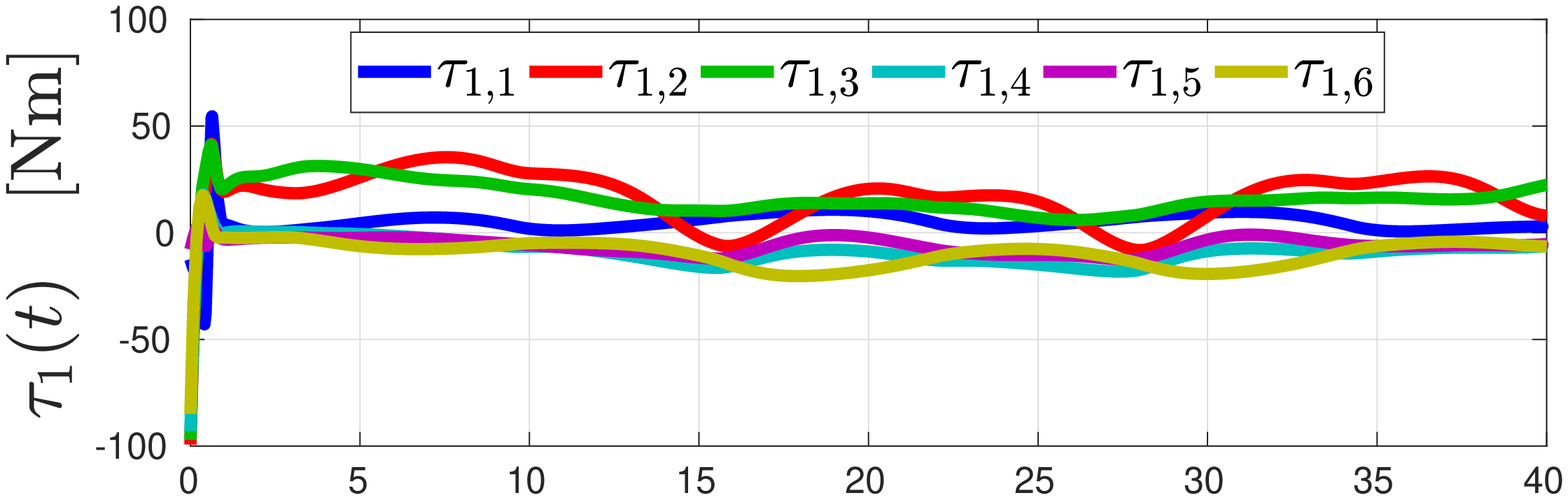}
		\caption{}
		\label{fig:Ng1} 
	\end{subfigure}
	
	\begin{subfigure}[b]{\columnwidth}
		\centering
		\includegraphics[width=.95\columnwidth]{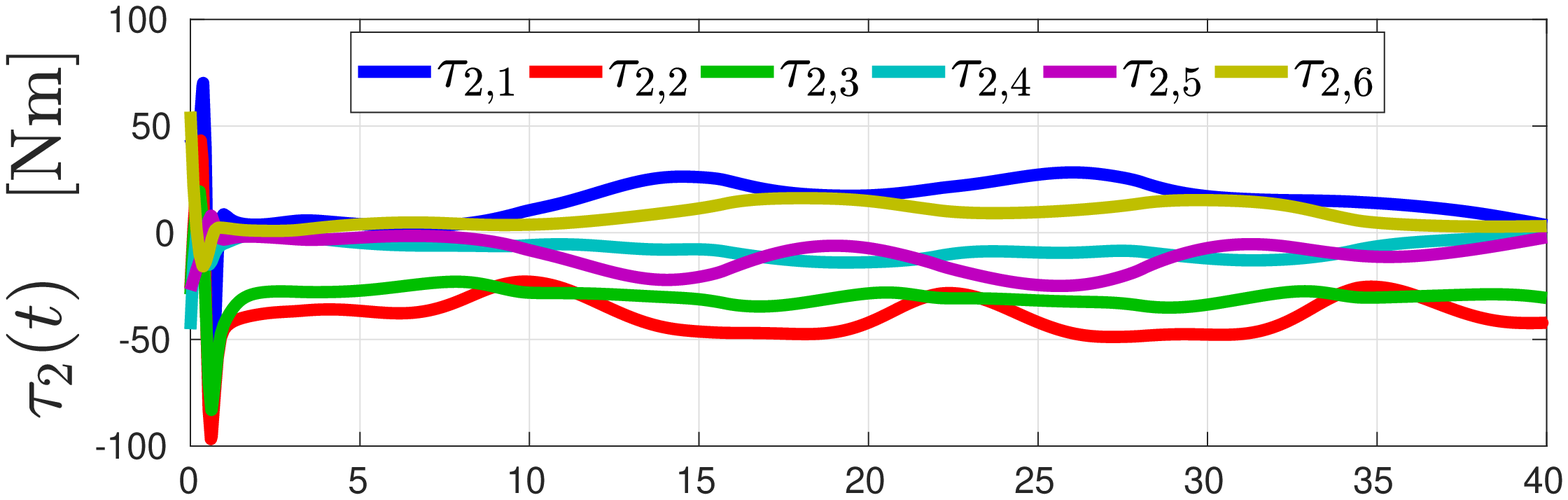}
		\caption{}
		\label{fig:Ng2}
	\end{subfigure}
	
	\begin{subfigure}[b]{\columnwidth}
		\centering
		\includegraphics[width=.95\columnwidth]{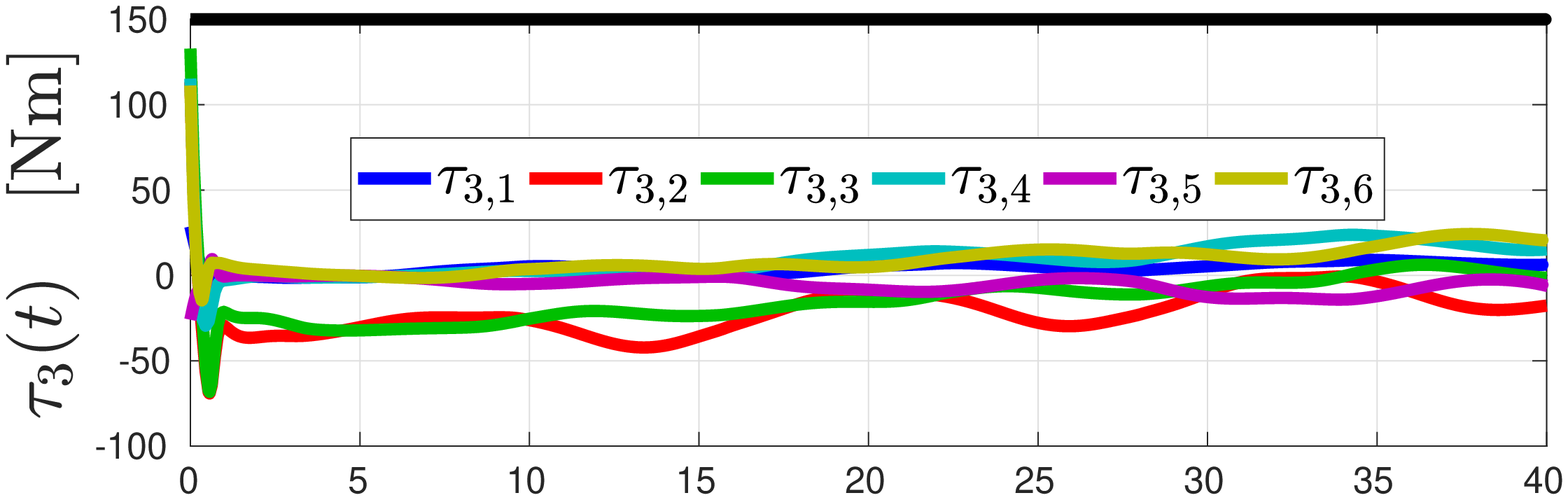}
		\caption{}
		\label{fig:Ng2}
	\end{subfigure}
	
	\begin{subfigure}[b]{\columnwidth}
		\centering
		\includegraphics[width=.95\columnwidth]{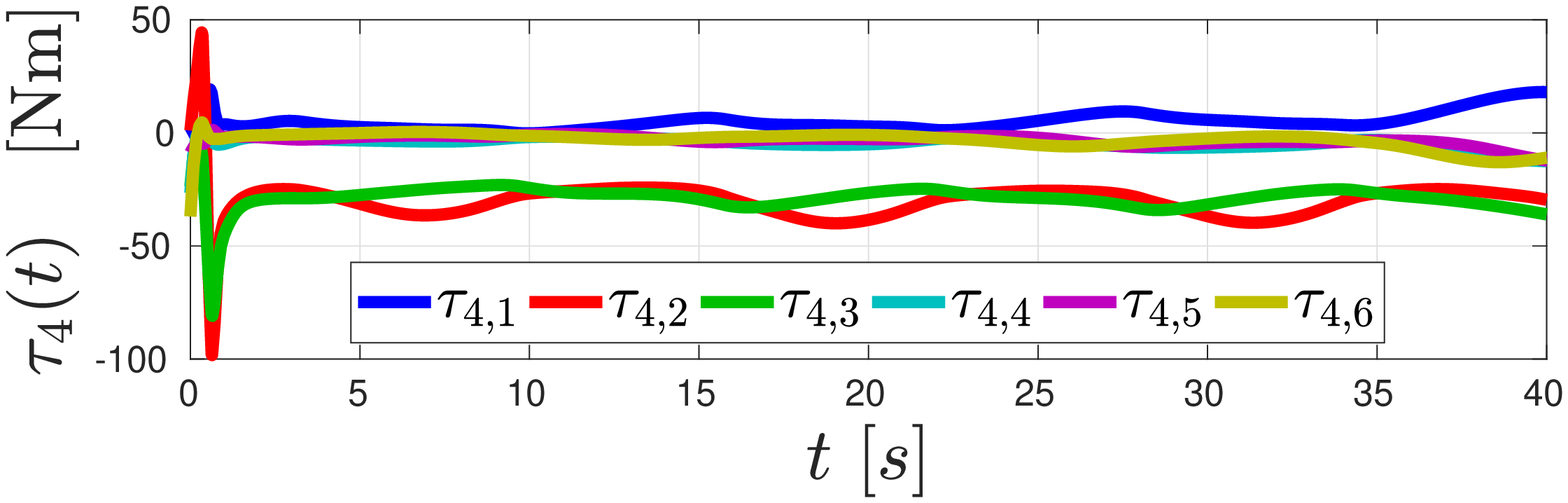}
		\caption{}
		\label{fig:Ng2}
	\end{subfigure}

	\caption[]{The agents' joint torques $\tau_i(t)$, $i\in\mathcal{N}$, (in (a)-(d), respectively) of the control scheme of Section \ref{subsec:Quaternion Controller} $\forall t\in[0,40]$, and the motor saturation (with black), which has not been plotted in (a), (b), (d) for better visualization.} \label{fig:adapt_sim_tau}
\end{figure}

\section{Simulation and Experimental Results} \label{sec:sim and exps}
{In this section, we provide simulation and experimental results for the two developed control schemes. More specifically, Section \ref{subsec:sims} presents computer simulation results and Section \ref{subsec:exps} presents experimental results for both control algorithms.}
\begin{figure*}
	\centering
	\includegraphics[trim = 0cm 0cm 0cm 0cm, width=2\columnwidth]{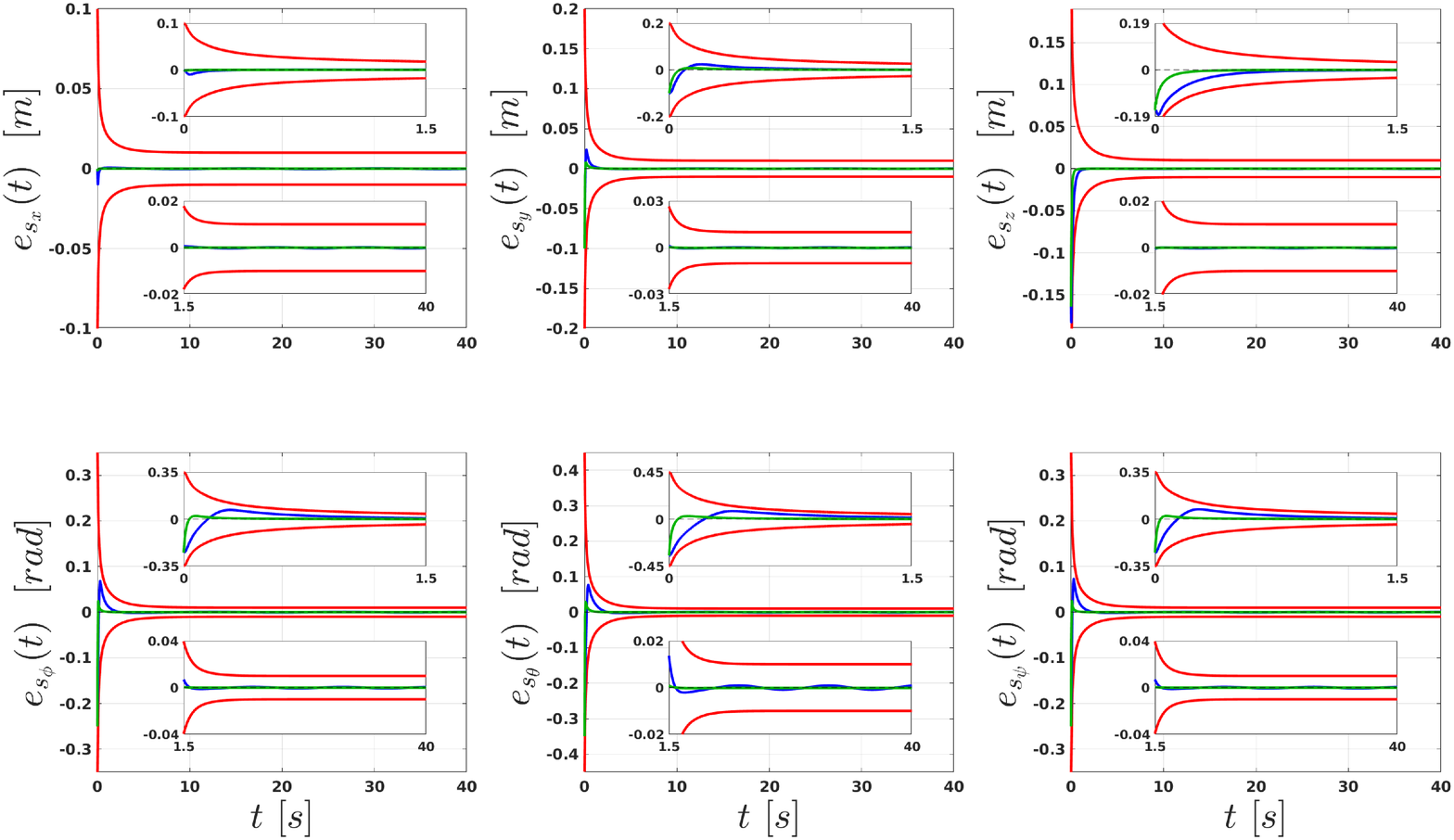} 
	\caption{Simulation results for the controller of Section \ref{subsec:PPC Controller}, {and of \cite{verginis2017timed}}; Top: The position errors $e_{s_x}(t)$, $e_{s_y}(t)$, $e_{s_z}(t)$ (with blue {and green, respectively}) along with the respective performance functions (with red); Bottom: The orientation errors $e_{s_\phi}(t)$, $e_{s_\theta}(t)$, $e_{s_\psi}(t)$ (with blue {and green, respectively}) along with the respective performance functions (with red), $\forall t\in[0,40]$. Zoomed versions of the transient and steady state response have been included for all plots.}
	\label{fig:ppc_pose_errors}
\end{figure*}

\begin{figure}
	\centering
	\includegraphics[width=\columnwidth]{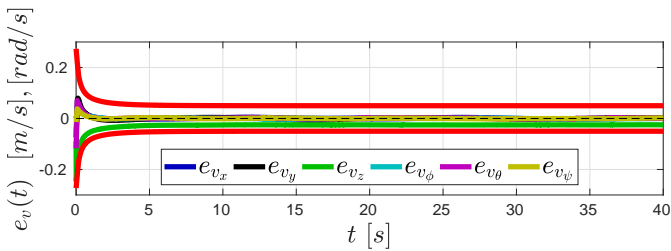} 
	\caption{The velocity errors $e_v(t)$ along with the respective performance functions (with red) for the controller of Section \ref{subsec:PPC Controller}, $\forall t\in[0,40]$.}
	\label{fig:ppc_velocity_errors}
\end{figure}

\subsection{Simulation Results} \label{subsec:sims}
\begin{figure}
	\centering
	\begin{subfigure}[b]{\columnwidth}
		\centering
		\includegraphics[width=.95\columnwidth]{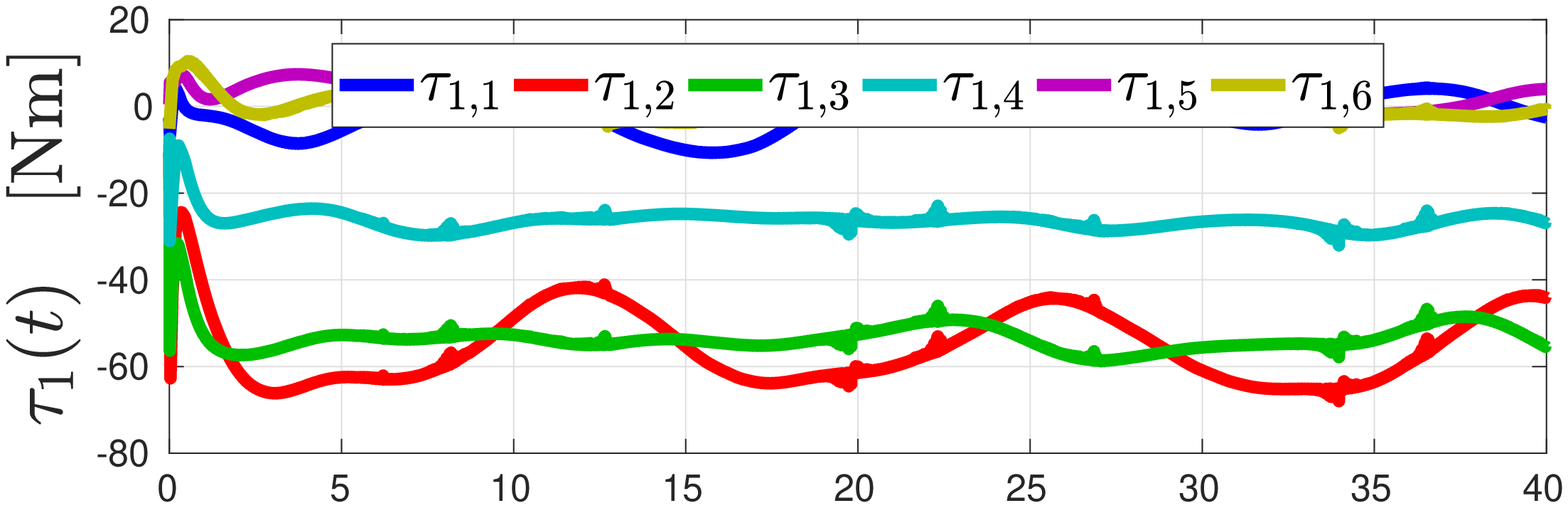}
		\caption{}
		\label{fig:Ng1} 
	\end{subfigure}
	
	\begin{subfigure}[b]{\columnwidth}
		\centering
		\includegraphics[width=.95\columnwidth]{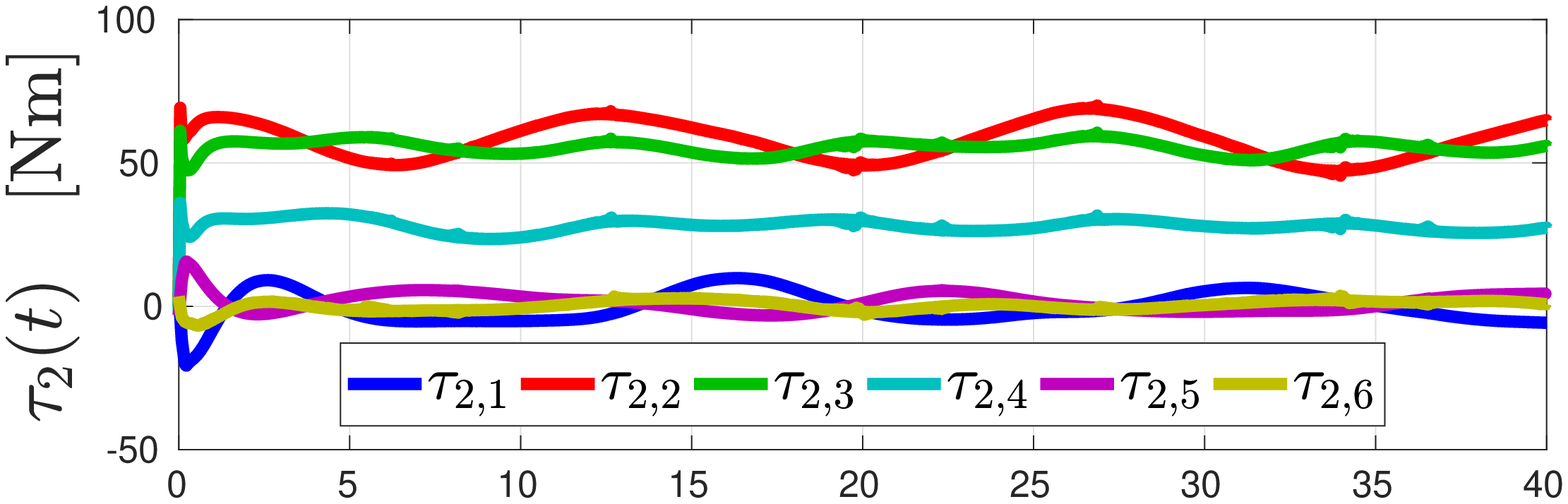}
		\caption{}
		\label{fig:Ng2}
	\end{subfigure}
	
	\begin{subfigure}[b]{\columnwidth}
		\centering
		\includegraphics[width=.95\columnwidth]{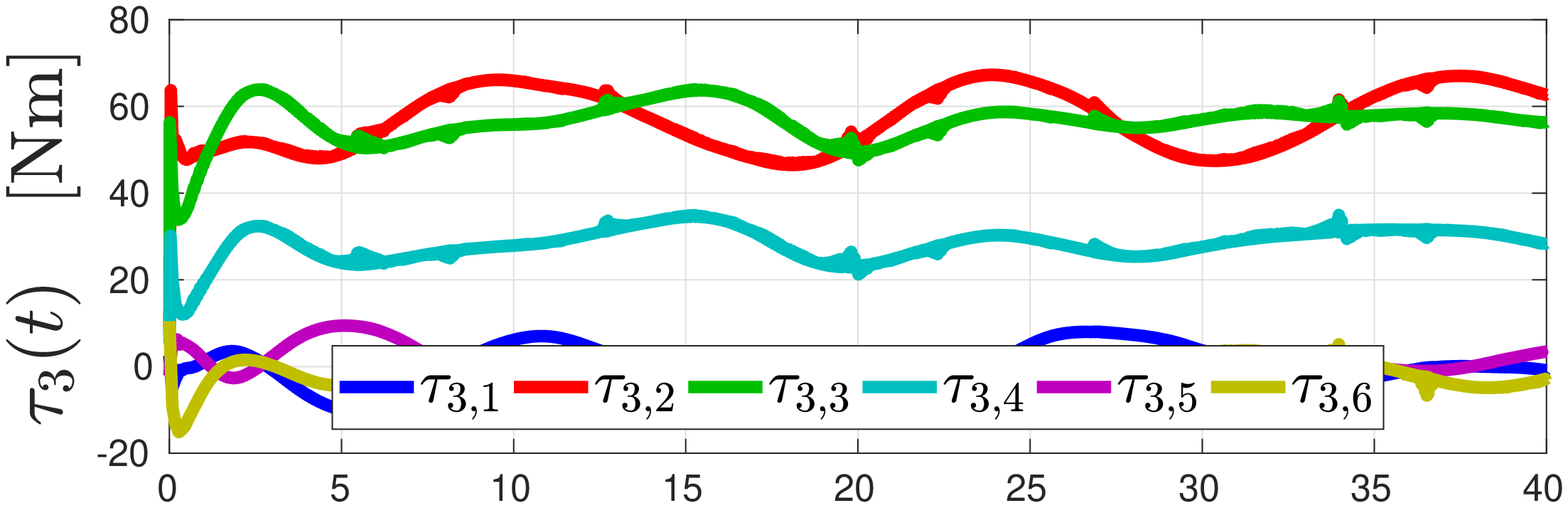}
		\caption{}
		\label{fig:Ng2}
	\end{subfigure}
	
	\begin{subfigure}[b]{\columnwidth}
		\centering
		\includegraphics[width=.95\columnwidth]{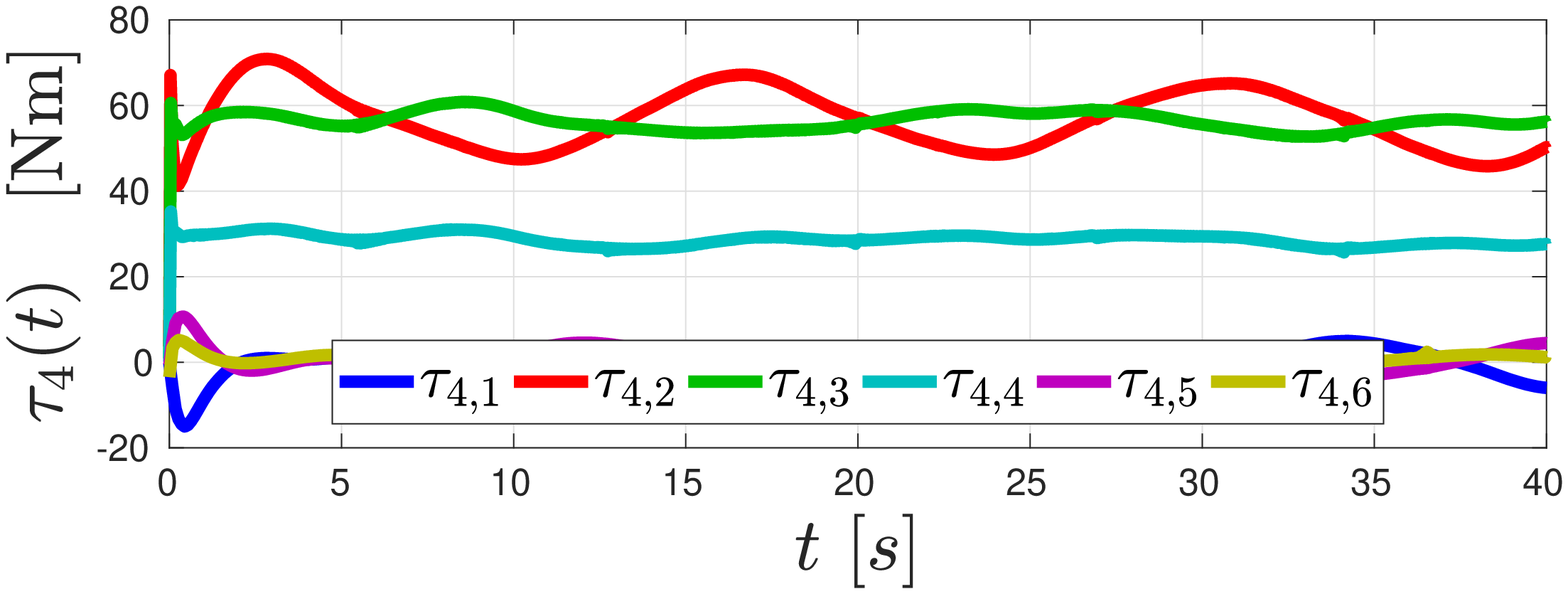}
		\caption{}
		\label{fig:Ng2}
	\end{subfigure}
		
		\caption[]{The agents' joint torques $\tau_i(t)$, $i\in\mathcal{N}$, (in (a)-(d), respectively) of the control scheme of Section \ref{subsec:PPC Controller} $\forall t\in[0,40]$.} \label{fig:ppc_tau}
\end{figure}

\begin{figure}[t]
	\centering
	\includegraphics[width=.95\columnwidth]{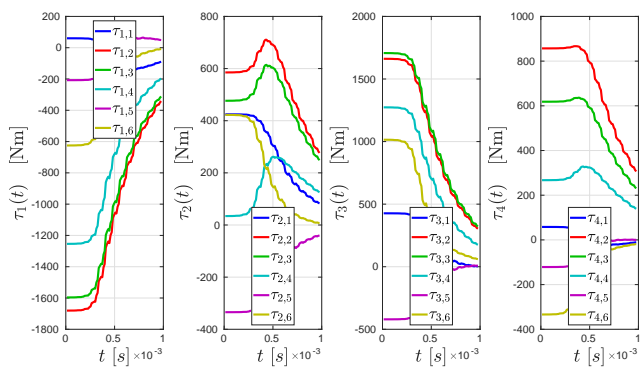} 
	\caption{The agents' joint torques $\tau_i(t)$, $i\in\mathcal{N}$, (in (a)-(d), respectively) of the control scheme of \cite{verginis2017timed} $\forall t\in[0,0.001]$.}
	\label{fig:ppc_tau_comp}
\end{figure}

{
The tested scenario consists of four UR$5$ robotic manipulators rigidly grasping a rectangular object. The object's initial pose is $x_{\scr O}(0) = [-0.225,-0.612,0161,-\pi,\frac{\pi}{3},0]^\top \ ([\text{m}],[\text{rad}])$ with respect to a chosen inertial frame and the desired trajectory is set as $p_\text{d}(t) = [-0.225 + 0.1\sin(0.5t),-0.612 + 0.2\cos(0.5t),0.25 + 0.05\sin(0.5t)]^\top \ [\text{m}]$, $\eta_\text{d}(t)=[-\pi + 0.25\cos(0.5t),\frac{\pi}{3} + A_\theta\sin(0.25t),0.25\cos(0.5t)]^\top \ \text{rad}$,
where $A_\theta$ takes different values for the two control schemes. In particular, we set $A_\theta = \frac{\pi}{6}$ for the adaptive quaternion-feedback control scheme, meaning that the desired pitch angle reaches the configuration of $\frac{\pi}{2}$. This would be singular for the Prescribed Performance Control scheme, for which we set $A_\theta = \frac{\pi}{9}$. In view of Assumption \ref{ass:disturbance bound}, we set $d_i = (\|q_i\|\sin(\omega_{d_i}t + \phi_{d_i}) + \dot{q}_i)\bar{d}_i$ and $d_{\scriptscriptstyle O} = (\|\dot{x}_{\scriptscriptstyle  O}\|\sin(\omega_{d_{\scriptscriptstyle O}}t + \phi_{d_{\scriptscriptstyle O}}) + v_{\scriptscriptstyle O})\bar{d}_{\scriptscriptstyle  O}$, where the constants $\omega_{d_i}, \phi_{d_i}$, $\omega_{d_{\scriptscriptstyle O}}, \omega_{d_{\scriptscriptstyle O}}$ are randomly chosen in the interval $(0,1)$, $\forall i\in\mathcal{N}$. Regarding the force distribution matrix \eqref{eq:J_Hirche}, we set $m_i^\star = 1$, $\forall i\in\mathcal{N}$, and $J_1^\star = 0.6I_3$, $J_2^\star = 0.4I_3$, $J_3^\star = 0.75I_3$, $J_4^\star = 0.25I_3$ to demonstrate a potential difference in the agents' power capabilities. In addition, we set an artificial saturation limit for the joint  motors as $\bar{\tau} = 150 \ \text{Nm}$. 
For the adaptive quaternion-feedback control scheme of Section \ref{subsec:Quaternion Controller}, we set the control gains appearing in \eqref{eq:control laws adaptive quat} and \eqref{eq:adaptation laws} as $k_p = \text{diag}\{[5,5,2]\},$ $k_\zeta = 3I_3$, $K_v = 400I_6$, $\gamma_i = \gamma_{\scr O} = \beta_i = \beta_{\scr O} = 1$, $\forall i\in\mathcal{N}$. The simulation results are depicted in Figs. \ref{fig:adapt_sim_errors}-\ref{fig:adapt_sim_tau} for $t\in[0,40]$ seconds. More specifically, Fig. \ref{fig:adapt_sim_errors} shows the evolution of the pose and velocity errors $e_p(t), e_\zeta(t)$, $e_{v_f}(t)$, Fig. \ref{fig:adapt_sim_theta_dis_errors} depicts the norms of the adaptation errors $e_{\vartheta_i}(t), e_{\vartheta_{\scr O}}(t)$,   $e_{d_i}(t), e_{d_{\scr O}}(t)$, and Fig. \ref{fig:adapt_sim_tau} shows the resulting joint torques $\tau_i(t)$, $\forall i\in\{1,\dots,4\}$. Note that $e_p(t), e_\zeta(t)$ and $e_{v_f}(t)$ converge to the desired values and the adaptation errors are bounded, as predicted by the theoretical analysis.
For the Prescribed Performance Control scheme of Section \ref{subsec:PPC Controller}, we set the performance functions as $\rho_{s_k}(t)= (|e_{s_k}(0)| + 0.09)\exp(-0.5t)+0.01$, $\rho_{v_k}(t) =(|e_{v_k}(0)| + 0.95)\exp(-0.5t)+0.05$, $\forall k\in\mathcal{K}$, and the control gains of \eqref{eq:v_r}, \eqref{eq:control_law_ppc} as $g_s = 0.005$, $g_v = 10$, respectively, by following Appendix \ref{app: A} and considering known dynamic bounds. The simulation results are depicted in Figs. \ref{fig:ppc_pose_errors}-\ref{fig:ppc_tau}, for $t\in[0,40]$ seconds. In particular, Fig. \ref{fig:ppc_pose_errors} depicts the evolution of the pose errors $e_s(t)$ {(in blue)}, along with the respective performance functions $\rho_s(t)$ (in red), Fig. \ref{fig:ppc_velocity_errors} depicts the evolution of the velocity errors $e_v(t)$, along with the respective performance functions $\rho_v(t)$, and Fig. \ref{fig:ppc_tau} shows the resulting joint torques $\tau_i(t)$, $\forall i\in\{1,\dots,4\}$.

 One can conclude from the aforementioned figures that the simulation results verify the theoretical findings, since the errors $e_s(t)$, $e_v(t)$ stay confined in the performance function funnels. Moreover, the joint torques in both control schemes respect the saturation values we set. {For comparison purposes, we also simulate the same system by using the Prescribed Performance Control methodology of \cite{verginis2017timed}, without taking into account any input constraints, since the input constraint analysis of Appendix \ref{app: A} is not performed in \cite{verginis2017timed}. In order to achieve good performance in terms of overshoot, rise, and settling time, we set the control gains as $g_s = 1$, $g_v = 200$. The resulting pose errors are depicted in Fig. \ref{fig:ppc_pose_errors} for $t\in[0,40]$ seconds (with green) along with the performance functions (with red), and the resulting torques are depicted in Fig. \ref{fig:ppc_tau_comp} for $t\in[0,0.001]$ seconds. This small time interval is sufficient to observe the high-value initial peaks of the torque inputs that do not satisfy the desired constraint of  $\bar{\tau} = 150 \ \text{Nm}$, which can be attributed to the lack of gain calibration. Nevertheless, note also the better performance of the pose errors, in terms of overshoot, rise and settling time, as pictured in Fig. \ref{fig:ppc_pose_errors}. Finally, note that any Prescribed Performance Control methodology would fail to solve Problem \ref{prob:problem1} with $\theta(0) = \frac{\pi}{2}$ or $\theta_d(t) = \frac{\pi}{2}$ for some $t\in\mathbb{R}_{\geq 0}$, in contrast to the adaptive quaternion-feedback control scheme of Section \ref{subsec:Quaternion Controller}. The torque illustration for the remaining time as well as the velocity error convergence are omitted due to space constraints.}
The simulations were carried out in the MATLAB R2017a environment on a $i7$-$5600$ laptop computer at $2.6$Hz, with $8$gB of RAM.}

\subsection{Experimental Results} \label{subsec:exps}

\begin{figure}[t]
	\centering
	\begin{subfigure}[b]{\columnwidth}
		\centering
		\includegraphics[trim =0cm 0cm 0cm 0cm, width=.95\columnwidth]{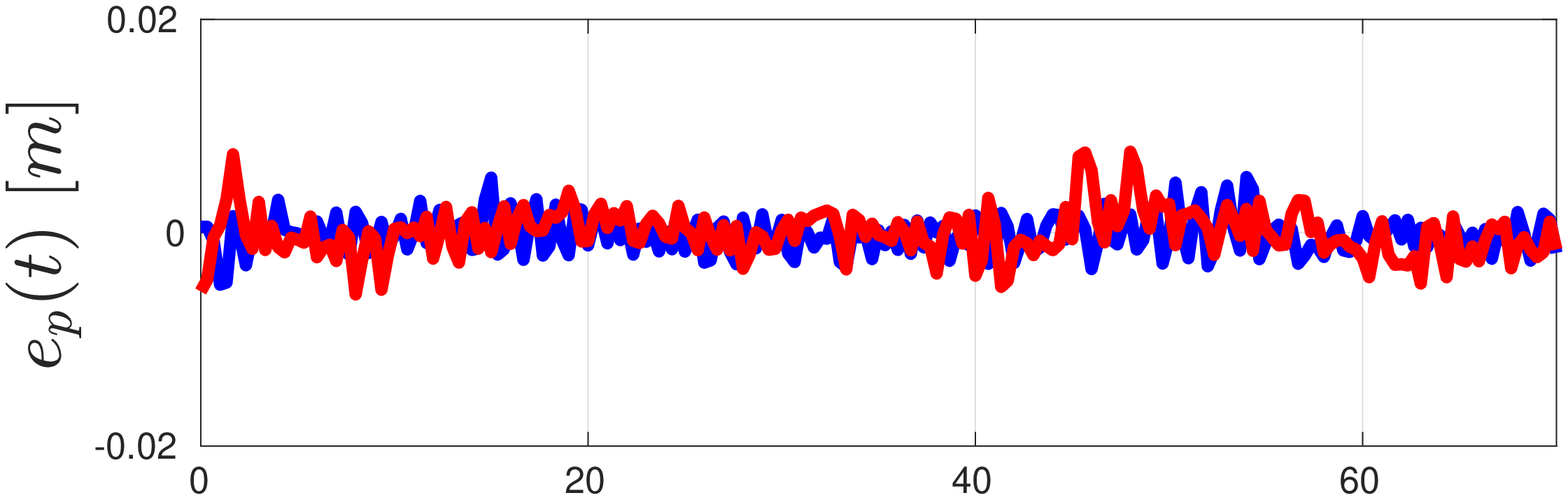}
		\caption{}
		\label{fig:Ng1} 
	\end{subfigure}
	
	\begin{subfigure}[b]{\columnwidth}
		\centering
		\includegraphics[width=.95\columnwidth]{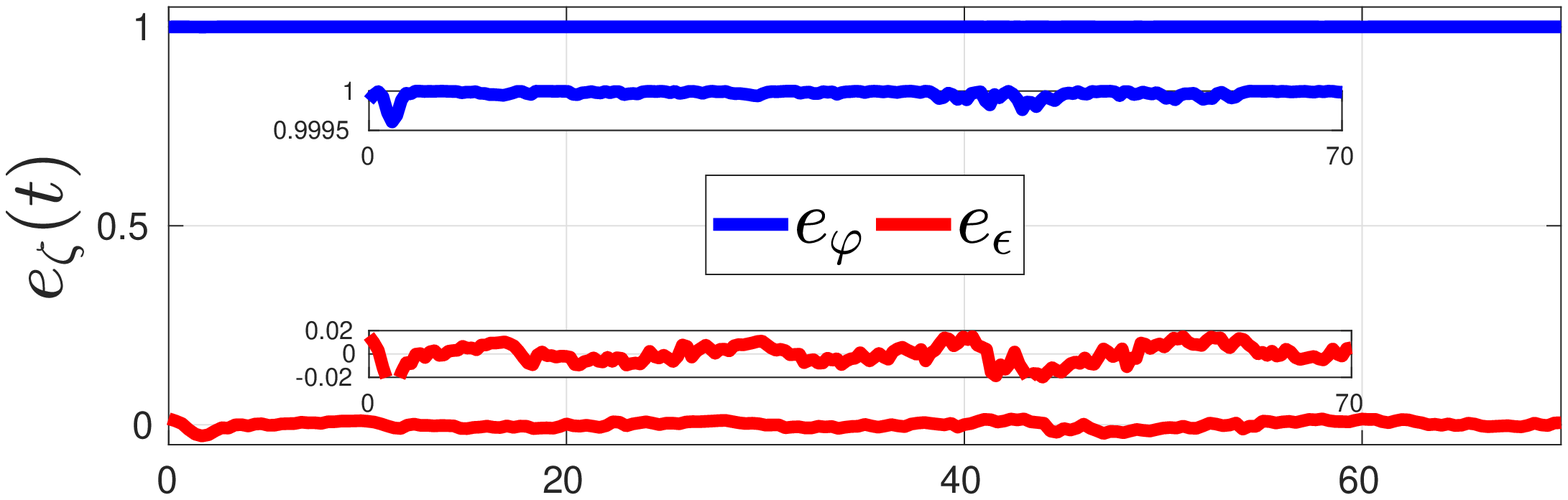}
		\caption{}
		\label{fig:Ng2}
	\end{subfigure}
	
	\begin{subfigure}[b]{\columnwidth}
		\centering
		\includegraphics[width=.95\columnwidth]{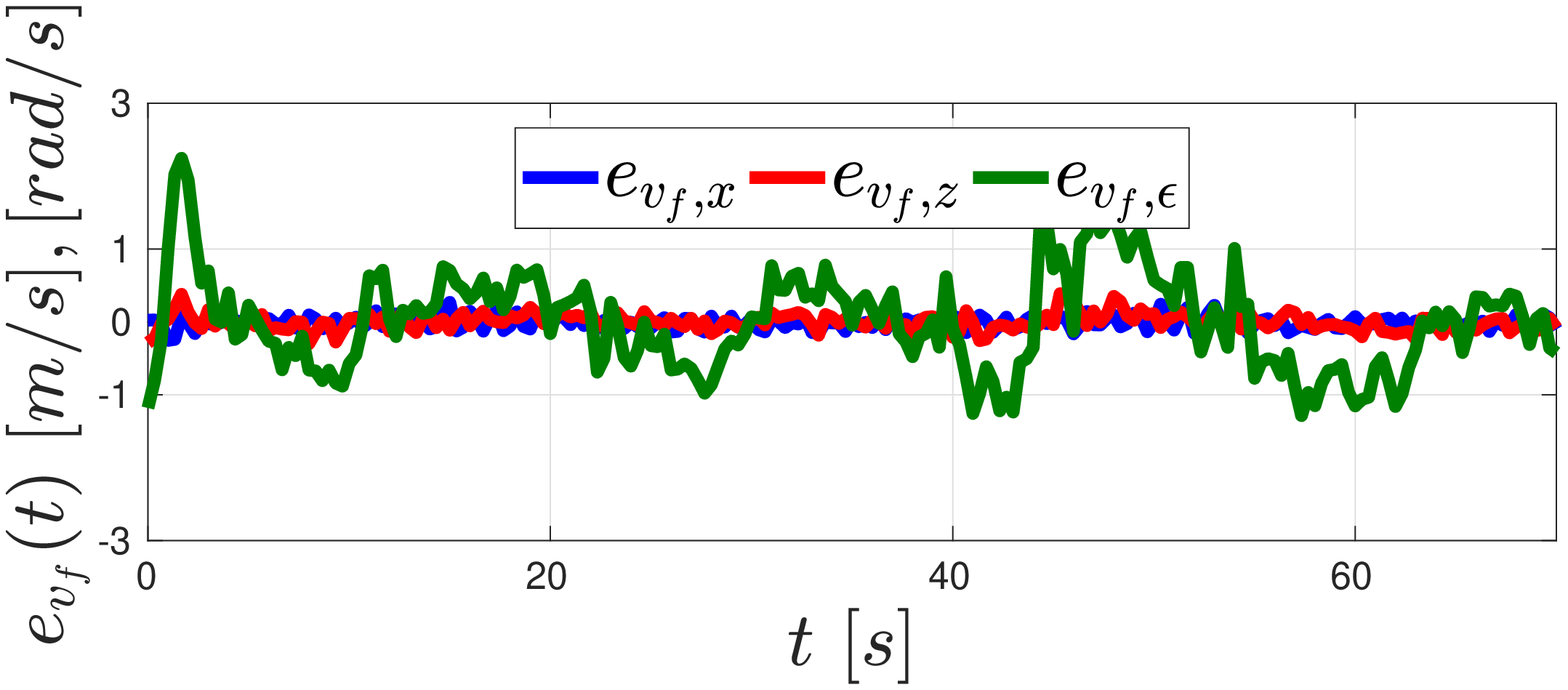}
		\caption{}
		\label{fig:Ng2}
	\end{subfigure}
	
	\caption[]{Experimental results for the control scheme of Section \ref{subsec:Quaternion Controller}; (a): The position errors $e_p(t)$; (b): The quaternion errors $e_\varphi(t)$, $e_\varepsilon(t)$; (c) The velocity errors $e_{v_f}(t)$, $\forall t\in[0,70]$.} \label{fig:adapt_quat_errors}
\end{figure}

\begin{figure}
	\centering
	\includegraphics[width=.95\columnwidth]{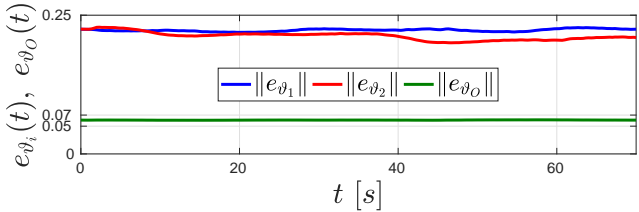} 
	\caption{The norms of the adaptation signals $e_{\vartheta_i}(t), \forall i\in\{1,2\}$ (left) and $e_{\vartheta_{\scriptscriptstyle O}}(t)$, (right) $\forall t\in[0,70]$ of the experiment of the controller in Section \ref{subsec:Quaternion Controller}.}
	\label{fig:adaptive_exp_theta_adaptation_signals}
\end{figure}

\begin{figure}
	\centering
	\includegraphics[width=.9\columnwidth]{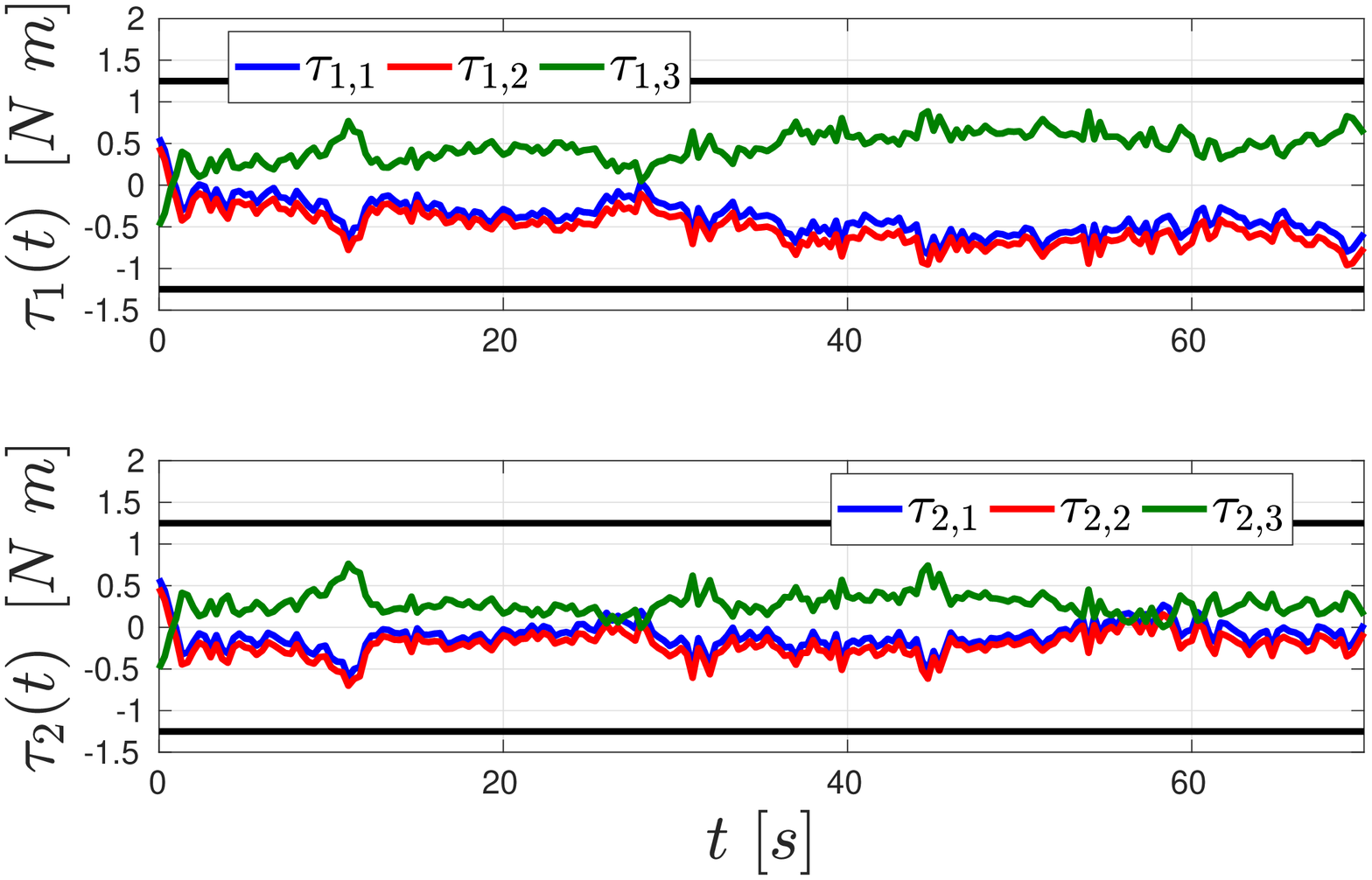} 
	\caption{The agents' joint torques of the experiment of the controller in Section \ref{subsec:Quaternion Controller}, for $t\in[0,70]$, with their respective limits (with black).}
	\label{fig:adaptive_exp_theta_inputs}
\end{figure}

\begin{figure*}
	\centering
	\includegraphics[width=1.7\columnwidth]{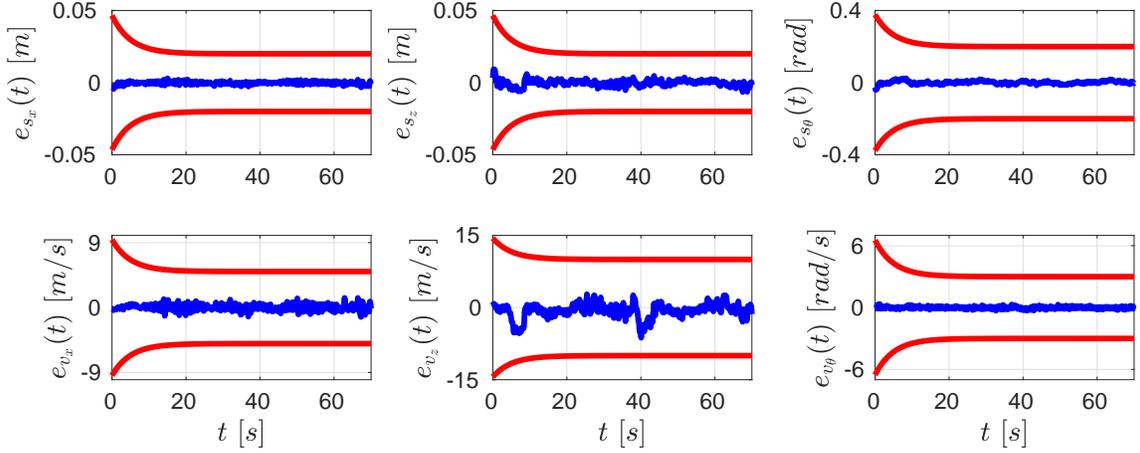} 
	\caption{Experimental results for the controller of Section \ref{subsec:PPC Controller}; Top: the pose errors $e_{s_x}(t)$, $e_{s_z}(t)$, $e_{s_\theta}(t)$ (with blue) along with the respective performance functions (with red); Bottom: The velocity errors $e_{v_x}(t)$, $e_{v_z}(t)$, $e_{v_\theta}(t)$ (with blue) along with the respective performance functions (with red), $\forall t\in[0,70]$.}
	\label{fig:ppc_exp_theta_errors}
\end{figure*}

\begin{figure}
	\centering
	\includegraphics[width=.9\columnwidth]{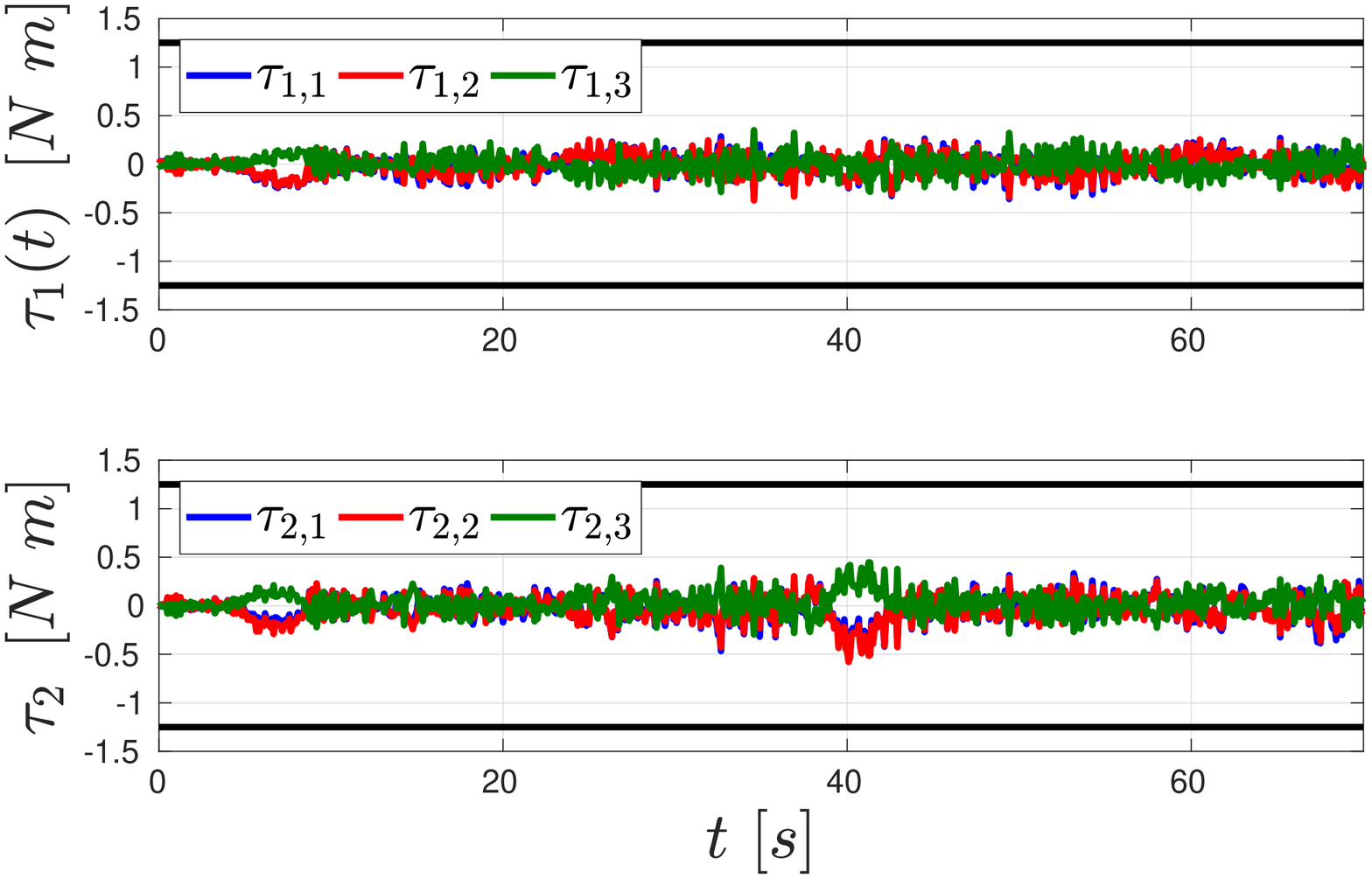} 
	\caption{The agents' joint torques of the experiment of the controller in Section \ref{subsec:PPC Controller}, $\forall t\in[0,70]$, with their respective limits (with black).}
	\label{fig:ppc_exp_theta_inputs}
\end{figure}

The tested scenario for the experimental setup consists of two WidowX Robot Arms rigidly grasping a wooden cuboid object of initial pose $x_{\scr O}(0) = [0.3, 0, 0.15, 0, 0, 0]^\top \ ([\text{m}],[\text{rad}]) $, which has to track a planar time trajectory $p_{\text{d}}(t)= [0.3 + 0.05\sin(\frac{2\pi t}{35}), 0.15 - 0.05\cos(\frac{2\pi t}{35})]^\top \ [\text{m}]$, $\eta_\text{d}(t) 
= \frac{\pi}{20}\sin(\frac{5\pi t}{35}) \ [\text{rad}]$.
For that purpose, we employ the three rotational -with respect to the $y$ axis - joints of the arms. The lower joint consists of a MX-$64$ Dynamixel Actuator, whereas each of the two upper joints consists of a MX-$28$ Dynamixel Actuator from the MX Series. Both actuators provide feedback of the joint angle and rate $q_i, \dot{q}_i$, $\forall i\in\{1,2\}$.
The micro-controller used for the actuators of each arm is the ArbotiX-M Robocontroller, which is serially connected to an i-$7$ desktop computer  with $4$ cores and $16$GB RAM. All the computations for the real-time experiments are performed at a frequency of $120$ [Hz]. Finally, we consider that the MX-$64$ motor can exert a maximum torque of $3$ [Nm], and the MX-$28$ motors can exert a maximum torque of $1.25$ [Nm], values that are slightly more conservative than the actual limits.  
The load distribution coefficients are set as $m_1^\star = m_2^\star = 1$, and $J^\star_1 = 0.75I_3$, $J^\star_2 = 0.25I_3$. 
For the adaptive quaternion-feedback control scheme, {we set $\delta_{\scriptscriptstyle O}(x_{\scriptscriptstyle O},\dot{x}_{\scriptscriptstyle O}, t)=0_{6\times\mu_{\scr O}}$, $\delta_i(q_i,\dot{q}_i,t) = 0_{6\times\mu}$, $\forall i\in\mathcal{N}$, which essentially means that we do not model any external disturbances}. We also set the control gains appearing in  \eqref{eq:control laws adaptive quat} and \eqref{eq:adaptation laws} as $k_p = 50$, $k_\zeta = 80$, $K_v = \text{diag}\{3.5,0.5,0.5\}$. 
The experimental results are depicted in Fig. \ref{fig:adapt_quat_errors}-\ref{fig:adaptive_exp_theta_inputs} for $t\in[0,70]$ seconds. More specifically, Fig. \ref{fig:adapt_quat_errors} pictures the pose and velocity errors $e_p(t), e_\zeta(t), e_{v_f}(t)$, Fig. \ref{fig:adaptive_exp_theta_adaptation_signals} depicts the norms of the adaptation errors $e_{\vartheta_i}(t)$, $e_{\vartheta_{\scr O}}(t)$, and Fig. \ref{fig:adaptive_exp_theta_inputs} shows the joint torques  $\tau_1(t)$, $\tau_2(t)$ of the agents. {Although external disturbances and modeling uncertainties are not taken into account in the system model, they are indeed present during the experiment run time and one can observe that the errors converge to the desired values and the adaptation errors remain
bounded, verifying the theoretical findings.}
For the Prescribed Performance Control scheme, we set the performance functions as  $\rho_{s_{x}}(t) = \rho_{s_{z}}(t) = 0.03\exp(-0.2t) + 0.02$ [m], $\rho_{s_{\theta}}(t) = 0.2\exp(-0.2t) + 0.2$ [rad], $\rho_{v_x}(t) = 5\exp(-0.2t) + 5$ [m/s], $\rho_{v_z}(t) = 5\exp(-0.2t) + 10$ [m/s], and $\rho_{v_{\theta}}(t) = 4\exp(-0.2t) + 3$ [m/s], and the control gains of \eqref{eq:v_r} and \eqref{eq:control_law_ppc} as $g_s = 0.05$ and $g_v=10$, respectively, by following Appendix \ref{app: A}. The experimental results are depicted in Fig. \ref{fig:ppc_exp_theta_errors}-\ref{fig:ppc_exp_theta_inputs} for $t\in[0,70]$ seconds. In particular, Fig. \ref{fig:ppc_exp_theta_errors} shows the pose and velocity errors $e_s(t)$, $e_v(t)$ along with the respective performance functions, and Fig. \ref{fig:ppc_exp_theta_inputs} depicts the joint torques $\tau_1(t)$, $\tau_2(t)$ of the agents.
We can conclude that the experimental results verify the theoretical analysis, since the errors evolve strictly within the prespecified performance bounds. Note also that in both control schemes
the joint torques respect the saturation limits. {A video illustrating the results can be found on https://youtu.be/jJWeI5ZvQPY.}

\subsection{Discussion} \label{subsec:Dissusion}
{ In view of the aforementioned results, we mention some worth-noting differences between the two control schemes. Firstly, note that the PPC methodology allows for \textit{exponential} convergence of the errors to the set defined by the values $\rho_{s_k,\infty}$, $\rho_{v_k,\infty}$, achieving \textit{predefined} transient and steady-state performance, without the need to resort to tuning of the control gains. The adaptive quaternion-feedback methodology, however, can only guarantee that the errors converge to zero as $t\to\infty$. This is verified by the simulation results, where the error trajectories $e_p(t), e_\zeta(t)$ and $e_v(t)$ show an oscillatory behavior. Improvement of such performance (in terms of overshoot, rise, and settling time) would require appropriate gain tuning. 
Secondly, note that, as shown in the simulations section, the quaternion-feedback methodology allows for trajectories where the pitch angle of the object $(\theta_{\scriptscriptstyle O})$ can be $\pm 90$ degrees, in contrast to the PPC methodology, where that configuration is ill-posed, since the matrix $J_{\scriptscriptstyle O}(\eta_{\scriptscriptstyle O})$ is not defined. Finally, the adaptive quaternion-feedback methodology can be considered less robust to modeling uncertainties in real-time scenarios, since it accounts only for parametric uncertaintes (the unknown terms $\theta_i$, $\theta_{\scriptscriptstyle O}$, $d_i$, $d_{\scriptscriptstyle O}$), assuming a \textit{known} structure of the dynamic terms. The PPC methodology, however, does not require any information of the structure or the parameters of the dynamic model (note that the only requirements are the positive definiteness of the coupled inertia matrix, the locally Lipschitz and continuity properties of the dynamic terms and the boundedness - with respect to time -  of the disturbances $d_i, d_{\scriptscriptstyle O}$). In that sense, one would expect the PPC methodology to perform better in real-time experiments, where unmodeled dynamics are involved. The fact, however, that PPC is a control scheme that does not contain any information of the model structure makes it more difficult to tune (in terms of gain tuning) in order to achieve robot velocities and torques that respect specific bounds, especially when the bounds  of the dynamic terms are unknown. This has been noticed during both simulations and experiments.}

\section{Conclusion and Future Work} \label{sec:Conclusion and FW}
We presented two novel decentralized control protocols for the cooperative manipulation of a single object by $N$ robotics agents. Firstly, we developed a quaternion-based approach that avoids representation singularities with adaptation laws to compensate for dynamic uncertainties. Secondly, we developed a robust control law that guarantees prescribed performance for the transient and steady state of the object. Both methodologies were validated via realistic simulations and experimental results. Future efforts will be devoted towards applying the proposed techniques to cases with non rigid grasping points and uncertain object geometric characteristics.


%





\appendices
{\section{} \label{app: A}
	In the following, we derive explicit expressions for the terms $f_{\text{cl},q}$, $f_{\text{cl},s}$, $f_{\text{cl},v}$ of \eqref{eq:sigma_dot_1}, as well as bounds for the dynamics terms of the model and the velocity and control inputs $v_i$, $u_i$, respectively, $i\in\mathcal{N}$}.

{Note first from  \eqref{eq:ppc errors}, \eqref{eq:ksi_s}, \eqref{eq:e_v_r}, and \eqref{eq:ksi_v}, that the states $x_{\scriptscriptstyle O}, v_{\scriptscriptstyle O}$ can be expressed as 
	\begin{subequations} \label{eq:x_o, x_o_dot}
		\begin{align}
		x_{\scriptscriptstyle O} &= x_\text{d}(t) + \rho_s(t)\xi_s, \\
		v_{\scriptscriptstyle O} &= \rho_v(t)\xi_v + v_r(\xi_s,t),
		\end{align}
	\end{subequations}
	where, with a slight abuse of notation, we right $v_r$ as a function of $\xi_s$ and $t$.
	Then from \eqref{eq:sigma_dot_1} and \eqref{eq:v_r} we obtain: 
	\begin{align} \label{eq:q_dot_cl}
	f_{\text{cl},q}(\sigma,t) = \widetilde{J}(q) G(q)\Big( \rho_v(t)\xi_v + v_r(\xi_s,t) \Big)
	\end{align}	 
	Regarding $f_{\text{cl},s}$,  we obtain from \eqref{eq:sigma_dot_1} by using $\eqref{eq:object dynamics 1}$ and \eqref{eq:x_o, x_o_dot}:
	\begin{align}
	f_{\text{cl},s}(\sigma,t) =& \rho(t)^{-1}\Big[ J_{\scriptscriptstyle O}\Big(\eta_\text{d}(t)+\rho_{s_\eta}(t)\xi_{s_\eta}\Big)\rho_v(t)\xi_v - \dot{\rho}_s(t) \xi_s \notag \\ 
	& - g_s\rho(t)^{-1}r_s(\xi_s)\varepsilon_s(\xi_s) - \dot{x}_\text{d}(t) \Big], \label{eq:ksi_s_dot_1} 
	\end{align}
	where we also express $\varepsilon_s$, from \eqref{eq:epsilon_s}, as a function of $\xi_s$.}

{Next, we differentiate $v_r$ from \eqref{eq:v_r} and use \eqref{eq:x_o, x_o_dot}, \eqref{eq:ksi_s}, \eqref{eq:r_s}, to obtain:
	\begin{align}
	& \dot{v}_r = -g_s J_{\scriptscriptstyle O}\Big(\eta_\text{d}(t) + \rho_{s_\eta}(t)\xi_s\Big)^{-1}\Big[ \rho_s(t)^{-1}\dot{r}_s(\xi_s)\varepsilon_s  \notag \\
	&\hspace{2mm}+ \rho_s(t)^{-1} r_s(\xi_s)^2f_{\text{cl},s}(\sigma,t) - \rho_s(t)^{-2}\dot{\rho}_s(t)r_s(\xi_s)\varepsilon_s  \Big] \notag \\ 
	&\hspace{7mm} - g_s\frac{\partial }{\partial t}\Big[J_{\scriptscriptstyle O}(\eta_{\scr O} )^{-1}\Big]\rho_s(t)^{-1}r_s(\xi_s)\varepsilon_s(\xi_s), \label{eq:v^d_r2} 
	\end{align}
	where 
	\begin{align}
	\dot{r}_s(\xi_s) = 
	\text{diag}\Big\{ \Big[ \frac{2\xi_{s_k}}{(1-\xi_{s_k})^2}\Big]_{k\in\mathcal{K}} \Big\} \sum\limits_{k\in\mathcal{K}} \bar{E}_k f_{\text{cl},s}(\sigma,t) \bar{e}_k, \label{eq:r_s_dot}
	\end{align}
	with $\bar{E}_k\in \mathbb{R}^{6\times 6}$ being the matrix with $1$ in the element $(k,k)$ and zeros everywhere else, and $\bar{e}_k\in \mathbb{R}^6$ being the vector with $1$ in the element $k$ and zeros everywhere else. Note from \eqref{eq:v^d_r2}, \eqref{eq:x_o, x_o_dot}, and the fact that $\dot{x}_{\scr O} = J_{\scr O}(\eta_\text{d}(t) + \rho_{s_\eta}(t)\xi_{s_\eta})v_{\scr O}$  that $\dot{v}_r$ can be expressed as a function of $\sigma$ and $t$.
	Hence, in view of \eqref{eq:coupled dynamics}, \eqref{eq:control_law_ppc_vector_form}, and $[G(q)]^\top G^{+}_M = I_6$, one obtains from \eqref{eq:sigma_dot_1} 
	\begin{align}
	\dot{\xi}_v =& \rho_v(t)^{-1}\Big( -  \dot{\rho}_v(t) \xi_v - \widetilde{M}(x(\sigma,t))\Big[\widetilde{C}(x(\sigma,t))[\rho_v(t)\xi_v + \notag \\  & 
	v_r(\xi_s,t)] + \widetilde{g}(x(\sigma,t)) + \widetilde{d}(x(\sigma,t),t) - \notag \\
	&\hspace{-6mm} g_v \rho_v(t)^{-1}r_v(\xi_v)\varepsilon_v(\xi_v) \Big] -  \dot{v}_{r}(\sigma,t)\Big) =: f_{\text{cl},v}(\sigma,t) \label{eq:ksi_v_dot_1}
	\end{align}
	and where, by using \eqref{eq:x_o, x_o_dot} and \eqref{eq:q_dot_cl}, we have written $x$ (that was first defined in \eqref{eq:coupled dynamics}) as a function of $\sigma$ and $t$, i.e., 
	\begin{equation}
	x(\sigma,t) = 
	\begin{bmatrix}
	q \\ 
	\dot{q} \\
	x_{\scriptscriptstyle O} \\
	\dot{x}_{\scriptscriptstyle O}
	\end{bmatrix}
	=
	\begin{bmatrix}
	q \\
	f_{\text{cl},q}(\sigma,t) \\
	x_\text{d}(t) + \rho_s(t)\xi_s \\
	J_{\scriptscriptstyle O}\Big(\eta_\text{d}(t)+\rho_\eta(t)\xi_{s_\eta}\Big)[\rho_v(t)\xi_v + v_r(\xi_s,t)]
	\end{bmatrix}. \notag
	\end{equation}
	We proceed by deriving expressions for the bounds of the agent velocities and control inputs $v_i$, $u_i$, $i\in\mathcal{N}$. By inspecting \eqref{eq:V_s_dot for Bs} and \eqref{eq:V_s_dot for Bs_2} we can conclude that $\bar{B}_s\coloneqq \sqrt{6}\bar{J}_{\scriptscriptstyle O}(\|v_r(0)\| + \alpha) + \bar{\dot{x}}_\text{d} + \sqrt{6}\max\limits_{k\in\mathcal{K}}\{l_k(\rho_{s_k,0} - \rho_{s_k,\infty})\}$ where $\bar{\dot{x}}_\text{d}$ is the bound of $\dot{x}_\text{d}(t)$. Moreover, in view of \eqref{eq:r_s}, \eqref{eq:ksi_s_bar}, one obtains $\|r_s(\xi_s(t))\| \leq \bar{r}_s \coloneqq \frac{2}{1-\bar{\xi}^2_{s}} = \frac{(\exp(\bar{\varepsilon}_s)+1)^2}{2\exp(\bar{\varepsilon}_s)}$. Therefore, we obtain from \eqref{eq:v_r} 
	\begin{equation}
	\|v_r(t)\| \leq  \bar{v}_r \coloneqq g_s\sqrt{2}\frac{\bar{\varepsilon}_s(\exp(\bar{\varepsilon}_s)+1)^2}{2\min\limits_{k\in\mathcal{K}}\{\rho_{s_k,\infty}\}\exp(\bar{\varepsilon}_s)}. \notag
	\end{equation}
	From $v_{\scriptscriptstyle O} = v_r + \rho_v(t)\xi_v$ we also conclude 
	\begin{equation}
	\| v_{\scriptscriptstyle O}(t) \| \leq \bar{v}_{\scriptscriptstyle O} \coloneqq  g_s \sqrt{2}\frac{\bar{\varepsilon}_s(\exp(\bar{\varepsilon}_s)+1)^2}{2\min\limits_{k\in\mathcal{K}}\{\rho_{k,\infty}\}\exp(\bar{\varepsilon}_s)} + \sqrt{6}\max\limits_{k\in\mathcal{K}}\{\rho_{v_k,0}\}, \notag
	\end{equation}
	which, through \eqref{eq:J_o_i} and \eqref{eq:J_O_i bound}, leads to 
	\begin{equation}
	\| v_i(t) \| \leq \bar{v}_i \coloneqq (\| p^{\scriptscriptstyle E_i}_{\scriptscriptstyle O/E_i} \| + 1)\bar{v}_{\scriptscriptstyle O}, \forall i\in\mathcal{N}. \label{eq:v_i_bar}
	\end{equation}
	By considering the derivative of the reference velocity \eqref{eq:v^d_r2}, as well as \eqref{eq:epsilon_s}, \eqref{eq:ksi_s_dot_1}, \eqref{eq:r_s_dot}, and \eqref{eq:v_i_bar} we can obtain a bound  $\|\dot{v}_r(t) \| \leq \bar{\dot{v}}_r$, which is not written explicitly for presentation clarity. 
	From \eqref{eq:ksi_s_bar}, \eqref{eq:object dynamics 1}, and \eqref{eq:ppc errors}  we also obtain $\| x_{\scriptscriptstyle O}(t) \| \leq \bar{x}_{\scriptscriptstyle O} \coloneqq \bar{x}_\text{d} + \sqrt{6}\bar{\xi}_s\max_{k\in\mathcal{K}}\{\rho_{s_k,0}\}$, and $\|\dot{x}_{\scr O}(t)\|\leq \bar{J}_{\scr O}\bar{v}_{\scr O}$. Next, by using \eqref{eq:coupled_kinematics} and the fact that the rotation matrix $R_{\scriptscriptstyle E_i}(q_i)$ is an orthogonal matrix, we obtain $\| x_{\scriptscriptstyle E_i}(t) \| \coloneqq \| [p^\top_{\scriptscriptstyle E_i}(q_i(t), \eta^\top_{\scriptscriptstyle E_i}(q_i(t)) ]^\top \| \leq \| x_{\scriptscriptstyle O}(t) \| + \| [ (p^{\scriptscriptstyle E_i}_{\scriptscriptstyle E_i/O})^\top, \eta^\top_{\scriptscriptstyle E_i/O}]^\top\|$ and hence, in view of the inverse kinematics of the agents \cite{siciliano2010robotics}, we conclude the boundedness of $q(t)$ as $\| q(t) \| \leq \bar{q}$,
	where $\bar{q}$ is a positive constant. From Assumption \ref{ass:kinematic singularities} and the forward differential agent kinematics, we can also conclude that there exists a positive constant $\bar{J}$ such that $ \| \dot{q}(t) \| \leq \bar{J} \| v \| \leq \bar{J} \sum_{i\in\mathcal{N}}\bar{v}_i$, where $\bar{v}_i$ was defined in \eqref{eq:v_i_bar}. Therefore, we conclude that $\| x(t)\| \leq   \bar{x} \coloneqq  \bar{q} + \bar{J}\sum_{i\in\mathcal{N}}\bar{v}_i + \bar{x}_{\scriptscriptstyle O} +  \bar{J}_{\scriptscriptstyle O} \bar{v}_{\scriptscriptstyle O}$.
	Assumption \ref{ass:disturbance bound_ppc} and the boundedness of $x$ imply that $\|d_i(q_i,\dot{q}_i,t)\| \leq \underline{d}'_i$, $\|d_{\scr O}(x_{\scr O}, \dot{x}_{\scr O},t)\| \leq \underline{d}'_{\scr O}$
	for positive and finite constants $\underline{d}'_{\scr O}$ and $\underline{d}'_i$, respectively, $\forall i\in\mathcal{N}$. Hence, from \eqref{eq:J_O_i bound} and \eqref{eq:coupled dynamics}, we obtain $\| \widetilde{d}(x(t)) \| \leq \underline{d} \coloneqq \underline{d}'_{\scriptscriptstyle O} + \sum_{i\in\mathcal{N}}\{\| p^{\scriptscriptstyle E_i}_{\scriptscriptstyle O/E_i} \| + 1\}\underline{d}'_i$. 
	Similarly, the continuity of $\widetilde{C}(x), \widetilde{g}(x)$ along with the boundedness of $x$ implies the existence of positive and finite constants $\bar{c}, \bar{g}$ such that $\| \widetilde{C}(x(t)) \| \leq \bar{c}$, $\|\widetilde{g}(x(t))\| \leq \bar{g}$. Therefore, we can obtain from  \eqref{eq:V_v dot 1} and \eqref{eq:V_v dot 2}, after some algebraic manipulations, that 
	\begin{align}
	\bar{B}_v \coloneqq & \sqrt{6}\max\limits_{k\in\mathcal{K}}\{l_{v_k}(\rho_{v_k,0} - \rho_{v_k,\infty})\} + \bar{\dot{v}}_r + \bar{m}\Big( \bar{g} + \underline{d} + \notag \\ &   \bar{c}(\bar{v}_r + \sqrt{6}(\| v_r(0) \| + \alpha))  \Big).  \notag
	\end{align} 
	Moreover, by combining \eqref{eq:r_v} and \eqref{eq:ksi_v_bar}, one obtains  $\|r_v(\xi_v(t))\| \leq \bar{r}_v \coloneqq \frac{2}{1-\bar{\xi}^2_{v_k}} = \frac{(\exp(\bar{\varepsilon}_v)+1)^2}{2\exp(\bar{\varepsilon}_v)}$. Finally,
	it can be also shown, from the fact that $p_{\scr O/E_i} = R_{\scr O}(q_i)p^{\scr O}_{\scr O/E_i}, \forall i\in\mathcal{N}$, that the norm $\|J_{M_i}(q)\|$, as defined in \eqref{eq:J_Hirche}, is independent of $q$. Hence, we can also conclude the boundedness of the control inputs \eqref{eq:control_law_ppc} 
	\begin{align}
	& \| u_i(t) \| \leq  \bar{u}_i \coloneqq \notag \\
	&  g_v \|J_{M_i}(q) \| \max\limits_{k\in\mathcal{K}}\Big\{\frac{1}{\rho_{v_k,\infty}} \Big\}  \bar{r}_v\bar{\varepsilon}_v \label{eq:u_i_bar}, \ \forall t\in[0,\tau_{\max}).
	\end{align}}
{By considering \eqref{eq:Lyap_f}, \eqref{eq:V_dot adaptive final}, \eqref{eq:v_f} \eqref{eq:control laws adaptive quat vector form}, we can also derive the respective upper bounds for the controller of Section \ref{subsec:Quaternion Controller}.}

\ifCLASSOPTIONcaptionsoff
  \newpage
\fi



\bibliographystyle{IEEEtran}
\bibliography{bibl}

\begin{thebibliography}{10}
\providecommand{\url}[1]{#1}
\csname url@samestyle\endcsname
\providecommand{\newblock}{\relax}
\providecommand{\bibinfo}[2]{#2}
\providecommand{\BIBentrySTDinterwordspacing}{\spaceskip=0pt\relax}
\providecommand{\BIBentryALTinterwordstretchfactor}{4}
\providecommand{\BIBentryALTinterwordspacing}{\spaceskip=\fontdimen2\font plus
\BIBentryALTinterwordstretchfactor\fontdimen3\font minus
  \fontdimen4\font\relax}
\providecommand{\BIBforeignlanguage}[2]{{%
\expandafter\ifx\csname l@#1\endcsname\relax
\typeout{** WARNING: IEEEtran.bst: No hyphenation pattern has been}%
\typeout{** loaded for the language `#1'. Using the pattern for}%
\typeout{** the default language instead.}%
\else
\language=\csname l@#1\endcsname
\fi
#2}}
\providecommand{\BIBdecl}{\relax}
\BIBdecl

\bibitem{schneider1992object}
S.~A. Schneider and R.~H. Cannon, ``Object impedance control for cooperative
  manipulation: Theory and experimental results,'' \emph{IEEE Transactions on
  Robotics and Automation}, vol.~8, no.~3, pp. 383--394, 1992.

\bibitem{sugar2002control}
T.~G. Sugar and V.~Kumar, ``Control of cooperating mobile manipulators,''
  \emph{IEEE Transactions on robotics and automation}, 2002.

\bibitem{khatib1996decentralized}
O.~Khatib, K.~Yokoi, K.~Chang, D.~Ruspini, R.~Holmberg, and A.~Casal,
  ``Decentralized cooperation between multiple manipulators,'' \emph{IEEE
  International Workshop on Robot and Human Communication}, 1996.

\bibitem{liu1996decentralized}
Y.-H. Liu, S.~Arimoto, and T.~Ogasawara, ``Decentralized cooperation control:
  non-communication object handling,'' \emph{ICRA}, 1996.

\bibitem{liu1998decentralized}
Y.-H. Liu and S.~Arimoto, ``Decentralized adaptive and nonadaptive
  position/force controllers for redundant manipulators in cooperations,''
  \emph{The International Journal of Robotics Research}, 1998.

\bibitem{zribi1992adaptive}
M.~Zribi and S.~Ahmad, ``Adaptive control for multiple cooperative robot
  arms,'' \emph{IEEE Conference on Decision and Control (CDC)}, 1992.

\bibitem{gudino2004control}
J.~Gudi{\~n}o-Lau, M.~A. Arteaga, L.~A. Munoz, and V.~Parra-Vega, ``On the
  control of cooperative robots without velocity measurements,'' \emph{IEEE
  Transactions on Control Systems Technology}, vol.~12, no.~4, 2004.

\bibitem{wen1992motion}
J.~T. Wen and K.~Kreutz-Delgado, ``Motion and force control of multiple robotic
  manipulators,'' \emph{Automatica}, vol.~28, no.~4, pp. 729--743, 1992.

\bibitem{yoshikawa1993coordinated}
T.~Yoshikawa and X.-Z. Zheng, ``Coordinated dynamic hybrid position/force
  control for multiple robot manipulators handling one constrained object,''
  \emph{The International Journal of Robotics Research}, 1993.

\bibitem{kopf1989dynamic}
C.~D. Kopf, ``Dynamic two arm hybrid position/force control,'' \emph{Robotics
  and Autonomous Systems}, vol.~5, no.~4, pp. 369--376, 1989.

\bibitem{caccavale2000task}
F.~Caccavale, P.~Chiacchio, and S.~Chiaverini, ``Task-space regulation of
  cooperative manipulators,'' \emph{Automatica}, vol.~36, no.~6, 2000.

\bibitem{caccavale2008six}
F.~Caccavale, P.~Chiacchio, A.~Marino, and L.~Villani, ``Six-dof impedance
  control of dual-arm cooperative manipulators,'' \emph{IEEE/ASME Transactions
  On Mechatronics}, vol.~13, no.~5, pp. 576--586, 2008.

\bibitem{heck2013internal}
D.~Heck, D.~Kosti{\'c}, A.~Denasi, and H.~Nijmeijer, ``Internal and external
  force-based impedance control for cooperative manipulation,'' \emph{IEEE
  European Control Conference (ECC)}, pp. 2299--2304, 2013.

\bibitem{erhart2013adaptive}
S.~Erhart and S.~Hirche, ``Adaptive force/velocity control for multi-robot
  cooperative manipulation under uncertain kinematic parameters,''
  \emph{IEEE/RSJ International Conference on Intelligent Robots and Systems
  (IROS)}, pp. 307--314, 2013.

\bibitem{erhart2013impedance}
S.~Erhart, D.~Sieber, and S.~Hirche, ``An impedance-based control architecture
  for multi-robot cooperative dual-arm mobile manipulation,'' \emph{IROS}, pp.
  315--322, 2013.

\bibitem{kume2007coordinated}
Y.~Kume, Y.~Hirata, and K.~Kosuge, ``Coordinated motion control of multiple
  mobile manipulators handling a single object without using force/torque
  sensors,'' \emph{IROS}, pp. 4077--4082, 2007.

\bibitem{szewczyk2002planning}
J.~Szewczyk, F.~Plumet, and P.~Bidaud, ``Planning and controlling cooperating
  robots through distributed impedance,'' \emph{Journal of Robotic Systems},
  vol.~19, no.~6, pp. 283--297, 2002.

\bibitem{tsiamis2015cooperative}
A.~Tsiamis, C.~K. Verginis, C.~P. Bechlioulis, and K.~J. Kyriakopoulos,
  ``Cooperative manipulation exploiting only implicit communication,''
  \emph{IROS}, pp. 864--869, 2015.

\bibitem{ficuciello2014cartesian}
F.~Ficuciello, A.~Romano, L.~Villani, and B.~Siciliano, ``Cartesian impedance
  control of redundant manipulators for human-robot co-manipulation,''
  \emph{IROS}, pp. 2120--2125, 2014.

\bibitem{ponce2016cooperative}
A.-N. Ponce-Hinestroza, J.-A. Castro-Castro, H.-I. Guerrero-Reyes,
  V.~Parra-Vega, and E.~Olgu{\`y}n-D{\`y}az, ``Cooperative redundant
  omnidirectional mobile manipulators: Model-free decentralized integral
  sliding modes and passive velocity fields,'' \emph{ICRA}, pp. 2375--2380,
  2016.

\bibitem{gueaieb2007robust}
W.~Gueaieb, F.~Karray, and S.~Al-Sharhan, ``A robust hybrid intelligent
  position/force control scheme for cooperative manipulators,''
  \emph{Transactions on Mechatronics}, vol.~12, no.~2, pp. 109--125, 2007.

\bibitem{Li_fuzzy2015}
Z.~Li, C.~Yang, C.~Y. Su, S.~Deng, F.~Sun, and W.~Zhang, ``Decentralized fuzzy
  control of multiple cooperating robotic manipulators with impedance
  interaction,'' \emph{Transactions on Fuzzy Systems}, 2015.

\bibitem{tzierakis2003independent}
K.~Tzierakis and F.~Koumboulis, ``Independent force and position control for
  cooperating manipulators,'' \emph{Journal of the Franklin Institute}, 2003.

\bibitem{marino2017distributed}
A.~Marino, ``Distributed adaptive control of networked cooperative mobile
  manipulators,'' \emph{Trans. on Control Systems Technology}, 2017.

\bibitem{aghili2013adaptive}
F.~Aghili, ``Adaptive control of manipulators forming closed kinematic chain
  with inaccurate kinematic model,'' \emph{IEEE/ASME Transactions on
  Mechatronics}, vol.~18, no.~5, pp. 1544--1554, 2013.

\bibitem{jean1993adaptive}
J.-H. Jean and L.-C. Fu, ``An adaptive control scheme for coordinated
  multimanipulator systems,'' \emph{IEEE Transactions on Robotics and
  Automation}, vol.~9, no.~2, pp. 226--231, 1993.

\bibitem{sun2002adaptive}
D.~Sun and J.~K. Mills, ``Adaptive synchronized control for coordination of
  multirobot assembly tasks,'' \emph{IEEE Transactions on Robotics and
  Automation}, vol.~18, no.~4, pp. 498--510, 2002.

\bibitem{petitti2016decentralized}
A.~Petitti, A.~Franchi, D.~Di~Paola, and A.~Rizzo, ``Decentralized motion
  control for cooperative manipulation with a team of networked mobile
  manipulators,'' \emph{ICRA}, pp. 441--446, 2016.

\bibitem{wang2015multi}
Z.~Wang and M.~Schwager, ``Multi-robot manipulation with no communication using
  only local measurements,'' \emph{CDC}, pp. 380--385, 2015.

\bibitem{yun1993object}
X.~Yun, ``Object handling using two arms without grasping,'' \emph{The
  International Journal of Robotics Research}, vol.~12, no.~1, 1993.

\bibitem{alonso2017multi}
J.~Alonso-Mora, S.~Baker, and D.~Rus, ``Multi-robot formation control and
  object transport in dynamic environments via constrained optimization,''
  \emph{The International Journal of Robotics Research}, 2017.

\bibitem{bai2010cooperative}
H.~Bai and J.~T. Wen, ``Cooperative load transport: A formation-control
  perspective,'' \emph{Transactions on Robotics}, vol.~26, no.~4, 2010.

\bibitem{tanner2003nonholonomic}
H.~G. Tanner, S.~G. Loizou, and K.~J. Kyriakopoulos, ``Nonholonomic navigation
  and control of cooperating mobile manipulators,'' \emph{Transactions on
  Robotics and Automation}, vol.~19, no.~1, pp. 53--64, 2003.

\bibitem{murphey2008adaptive}
T.~D. Murphey and M.~Horowitz, ``Adaptive cooperative manipulation with
  intermittent contact,'' \emph{ICRA}, pp. 1483--1488, 2008.

\bibitem{MED17_mpc}
A.~Nikou, C.~K. Verginis, S.~Heshmati-alamdari, and D.~V. Dimarogonas, ``A
  nonlinear model predictive control scheme for cooperative manipulation with
  singularity and collision avoidance,'' \emph{Mediterranean Conference on
  Control and Automation}, 2017.

\bibitem{ECC18_mpc}
C.~K. Verginis, A.~Nikou, and D.~V. Dimarogonas, ``Communication-based
  decentralized cooperative object transportation using nonlinear model
  predictive control,'' \emph{European Control Conference}, 2018.

\bibitem{walker1991analysis}
I.~D. Walker, R.~A. Freeman, and S.~I. Marcus, ``Analysis of motion and
  internal loading of objects grasped by multiple cooperating manipulators,''
  \emph{The International journal of robotics research}, 1991.

\bibitem{williams1993virtual}
D.~Williams and O.~Khatib, ``The virtual linkage: a model for internal forces
  in multi-grasp manipulation,'' \emph{ICRA}, vol.~1, pp. 1025--1030, 1993.

\bibitem{chung2005analysis}
J.~H. Chung, B.-Y. Y.~W. K., and Kim, ``Analysis of internal loading at
  multiple robotic systems,'' \emph{Journal of mechanical science and
  technology}, vol.~19, no.~8, pp. 1554--1567, 2005.

\bibitem{erhart2015internal}
S.~Erhart and S.~Hirche, ``Internal force analysis and load distribution for
  cooperative multi-robot manipulation,'' \emph{Transactions on Robotics},
  2015.

\bibitem{erhart2016model}
------, ``Model and analysis of the interaction dynamics in cooperative
  manipulation tasks,'' \emph{Transactions on Robotics}, vol.~32, no.~3, 2016.

\bibitem{Slotine_adaptive87}
J.-J.~E. Slotine and W.~Li, ``On the adaptive control of robot manipulators,''
  \emph{The International Journal of Robotics Research}, 1987.

\bibitem{siciliano2010robotics}
B.~Siciliano, L.~Sciavicco, L.~Villani, and G.~Oriolo, \emph{Robotics:
  modelling, planning and control}.\hskip 1em plus 0.5em minus 0.4em\relax
  Springer Science \& Business Media, 2010.

\bibitem{bechlioulis2008robust}
C.~P. C.~P.~Bechlioulis and G.~A. Rovithakis, ``Robust adaptive control of
  feedback linearizable mimo nonlinear systems with prescribed performance,''
  \emph{Transactions on Automatic Control}, vol.~53, no.~9, 2008.

\bibitem{Verginis_ifac}
C.~K. Verginis, M.~Mastellaro, and D.~V. Dimarogonas, ``Robust quaternion-based
  cooperative manipulation without force/torque information,'' \emph{IFAC
  proceedings}, 2017.

\bibitem{verginis2017timed}
C.~K. Verginis and D.~V. Dimarogonas, ``Timed abstractions for distributed
  cooperative manipulation,'' \emph{Autonomous Robots}, pp. 1--19, 2017.

\bibitem{bechlioulis2014robust}
C.~P. Bechlioulis, M.~V. Liarokapis, and K.~Kyriakopoulos, ``Robust model free
  control of robotic manipulators with prescribed transient and steady state
  performance,'' \emph{IROS}, 2014.

\bibitem{karayiannidis2012model}
Y.~Karayiannidis and Z.~Doulgeri, ``Model-free robot joint position regulation
  and tracking with prescribed performance guarantees,'' \emph{Robotics and
  Autonomous Systems}, vol.~60, no.~2, pp. 214--226, 2012.

\bibitem{doulgeri2009prescribed}
Z.~Doulgeri, Y.~Karayiannidis, and O.~Zoidi, ``Prescribed performance control
  for robot joint trajectory tracking under parametric and model
  uncertainties,'' \emph{MED}, pp. 1313--1318, 2009.

\bibitem{sontag2013mathematical}
E.~D. Sontag, \emph{Mathematical control theory: deterministic finite
  dimensional systems}.\hskip 1em plus 0.5em minus 0.4em\relax Springer Science
  \& Business Media, 2013, vol.~6.

\bibitem{lavretsky13adaptive}
E.~Lavretsky and K.~Wise, ``Robust and adaptive control,''
  \emph{Springer-Verlag, London}, 2013.

\bibitem{campa2006kinematic}
R.~Campa, K.~Camarillo, and L.~Arias, ``Kinematic modeling and control of robot
  manipulators via unit quaternions: Application to a spherical wrist,''
  \emph{CDC}, pp. 6474--6479, 2006.

\bibitem{bhat2000topological}
S.~P. Bhat and D.~S. Bernstein, ``A topological obstruction to continuous
  global stabilization of rotational motion and the unwinding phenomenon,''
  \emph{Systems \& Control Letters}, vol.~39, no.~1, 2000.

\bibitem{mayhew2011quaternion}
C.~G. Mayhew, R.~G. Sanfelice, and A.~R. Teel, ``Quaternion-based hybrid
  control for robust global attitude tracking,'' \emph{IEEE Transactions on
  Automatic Control}, vol.~56, no.~11, pp. 2555--2566, 2011.

\bibitem{lee10}
T.~Lee, M.~Leok, and N.~H. McClamroch, ``Control of complex maneuvers for a
  quadrotor uav using geometric methods on se(3),'' \emph{arXiv:1003.2005},
  2010.

\bibitem{Automatica_formation_18}
C.~K. Verginis, A.~Nikou, and D.~V. Dimarogonas, ``Robust formation control in
  se(3) for tree-graph structures with prescribed transient and steady state
  performance,'' \emph{Automatica}, 2018. Under Review,
  https://arxiv.org/pdf/1803.07513.pdf.

\end{thebibliography}

\end{document}